\documentclass{article} 
\usepackage{iclr2027_conference,times}

\usepackage{hyperref}
\usepackage{url}

\usepackage{graphicx}
\usepackage{booktabs}
\usepackage{amsmath}
\usepackage{amssymb}
\usepackage[table]{xcolor}
\usepackage{colortbl}
\usepackage{multirow}
\usepackage{float}
\usepackage{tabularx}
\usepackage{wrapfig}
\usepackage{makecell}
\usepackage{needspace}
\usepackage{array}
\usepackage{xcolor}
\definecolor{projlink}{RGB}{200,30,120}

\definecolor{bestbg}{RGB}{214,230,247}
\definecolor{secondbg}{RGB}{238,246,253}

\newcommand{\best}[1]{\cellcolor{bestbg}\textbf{#1}}

\newcommand{\second}[1]{\cellcolor{secondbg}#1}

\definecolor{configheader}{RGB}{232,241,249}
\newcolumntype{L}[1]{>{\raggedright\arraybackslash}p{#1}}
\newcolumntype{Y}{>{\raggedright\arraybackslash}X}

\newcolumntype{Z}{>{\raggedright\arraybackslash}X}
\newcolumntype{C}{>{\centering\arraybackslash}X}

\definecolor{MFVSRGroupGray}{RGB}{242,244,247}
\definecolor{MFVSROursBlue}{RGB}{226,239,249}
\definecolor{MFVSRCheck}{RGB}{24,126,113}
\definecolor{MFVSRPartial}{RGB}{190,122,24}
\definecolor{MFVSRCross}{RGB}{151,160,168}
\definecolor{MFVSRMuted}{RGB}{116,128,138}

\providecommand{\mfmark}[2]{}
\renewcommand{\mfmark}[2]{%
    \raisebox{-0.08ex}{%
        \scalebox{1.20}{%
            \textcolor{#1}{\ensuremath{\boldsymbol{#2}}}%
        }%
    }%
}

\providecommand{\mfyes}{}
\providecommand{\mfpartial}{}
\providecommand{\mfno}{}
\providecommand{\mfna}{}

\renewcommand{\mfyes}{\mfmark{MFVSRCheck}{\checkmark}}
\renewcommand{\mfpartial}{\mfmark{MFVSRPartial}{\triangle}}
\renewcommand{\mfno}{\mfmark{MFVSRCross}{\times}}
\renewcommand{\mfna}{\textcolor{MFVSRMuted}{\textbf{--}}}

\title{MFVSR: A \underline{M}ulti-Style and
\underline{F}ine-Grained \\
Benchmark for \underline{V}ideo \underline{S}ubtitle \underline{R}emoval }

\author{%
\textbf{Xiyan Huang}$^{1}$, \quad
\textbf{Qingdong He}$^{2}$, \quad 
\textbf{Peng Tang}$^{2}$, \quad
\textbf{Bo Yin}$^{1}$, \\
\textbf{Chaoyi Wang}$^{3}$, \quad
\textbf{Xiaobin Hu}$^{1}$ \quad \\
{\mdseries $^{1}$National University of Singapore, \ $^{2}$Tencent Youtu Lab,} \\
{\mdseries $^{3}$University of Chinese Academy of Sciences}  \\[2pt]
\textbf{Code:} \textcolor{projlink}{\nolinkurl{https://github.com/hxyiiii/MFVSR}} \\
\textbf{Data:} \textcolor{projlink}{\nolinkurl{https://huggingface.co/datasets/siiian/MFVSR}}
}

\iclrfinalcopy

\definecolor{mycite}{HTML}{0072BC}
\hypersetup{
    colorlinks=true,
    citecolor=mycite,
    linkcolor=mycite,
    urlcolor=mycite
}

\begin{document}

\maketitle

\begin{abstract}
Video subtitle removal (VSR) is an important video restoration task, yet existing evaluation remains limited by low-resolution datasets, fragmented test sets, and generic video-inpainting protocols that fail to adequately capture localized restoration errors, unintended modifications, temporal instability, and incomplete outputs. In this paper, we introduce \textbf{MFVSR}, a large-scale multi-style benchmark for fine-grained and reliability-aware video subtitle removal evaluation. MFVSR distinguishes itself through three key features: \textbf{(1) Large-scale and diverse data}, comprising 3,000 paired videos from Real-HQ, Real-LQ, and AI-Generated sources, with an auxiliary set of 953 in-the-wild real videos for real-world coverage analysis; \textbf{(2) Controlled subtitle diversity}, spanning 16 languages and four factors covering spatial layout, temporal behavior, visual style, and language mode; \textbf{(3) Fine-grained evaluation}, introducing an information-conditioned, spatially decoupled protocol that separates blind from mask-guided removal and measures Restoration Fidelity, Preservation Fidelity, Temporal Fidelity, coverage, and failures. Experiments on six representative methods reveal substantial discrepancies between local restoration, outside-region preservation, temporal consistency, and full-set reliability that are obscured by conventional full-frame metrics. MFVSR therefore provides a more comprehensive and diagnostic foundation for evaluating and advancing video subtitle removal methods. 
\end{abstract}

\section{Introduction}
\label{sec:introduction}
Video Subtitle Removal (VSR) aims to remove burned-in subtitles while reconstructing the occluded background and preserving the original visual content~\citep{kim2019deep,xu2021joint}. Although closely related to video inpainting and scene text removal~\citep{zhang2024avid,ding2025homogen,peng2024viteraser,wang2024tmim}, VSR presents a distinct combination of challenges: subtitle regions must be accurately localized, covered scene content must be faithfully restored, unaffected regions should remain unchanged, and the reconstructed video must remain temporally consistent across frames. These challenges are further complicated by variations in subtitle layout, visual style, language, resolution, and temporal behavior. Consequently, a comprehensive evaluation of existing methods is essential for understanding their capabilities and limitations and for supporting reliable progress in VSR.

However, existing public resources remain insufficient for such systematic VSR evaluation, presenting three core limitations:
\textbf{(L1) Fragmented benchmark coverage.} ChaLearn~\citep{chalearn2018decaptioning,kim2019deep} is limited to short $128\times128$ clips, while recent methods such as SEDiT~\citep{hui2026sedit} and CLEAR~\citep{he2026clear} rely on independently constructed paper-specific test sets. As a result, current evaluations cover only a narrow range of video sources, resolutions, and real-world content distributions, making broad and reproducible comparison difficult. \textbf{(L2) Insufficiently structured subtitle variation.} Existing test sets include diverse subtitle appearances, but these variations are rarely organized along explicit and independently controlled factors such as spatial layout, temporal behavior, visual style, and language. This makes it difficult to attribute performance differences to specific subtitle conditions or to conduct systematic stratified analysis. \textbf{(L3) Generic Benchmarks Lack Subtitle-Specific Evaluation Design.}
Broader video-inpainting evaluation resources and benchmarks~\citep{perazzi2016benchmark,xu2018youtubevos,chang2019fvi,szeto2022devil,wu2025bvinet,miao2025rose} improve content diversity, motion coverage, temporal evaluation, and restoration diagnosis, but are not designed around multilingual subtitle structure, the information asymmetry between blind localization and mask-guided restoration, or spatially decoupled diagnosis with explicit coverage and failure accounting. The core gap is therefore not merely a lack of data, but the absence of a unified benchmark that combines high-resolution paired supervision, controlled subtitle diversity, complementary real-world subtitle coverage, and information-conditioned, spatially decoupled evaluation.

\begin{figure*}[t]
    \centering
    \includegraphics[width=0.98\textwidth]{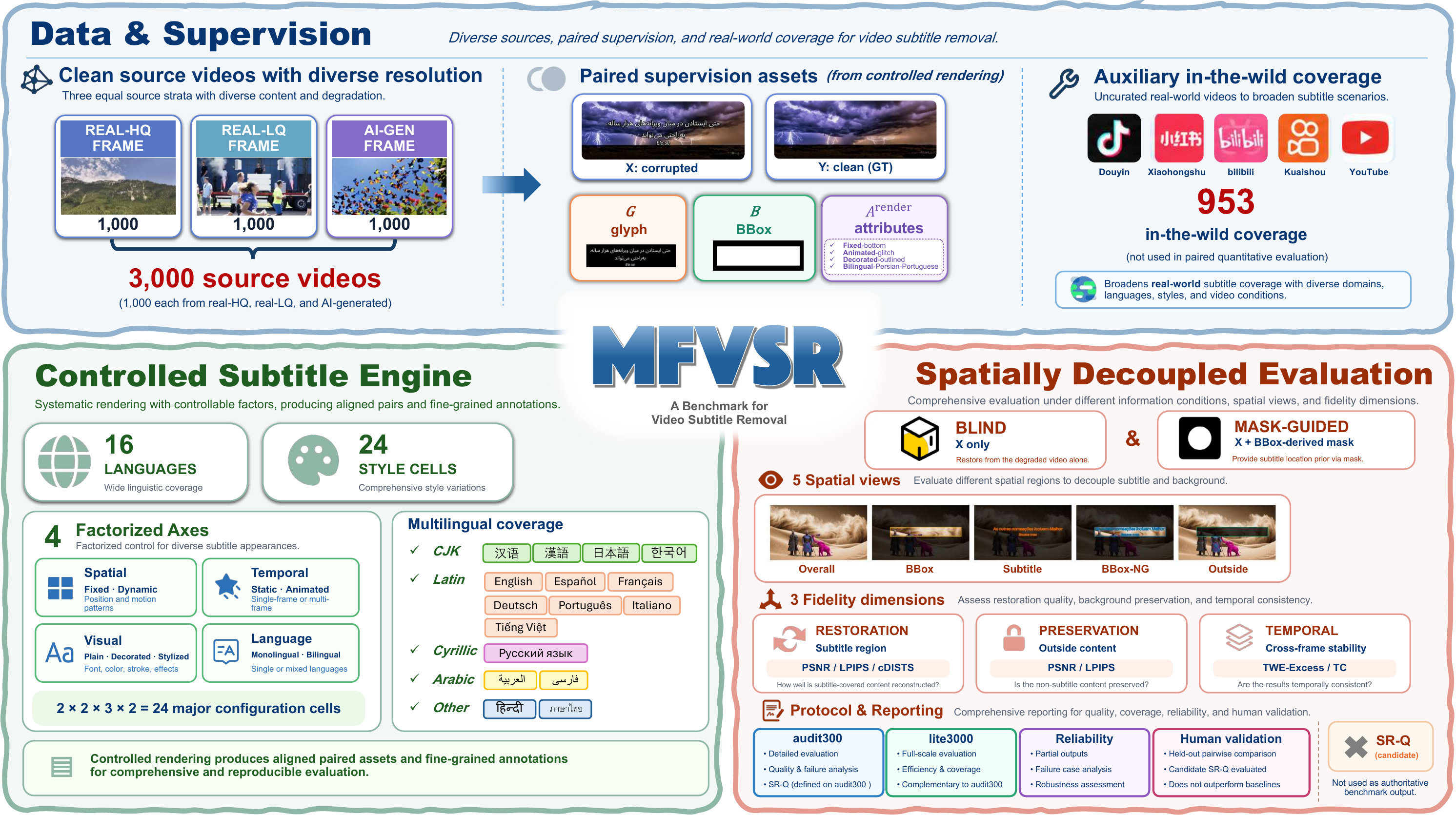}
    \vspace{-5pt}
    \caption{\textbf{Overview of MFVSR.} MFVSR contains 3,000 paired videos from Real-HQ, Real-LQ, and AI-generated sources, together with 953 auxiliary in-the-wild videos. It spans 16 languages and four subtitle factors, and evaluates blind and mask-guided removal through five spatial views and three complementary dimensions: Restoration Fidelity, Preservation Fidelity, and Temporal Fidelity.}
    \label{fig:mfvsr-overview}
    \vspace{-10pt}
\end{figure*}

To directly address these limitations, we propose \textbf{MFVSR}, a Multi-style and Fine-grained benchmark for Video Subtitle Removal,  as shown in Figure~\ref{fig:mfvsr-overview}. MFVSR is built around three key contributions:
\textbf{(1) Large-scale and diverse data.} MFVSR contains 3,000 resolution-diverse paired videos, balanced equally across the Real-HQ, Real-LQ, and AI-Generated source strata. Controlled subtitle rendering provides paired clean targets, pixel-level glyph masks, frame-level BBoxes, BBox-derived model-input masks, and rendering
metadata, enabling fine-grained evaluation against known clean content. In addition, we curate an
auxiliary set of 953 naturally subtitled in-the-wild real videos from Bilibili, Douyin, Xiaohongshu, Kuaishou, and YouTube to broaden coverage of real-world subtitle appearances and content
distributions.
\textbf{(2) Controlled subtitle diversity.}
MFVSR covers 16 languages and factorizes subtitle rendering into four independently configurable factors: Spatial Layout, Temporal Behavior, Visual Style, and Language Mode, yielding 24 configuration cells. This controlled design facilitates standardized cross-method comparison and stratified analysis, without presupposing that any language or rendering factor is inherently more challenging.
\textbf{(3) Fine-grained evaluation.}
MFVSR separates blind subtitle removal from mask-guided removal. Predictions are evaluated using five spatial views and grouped into Restoration Fidelity, Preservation Fidelity, and Temporal Fidelity. It also jointly reports conditional quality, spatial and temporal coverage, partial outputs, and full-benchmark failures.

We evaluate six representative subtitle-removal, video-inpainting, and object-removal methods under MFVSR.
The results reveal distinct trade-offs across restoration, preservation, temporal fidelity, and reliability that conventional full-frame similarity does not capture.
Held-out human comparisons provide stronger support for the spatial components than for the temporal component, and do not establish an advantage of SR-Q over conventional Overall metrics.
MFVSR thus provides a more diagnostic basis for comparing VSR systems.
Benchmark resources will be made publicly available to support reproducible research and future VSR development.

\section{MFVSR Benchmark Construction}
\label{sec:benchmark-construction}


\subsection{Multi-Source Collection and Base-Video Filtering}
\label{sec:base-video-filtering}

\textbf{Multi-source collection for MFVSR.}
To cover different content origins, resolutions, durations, scenes, and motion patterns, MFVSR contains 3,000 clean base videos from real and generated sources. We divide them into three equal strata. Real-HQ contains 1,000 real videos from OpenVidHD and Pexels~\citep{nan2024openvid,pexelslicense}, with retained resolution at or above 1080p. Real-LQ contains 1,000 real videos from Panda-70M and Pexels~\citep{chen2024panda70m,pexelslicense}, with retained resolution at or above 512p and below 1080p. AI-Generated contains 1,000 videos from the VBench-2.0 sampled-video release~\citep{zheng2025vbench2} and the Rapidata Sora video-generation alignment dataset~\citep{rapidataSora}. Real-HQ and Real-LQ are resolution-based strata rather than perceptual-quality labels, and the three sources do not imply exhaustive content coverage.

\begin{figure*}[t]
    \centering
    \includegraphics[width=1.0\linewidth]{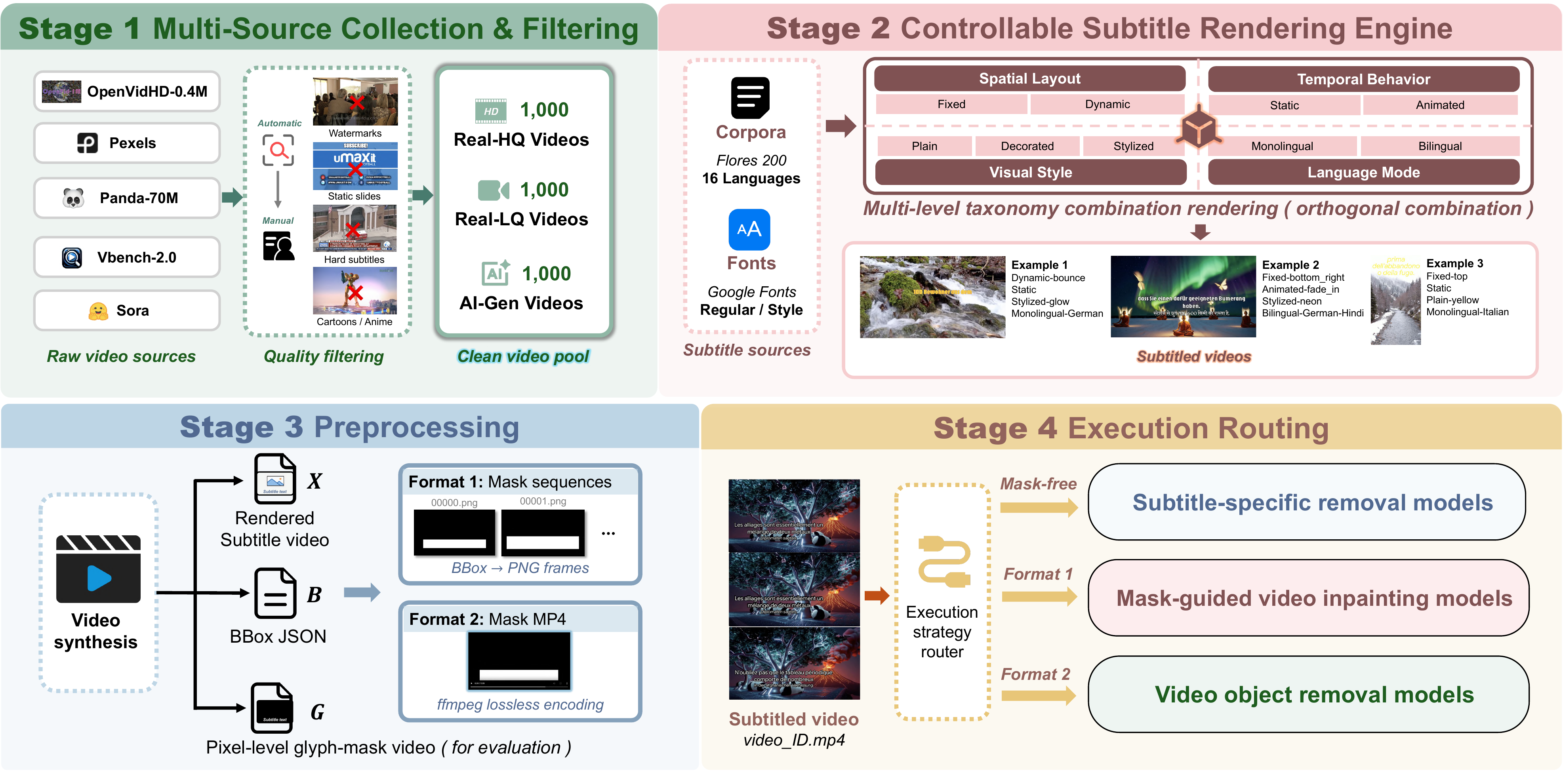}
    \caption{\textbf{Dataset construction and unified processing of MFVSR} consists of: \textbf{(1)} Multi-source collection and filtering to provide a clean video pool (Section~\ref{sec:base-video-filtering}); \textbf{(2)} Controllable subtitle rendering to generate diverse subtitled videos with different attributes (Section~\ref{sec:subtitle-rendering}); \textbf{(3)} Preprocessing to construct paired supervision assets (Section~\ref{sec:benchmark-assets}); \textbf{(4)} Execution routing to assign each sample to the corresponding model (Section~\ref{sec:model-interfaces}).}
    \label{fig:Dataset construction}
\end{figure*}
\textbf{Automatic and manual filtering.}
We use automatic screening followed by manual verification. Automatic screening samples frames, checks decoding and basic validity, filters by resolution and preliminary clarity, and uses OCR to flag burned-in text, URLs, and other overlays. Manual verification removes videos with burned-in subtitles, salient watermarks, subtitle-like text overlays, static slideshows, cartoons, anime, or other content unsuitable for paired subtitle rendering. Small logos may remain if they are fixed in a corner and cannot be confused with the rendered subtitles. Large, prominent, textual, or subtitle-region watermarks are excluded. The resulting clean videos enter the rendering pipeline in Figure~\ref{fig:Dataset construction}. Additional asset-validation checks are described in Appendix~\ref{app:dataset-qa}.
\subsection{Controllable Subtitle Rendering Engine}
\label{sec:subtitle-rendering}


\textbf{Text and font resources.}
Subtitle text is drawn from FLORES-200~\citep{nllb2022,goyal2022flores,guzman2019two}, and fonts are obtained from Google Fonts~\citep{googlefonts}. MFVSR uses 16 languages: Simplified Chinese, Traditional Chinese, Japanese, and Korean for CJK scripts; English, Spanish, French, German, Portuguese, Italian, and Vietnamese for Latin scripts; Russian for Cyrillic; Arabic and Persian for right-to-left scripts; and Hindi and Thai for additional writing systems. These languages introduce different glyph structures, shaping rules, reading directions, and layout characteristics without assuming a language-specific difficulty order. Fonts are divided into regular and style-oriented resources.

\begin{figure}[t]
    \centering
    \includegraphics[width=0.95\linewidth]{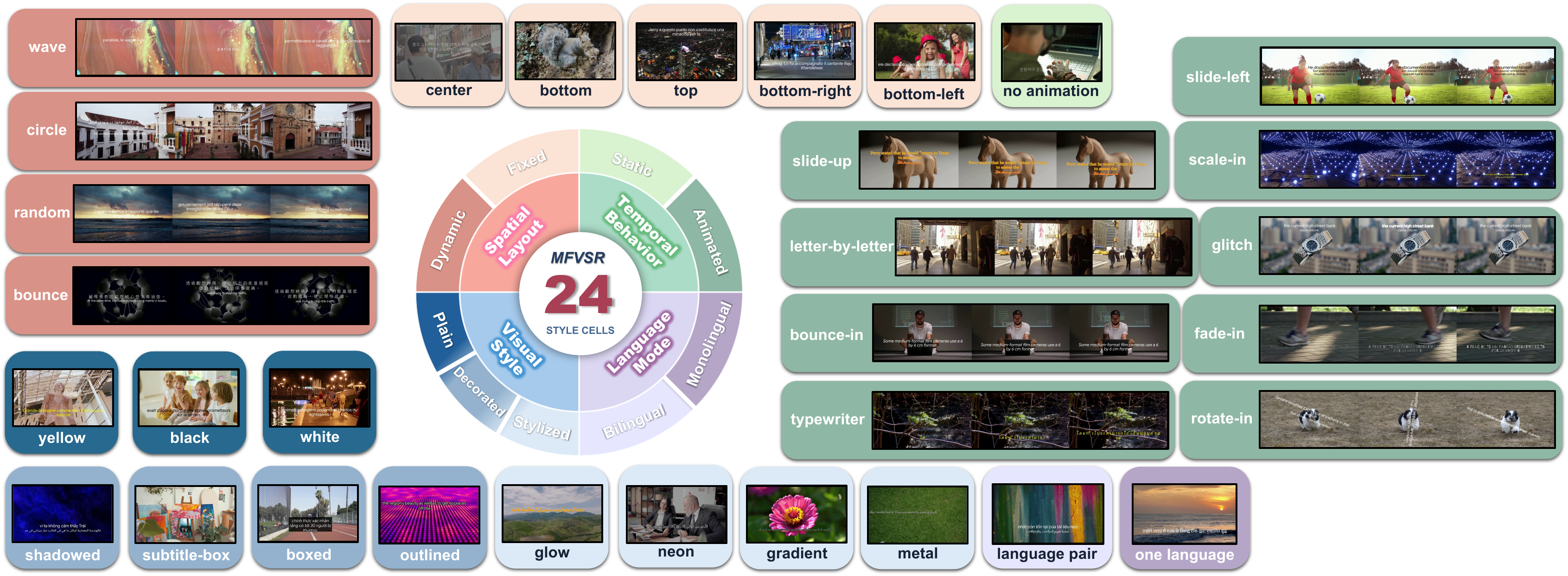}
    \vspace{-5pt}
    \caption{\textbf{Controlled subtitle-rendering taxonomy in MFVSR.} The two-level ring summarizes four independently configurable factors: Spatial Layout, Temporal Behavior, Visual Style, and Language Mode and their major options, forming 24 configuration cells. Surrounding examples are color-coded by factor to illustrate finer-grained within-cell variations.}
    \label{fig:subtitle-taxonomy}
\end{figure}

\textbf{Multi-level taxonomy combination rendering.} MFVSR structures the taxonomy around the changing aspect of a subtitle over its display interval, spanning four independently configurable axes. The four independently configurable factors form
\[
\text{Spatial Layout} \times \text{Temporal Behavior} \times
\text{Visual Style} \times \text{Language Mode},
\]
which yields $2 \times 2 \times 3 \times 2 = 24$ major configuration cells. And finer-grained subtypes add further within-cell variation, as shown in Figure~\ref{fig:subtitle-taxonomy}.

\textbf{Spatial layout} governs where a subtitle appears and whether its coordinates change over time: fixed subtitles occupy one of five anchors with constant coordinates, whereas dynamic subtitles follow one of four trajectories, making position a function of time. 
\textbf{Temporal behavior} captures how a subtitle evolves between appearance and disappearance. Static subtitles remain unchanged while visible, whereas animated subtitles comprise nine subtypes that cover entrance transformations of whole blocks, reveals at the character level, and persistent perturbations. Spatial layout and temporal behavior are orthogonal and may occur together. For example, a caption can translate from point $\mathcal{A}$ to point $\mathcal{B}$ (spatial, dynamic), while a fade-in or typewriter effect alters its form without displacing it (temporal, animated). Therefore, we retain them as separate axes. 
\textbf{Visual style} determines how glyphs are rendered, ranging from plain (color variation only), to decorated (shadows, outlines, or background boxes for functional separation), to stylized (glow, gradients, neon, textures, or expressive fonts). 
\textbf{Language mode} sets how many languages appear simultaneously: monolingual subtitles contain a single language, while bilingual subtitles pair two randomly selected languages, typically arranged in two lines, and configurations with more than two languages are excluded as uncommon in practice. 

The four axes combine to produce 24 major cells. While the factors are independent in configuration, their effects on removal difficulty may not be statistically independent. Therefore, potential mediators like subtitle area and line count are evaluated separately for each factor.
\subsection{Preprocessing and Paired-Asset Generation}
\label{sec:benchmark-assets}

For each clean base, video synthesis yields three aligned outputs: a rendered subtitle video, a pixel-level glyph-mask video, and per-frame BBox JSON. The clean base is retained as ground truth, and the rendering configuration is stored as metadata. The BBox JSON is the canonical rectangular annotation for mask-guided models. From it, we generate two equivalent mask representations:  \emph{Mask Sequences} (frame-wise binary PNG masks) and \emph{Mask MP4} (the same sequence losslessly encoded as a video). These formats encode identical spatial and temporal mask information and are not separate ground-truth annotations. The glyph mask is reserved for fine-grained evaluation, whereas the BBox-derived mask is used as model input. 

\begin{table*}[t]
    \centering
    \caption{
        \textbf{Characteristics and MFVSR input interfaces of evaluated methods.}
        Detailed inference configurations are in Table~\ref{tab:inference-configurations}.
    }
    \label{tab:model-interfaces}

    \small
    \renewcommand{\arraystretch}{1.15}
    \setlength{\tabcolsep}{3pt}
    \resizebox{\textwidth}{!}{%
        \begin{tabularx}{\textwidth}{@{} l p{5cm} c X p{3.3cm} @{}}
            \toprule
            \textbf{Category} & \textbf{Method} & \textbf{Mask} & \textbf{Backbone} & \textbf{Temporal mechanism} \\
            \midrule
            Inpainting & ProPainter~\citep{zhou2023propainter} & Seq. & Flow + Sparse ViT & Flow-guided propagation \\
            Inpainting & DiffuEraser~\citep{li2025diffueraser} & MP4 & SD~1.5 + BrushNet & Motion module w/ prior \\
            \addlinespace[1.5pt]
            Object & MiniMax-Remover~\citep{zi2026minimax} & MP4 & Wan DiT & Full-seq.\ attention \\
            Object & OmnimatteZero~\citep{samuel2025omnimattezero} & MP4 & LTX DiT & Attn.\ guidance \\
            Object & EffectErase~\citep{fu2026effecterase} & MP4 & Wan DiT & Region guidance \\
            \addlinespace[1.5pt]
            Subtitle & CLEAR~\citep{he2026clear} & None & Wan DiT & Spatio-temporal attention \\
            
            \bottomrule
        \end{tabularx}
    }
\end{table*}
\subsection{Execution Routing}
\label{sec:model-interfaces}
MFVSR routes each integrated model through deterministic input preprocessing.
All mask-guided methods start from the same canonical BBox-derived annotation.
Model-specific operations, such as resizing, thresholding, and dilation when
required, follow the evaluated inference configuration, and no routing or
preprocessing step accesses the glyph mask, clean target, or manually adjusted
sample-specific conditions. Blind removal takes only the subtitle-corrupted video as input, whereas mask-guided removal additionally receives the canonical BBox-derived mask defined above. MFVSR therefore ranks the two families separately.
Table~\ref{tab:model-interfaces} lists the resulting model interfaces.

\subsection{Statistics and Analysis}
\label{sec:benchmark-statistics}
\begin{figure*}[t]
    \centering
    \includegraphics[width=1\textwidth]{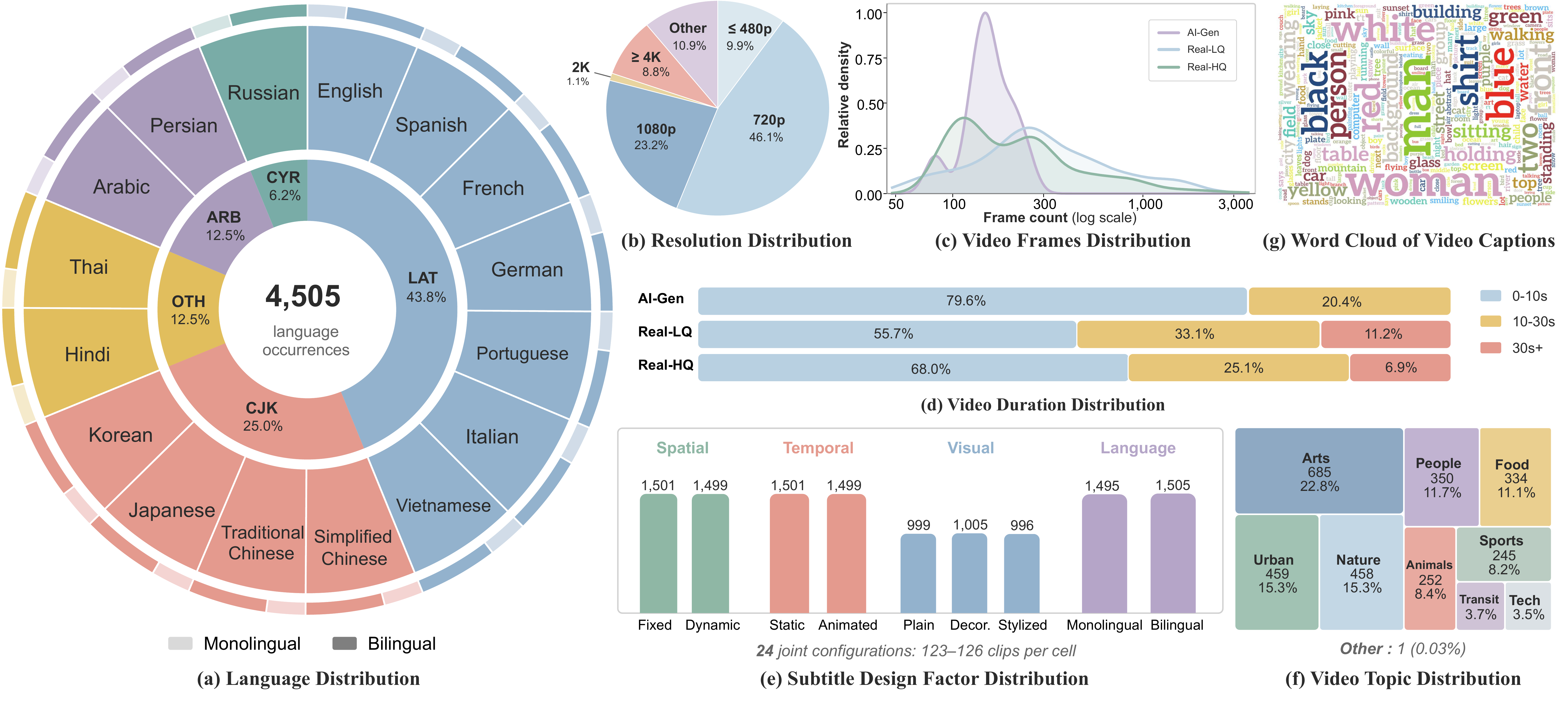}
    \vspace{-18pt}
    \caption{
    \textbf{MFVSR demonstrates rich diversity and balanced coverage.}
    \textbf{Subtitle Statistics:}
    (a) shows a sunburst chart of language coverage across 16 languages, five script groups, and monolingual/bilingual settings;
    (e) shows balanced distributions across spatial, temporal, visual style, and language factors, whose full factorial combination forms 24 configurations with 123-126 videos each.
    \textbf{Video Statistics:}
    (b) indicates diverse video resolutions across AI-Gen, Real-LQ, and Real-HQ;
    (c) presents frame-count distributions with substantial variation in video length;
    (d) shows video durations ranging from short clips to sequences longer than 30 seconds;
    (f) illustrates video content distributed across nine semantic categories;
    (g) presents a word cloud of video captions, covering diverse objects, scenes, attributes, and activities.
    }
    \label{fig:benchmark_statistics}
\end{figure*}

\noindent\textbf{Video statistics.}
MFVSR contains a diverse collection of videos representing a wide range of visual content, source qualities, and temporal characteristics.
Figure~\ref{fig:benchmark_statistics}(b) presents the resolution distributions of the three source subsets, where Real-LQ is dominated by 720p videos and Real-HQ primarily contains 1080p and 4K videos, while AI-Gen exhibits a distinct lower-resolution profile.
Figure~\ref{fig:benchmark_statistics}(c) shows the frame-count distributions, capturing substantial variation in sequence length across the three subsets.
Figure~\ref{fig:benchmark_statistics}(d) presents the distribution of video durations, ranging from short clips below 10 seconds to longer sequences exceeding 30 seconds.
Figure~\ref{fig:benchmark_statistics}(f) shows the distribution of video content across nine major topics, with Arts, Urban, and Nature forming the largest categories.
Figure~\ref{fig:benchmark_statistics}(g) illustrates a word cloud constructed from video captions generated using the LLaVA-Llama-3-8B-v1.1~\citep{liu2024improved,xtuner_llava_llama3_8b_v11}, highlighting diverse people, objects, scenes, colors, and activities represented in the benchmark.

\noindent\textbf{Subtitle statistics.}
MFVSR provides diverse multilingual subtitles with balanced coverage of controlled subtitle configurations.
Figure~\ref{fig:benchmark_statistics}(a) shows a sunburst chart of 4,505 language occurrences across 16 languages and five script groups, including both monolingual and bilingual settings. These groups capture script-level visual differences in glyph density, stroke complexity, diacritics, connected writing, and writing direction that directly affect subtitle localization, removal, and reconstruction.
Figure~\ref{fig:benchmark_statistics}(e) summarizes four controlled subtitle design factors and their full factorial combination yields 24 joint configurations, with 123-126 videos per configuration, providing nearly uniform coverage of the predefined subtitle design space.

\noindent\textbf{Comparison with existing evaluation resources.}
Table~\ref{tab:benchmark-comparison} compares MFVSR with representative video inpainting, object-removal, and subtitle-removal evaluation resources. MFVSR combines 3,000 paired clips and controlled subtitle strata with separate blind and mask-guided reporting, spatially decoupled and temporal evaluation, and explicit statistics on execution failures and valid evaluation coverage.

\begin{table*}[t]
    \centering
    \caption{
        \textbf{Comparison with related evaluation resources.}
        $\checkmark$, $\triangle$, $\times$, and -- denote full, partial,
        unsupported, and not-applicable settings, respectively.
    }
    \label{tab:benchmark-comparison}

    \footnotesize
    \setlength{\tabcolsep}{2.7pt}
    \renewcommand{\arraystretch}{1.24}

    \resizebox{\textwidth}{!}{%
    \begin{tabular}{l r c c c c c c c c c}
        \toprule[1.5pt]

        \multirow{2}{*}{\textbf{Benchmark}} &
        \multicolumn{2}{c}{\textbf{Resource}} &
        \multicolumn{2}{c}{\textbf{Task setting}} &
        \multicolumn{5}{c}{\textbf{Evaluation design}} &
        \multirow{2}{*}{\textbf{Venue}}
        \\

        \cmidrule(lr){2-3}
        \cmidrule(lr){4-5}
        \cmidrule(lr){6-10}

        &
        Videos &
        Paired GT &
        Input &
        Sep. &
        Strat. &
        Region &
        Temp. &
        Failure &
        Coverage &
        \\

        \specialrule{0.65pt}{0pt}{0pt}

        \rowcolor{MFVSRGroupGray}
        \multicolumn{11}{l}{%
            \rule[-0.72ex]{0pt}{3.15ex}%
            \hspace{4pt}%
            \textit{\textbf{Video Inpainting Evaluation}}%
        }
        \\

        FVI test set~\citep{chang2019fvi} &
        100 &
        \mfyes & M & \mfna &
        \mfyes & \mfno & \mfpartial & \mfno & \mfno &
        ICCV 2019
        \\

        DEVIL~\citep{szeto2022devil} &
        1,250 &
        \mfyes & M & \mfna &
        \mfyes & \mfpartial & \mfyes & \mfno & \mfno &
        CVPR 2022
        \\

        BVINet evaluation~\citep{wu2025bvinet} &
        568 &
        \mfyes & B+M & \mfyes &
        \mfno & \mfno & \mfyes & \mfno & \mfno &
        ICCV 2025
        \\

        \specialrule{0.65pt}{0pt}{0pt}

        \rowcolor{MFVSRGroupGray}
        \multicolumn{11}{l}{%
            \rule[-0.72ex]{0pt}{3.15ex}%
            \hspace{4pt}%
            \textit{\textbf{Video Object Removal}}%
        }
        \\

        HQVI~\citep{cho2025hqvi} &
        \textit{N/V} &
        \mfyes & M & \mfna &
        \mfno & \mfno & \mfpartial & \mfno & \mfno &
        AAAI 2025
        \\

        ROSE-Bench (syn.)~\citep{miao2025rose} &
        60 &
        \mfyes & M & \mfna &
        \mfyes & \mfno & \mfno & \mfno & \mfno &
        NeurIPS 2025
        \\

        PROVE-Bench~\citep{li2026prove} &
        180$^{\dagger}$ &
        \mfpartial & M & \mfna &
        \mfyes & \mfyes & \mfyes & \mfno & \mfno &
        ACM MM 2026
        \\

        \specialrule{0.65pt}{0pt}{0pt}

        \rowcolor{MFVSRGroupGray}
        \multicolumn{11}{l}{%
            \rule[-0.72ex]{0pt}{3.15ex}%
            \hspace{4pt}%
            \textit{\textbf{Video Subtitle Removal}}%
        }
        \\

        ChaLearn Decaptioning~\citep{chalearn2018decaptioning} &
        $\approx$13.5K$^{\ddagger}$  &
        \mfyes & B & \mfna &
        \mfno & \mfno & \mfno & \mfno & \mfno &
        WCCI/ECCV 2018
        \\

        CLEAR test set~\citep{he2026clear} &
        400 &
        \mfyes & B+M & \mfno &
        \mfno & \mfno & \mfyes & \mfno & \mfno &
        ICML 2026
        \\

        VSR-Bench-400~\citep{hui2026sedit} &
        400 &
        \mfyes & B+M & \mfno &
        \mfno & \mfno & \mfpartial & \mfno & \mfno &
        arXiv
        \\

        \specialrule{1.05pt}{0pt}{0pt}

        \rowcolor{MFVSROursBlue}
        \textbf{\rule[-0.78ex]{0pt}{3.30ex}MFVSR (Ours)} &
        \textbf{3,000} &
        \mfyes & B+M & \mfyes &
        \mfyes & \mfyes & \mfyes & \mfyes & \mfyes &
        \mfna
        \\

        \specialrule{1.5pt}{0pt}{0pt}
    \end{tabular}%
    }

    \vspace{2pt}
    \begin{minipage}{\textwidth}
    \scriptsize\raggedright
    B/M: blind/mask-guided; Sep.: separate reporting by input setting;
    Strat.: stratified evaluation; N/V: count not verified.
    $^{\dagger}$80 paired + 100 unpaired videos.
    $^{\ddagger}$Estimated test size from the reported $\sim$90K clips and 15\% test split.
    \end{minipage}
\end{table*}

\section{Experiments}
\label{sec:experiments}

\subsection{Experimental Setting}
\label{sec:evaluation-protocol}

\noindent\textbf{Methods and fairness.}
Section~\ref{sec:benchmark-construction} constructs each MFVSR instance
by rendering temporally controlled subtitles over a clean base video. Given a clean target $\mathbf{Y}$, the renderer produces a subtitle-corrupted video $\mathbf{X}$, glyph mask $\mathbf{G}$, frame-level BBoxes $\mathbf{B}$, and identifier-encoded rendering attributes $\mathcal{A}^{\mathrm{render}}$:
\begin{equation}
    \mathcal{I}=(\mathbf{X},\mathbf{Y},\mathbf{G},\mathbf{B},\mathcal{A}^{\mathrm{render}}).
    \label{eq:mfvsr-instance}
\end{equation}
The attributes describe language mode, visual style, spatial layout, temporal behavior, and the stored font, color, decoration, animation, and trajectory parameters. We evaluate one blind method, CLEAR, and five mask-guided methods: ProPainter, DiffuEraser, MiniMax-Remover, OmnimatteZero, and EffectErase. Blind removal receives only $\mathbf{X}$; mask-guided removal additionally receives the same canonical BBox-derived mask. The glyph mask is reserved for evaluation and is never exposed as model input. Because the two settings provide different information, they share targets and metrics but remain in separate result blocks. Every method retains its released resolution limit, precision, chunking, sampling, and deterministic mask adapter. Raw outputs are immutable: missing frames are neither copied nor padded, and a metric is withheld when its spatial or temporal correspondence cannot be verified. All model interfaces are frozen in Table~\ref{tab:inference-configurations}.

\noindent\textbf{Spatially decoupled evaluation.}
As illustrated in Figure~\ref{fig:mfvsr-overview}, MFVSR evaluates five views: Overall, BBox, Subtitle, BBox-NG, and Outside. Subtitle, BBox-NG, and Outside partition each frame, BBox is the union of the first two, and Overall is the full frame. Because the glyph mask can omit shadows, outlines, glow, or subtitle boxes, BBox-NG mixes nearby background with non-core subtitle effects and is used only as a local diagnostic. Boxed and Subtitle-box styles are analyzed separately because their affected support can extend beyond the text BBox.

The five views are organized into three capabilities (Table~\ref{tab:metric-view-main}). \textbf{Restoration Fidelity} asks whether the subtitle core is replaced by the clean target; \textbf{Preservation Fidelity} asks whether content outside the operational BBox remains unchanged; and \textbf{Temporal Fidelity} asks whether the prediction introduces temporal error beyond the target motion. PSNR, SSIM~\citep{wang2004ssim}, LPIPS~\citep{zhang2018lpips}, DISTS~\citep{ding2020dists}, and RAFT~\citep{teed2020raft} retain their released implementations and weights. Regional PSNR, SSIM, and LPIPS change only the aggregation support; conditioned DISTS is explicitly a controlled-input intervention, not an official dense DISTS variant. Overall metrics are retained for conventional full-frame compatibility rather than used as the primary evidence.

\begin{table*}[t]
    \centering
    \caption{\textbf{Metric--view design. }Bold views and metrics form the three primary dimensions; the remaining views are diagnostic or compatibility measurements. TC denotes flow-free Temporal Change Error, and TWE-X denotes RAFT-based TWE-Excess.}
    \label{tab:metric-view-main}
    \small
    \renewcommand{\arraystretch}{1.1}
    \setlength{\tabcolsep}{6pt}
    \begin{tabular}{@{}>{\raggedright\arraybackslash}p{0.10\textwidth}
                       >{\raggedright\arraybackslash}p{0.42\textwidth}
                       >{\raggedright\arraybackslash}p{0.14\textwidth}
                       >{\raggedright\arraybackslash}p{0.24\textwidth}@{}}
        \toprule[1.2pt]
        \textbf{Dimension} & \textbf{Central question} & \textbf{Primary views} & \textbf{Representative evidence} \\
        \midrule
        \addlinespace[1pt]
        Restoration Fidelity
            & Is content covered by the subtitle reconstructed toward the paired clean target?
            & Subtitle
            & PSNR and LPIPS; conditioned DISTS\\
        \addlinespace[2pt]
        Preservation Fidelity
            & Does content outside the operational edit box remain close to the paired clean target?
            & Outside
            & PSNR, LPIPS \\
        \addlinespace[2pt]
        Temporal Fidelity
            & Does the output preserve legitimate target dynamics without model-induced instability?
            & BBox, Outside
            & Temporal Change Error; TWE-Excess\\
        \addlinespace[1pt]
        \bottomrule[1.2pt]
    \end{tabular}
    \vspace{-10pt}
\end{table*}

\noindent\textbf{Evaluation tracks and Audit300.}
We separate benchmark coverage from measurement density.
Lite3000 provides broad spatial and temporal evaluation, whereas Audit300 adds more expensive structural and motion-compensated metrics on a fixed 300-video subset.
Sampling is frozen and shared across methods, and Audit300 is drawn from the common-evaluable pool using only dataset attributes, without access to model outputs, metric values, rankings, or human judgments. This provides a controlled shared-ID set for detailed cross-method analysis.
Selection details and composition are reported in
Appendix~\ref{app:audit300-selection}.

\noindent\textbf{Statistics and human alignment.}
Unless otherwise stated, quality results are video-level macro estimates with bootstrap 95\% confidence intervals, while delivery statistics use all scheduled videos.
Cross-model comparisons use shared IDs and symmetric metric support, and model-native results are interpreted only conditionally.
Reliability is reported separately from quality through complete, partial, and failed or missing outputs.
Held-out human comparisons show stronger alignment for the spatial dimensions than for Temporal Fidelity, while the preregistered SR-Q summary shows no advantage over conventional Overall metrics and is retained only as a secondary analysis.
Full statistical and human-study procedures are provided in
Appendix~\ref{app:evaluation-details}.

\subsection{Overall Results}
\begin{table*}[t]
    \centering
    \caption{
    \textbf{Raw metrics on $\mathcal{A}_{300}$.}
    PSNR is in dB; TC and TWE-X are scaled by $10^2$ and $10^3$.
    Dark and light shading mark the best and second-best mask-guided means.
    $^{*}$EffectErase is evaluated on its verified prefix.
    }
    \label{tab:audit300-raw-results}

    \scriptsize
    \setlength{\tabcolsep}{3.2pt}
    \renewcommand{\arraystretch}{1.12}

    \resizebox{\textwidth}{!}{%
    \begin{tabular}{@{}llcccccccccc@{}}
        \toprule[1.4pt]

        \multirow{2}{*}{\textbf{Setting}}
        & \multirow{2}{*}{\textbf{Method}}
        & \multicolumn{3}{c}{\textbf{Restoration: Subtitle}}
        & \multicolumn{2}{c}{\textbf{Preservation: Outside}}
        & \multicolumn{4}{c}{\textbf{Temporal Fidelity}}
        & \multirow{2}{*}{
            \makecell{\textbf{Joint cov.}\\\textbf{(\%) $\uparrow$}}
        } \\

        \cmidrule(lr){3-5}
        \cmidrule(lr){6-7}
        \cmidrule(lr){8-11}

        &
        & \textbf{PSNR $\uparrow$}
        & \textbf{LPIPS $\downarrow$}
        & \textbf{cDISTS $\downarrow$}
        & \textbf{PSNR $\uparrow$}
        & \textbf{LPIPS $\downarrow$}
        & \textbf{BBox TC $\downarrow$}
        & \textbf{Out. TC $\downarrow$}
        & \textbf{BBox TWE-X $\downarrow$}
        & \textbf{Out. TWE-X $\downarrow$}
        & \\

        \midrule

        Blind
        & CLEAR
        & 11.99
        & 0.621
        & 0.236
        & 21.48
        & 0.221
        & 5.00
        & 1.77
        & 15.042
        & 0.076
        & 85.4 \\

        \midrule

        \multirow{5}{*}{Mask-guided}
        & ProPainter
        & 19.66
        & 0.364
        & 0.138
        & \best{34.11}
        & \best{0.050}
        & 2.57
        & \best{0.70}
        & 0.683
        & 0.072
        & \best{100.0} \\

        & DiffuEraser
        & \best{21.36}
        & \second{0.269}
        & \best{0.116}
        & \second{32.87}
        & \second{0.062}
        & 2.53
        & \second{0.73}
        & 0.686
        & 0.068
        & \second{99.9} \\

        & MiniMax-Remover
        & 20.32
        & 0.321
        & \second{0.132}
        & 30.73
        & 0.086
        & \second{2.52}
        & 1.12
        & 1.142
        & \best{0.028}
        & 99.4 \\

        & OmnimatteZero
        & 17.69
        & 0.369
        & 0.160
        & 26.00
        & 0.131
        & 2.77
        & 1.51
        & \second{0.653}
        & \second{0.048}
        & 98.5 \\

        & EffectErase$^{*}$
        & \second{20.58}
        & \best{0.251}
        & 0.138
        & 25.77
        & 0.120
        & \best{2.39}
        & 1.51
        & \best{0.634}
        & 0.129
        & 43.1 \\

        \bottomrule[1.4pt]
    \end{tabular}%
    }
\end{table*}

Table~\ref{tab:audit300-raw-results} reports Audit300 results under matched per-metric support across all six methods.
Confidence intervals and additional metrics are provided in
Appendix~\ref{app:audit300-complete-results}.
Blind and mask-guided methods are reported separately because they use different input information.

\noindent\textbf{Shared-video quality decomposes into non-interchangeable capabilities.}
The strongest methods differ across Restoration, Preservation, and Temporal fidelity even on the same Audit300 videos.
DiffuEraser favors glyph-core reconstruction, ProPainter Outside preservation and temporal consistency, and MiniMax-Remover Outside TWE-X, while EffectErase achieves favorable conditional values only on its verified prefix.
Thus, gains in one capability do not imply gains in another, and conditional quality must be interpreted together with output support. These results motivate a capability profile rather than a single aggregate ordering.

\subsection{Findings of MFVSR}
\label{sec:findings}
We present five findings from MFVSR, with supporting analyses in
Appendix~\ref{app:detailed-findings}.

\begin{figure*}[t]
    \centering
    \includegraphics[width=0.85\textwidth]{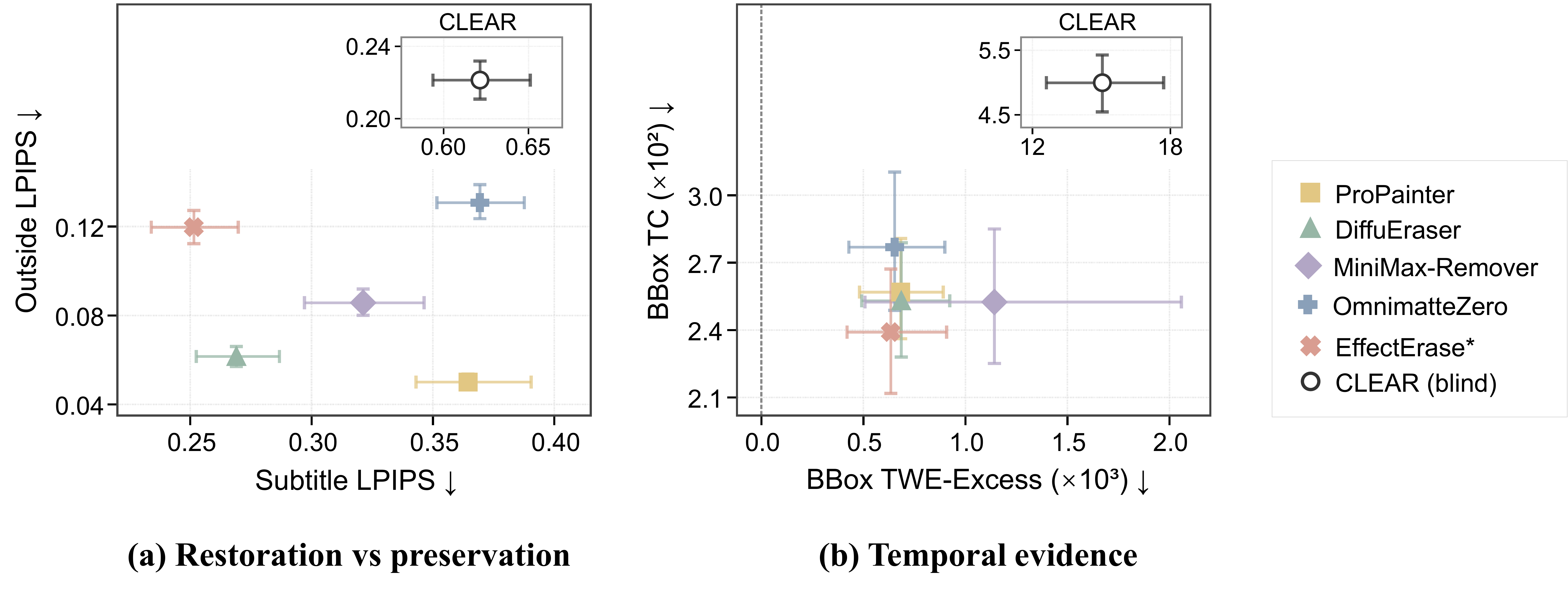}
    \vspace{-12pt}
    \caption{
    \textbf{Decoupled quality on Audit300.}
    (a) Subtitle versus Outside LPIPS.
    (b) BBox TWE-Excess versus BBox TC.
    Points show video-level macro means with video-bootstrap 95\% CIs;
    lower-left is better.
    CLEAR (blind) is plotted separately with independent axis ranges;
    cross-facet distances are not comparable.
    The dashed line marks zero TWE-Excess; negative values are retained.
    $^{*}$EffectErase is evaluated on its verified prefix.
    }
    \label{fig:finding1-quality-tradeoffs}
\end{figure*}
\noindent\textbf{Decoupled evaluation reveals quality trade-offs.}
Figure~\ref{fig:finding1-quality-tradeoffs} and
Table~\ref{tab:audit300-raw-results} show that Subtitle and Outside LPIPS
disagree on five of ten mask-guided method pairs.
For example, DiffuEraser has lower Subtitle LPIPS than ProPainter but
higher Outside LPIPS.
BBox TWE-Excess and BBox TC likewise disagree on six of ten pairs,
highlighting non-interchangeable temporal evidence.
Overall LPIPS follows Outside LPIPS more closely than Subtitle LPIPS.
These descriptive comparisons include EffectErase's verified-prefix
results and support reporting Restoration, Preservation, and Temporal
Fidelity separately rather than relying on full-frame scores alone.
See Appendix~\ref{app:finding1-details} for detailed comparisons.

\begin{figure*}[t]
\centering
\includegraphics[width=\textwidth]{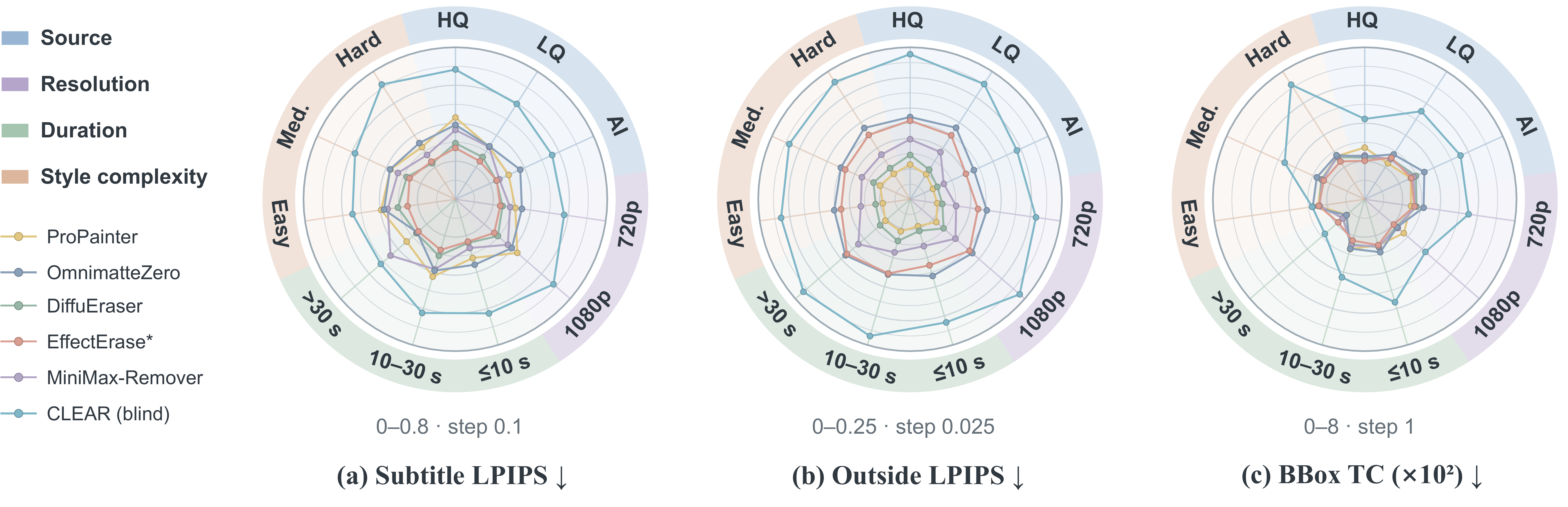}
\vspace{-17pt}
\caption{
\textbf{Performance across Audit300 dimensions.} (a) Subtitle LPIPS, (b) Outside LPIPS, and (c) BBox TC ($\times10^{2}$) represent Restoration, Preservation and Temporal fidelity. Points are video-level macro means and colored sectors indicate attribute domains. CLEAR is blind; $^{*}$EffectErase uses its verified prefix.
}
\label{fig:finding2-attributes}
\vspace{-10pt}
\end{figure*}

\noindent\textbf{Subtitle removal performance varies across dataset dimensions and capabilities.}
Figure~\ref{fig:finding2-attributes} shows that source, resolution, duration, and style complexity affect different capabilities in different ways. Real-HQ yields higher Subtitle and Outside LPIPS than AI-generated videos for all six methods, yet lower BBox TC for five, indicating that visually harder restoration does not necessarily imply greater temporal instability. Because source and resolution are coupled in Audit300, this contrast should be interpreted as a property of the joint data strata rather than an isolated resolution effect. Duration and style further reveal method-dependent shifts in error: all five mask-guided methods have higher Subtitle LPIPS on 10--30\,s clips, while Hard styles reduce their Subtitle LPIPS but increase Outside LPIPS; CLEAR instead worsens on both. Thus, increasing data complexity can redistribute errors across restoration and preservation rather than degrade all capabilities uniformly, motivating capability-wise evaluation over a single aggregate score. Further analyses are provided in Appendix~\ref{app:finding2-stratified-diagnostics}.

\Needspace{16\baselineskip}
\begin{wraptable}[11]{r}{0.45\columnwidth}
    \vspace{-\intextsep}
    \centering

    \caption{
    \textbf{Full-set delivery (\%).}
    Rates use all 3,000 videos; temporal coverage is averaged over evaluable outputs.
    }
    \label{tab:fullset-delivery-main}

    \scriptsize
    \setlength{\tabcolsep}{2.0pt}
    \renewcommand{\arraystretch}{1.05}

    \resizebox{\linewidth}{!}{%
    \begin{tabular}{@{}lrrrr@{}}
        \toprule[1pt]

        \textbf{Method}
        & \textbf{Comp.}
        & \textbf{Part.}
        & \textbf{Non-eval.}
        & \textbf{Time cov.} \\

        \midrule

        CLEAR
        & 5.6 & 94.4 & $<0.1$ & 86.2 \\

        ProPainter
        & \textbf{98.3} & 0.0 & 1.7 & \textbf{100.0} \\

        DiffuEraser
        & 72.4 & 26.6 & 1.0 & 99.8 \\

        MiniMax-Remover
        & 30.8 & 64.9 & 4.3 & 99.3 \\

        OmnimatteZero
        & 21.0 & 75.2 & 3.8 & 98.1 \\

        EffectErase
        & 3.9 & 93.7 & 2.4 & 47.1 \\

        \bottomrule[1pt]
    \end{tabular}%
    }
\end{wraptable}

\noindent\textbf{High evaluability masks different delivery regimes.}
Table~\ref{tab:fullset-delivery-main} shows sharply different delivery regimes across all 3,000 scheduled videos: complete rates range from 3.9\% to 98.3\%, although every method produces evaluable output for more
than 95\% of instances. High temporal coverage likewise does not imply complete-video delivery. Quality and delivery should therefore be reported as complementary properties rather than collapsed into a single successful-output score.Further analyses are provided in Appendix~\ref{app:finding3-delivery}.

\noindent\textbf{Evaluable scope and conditional quality require separate interpretation.}
Table~\ref{tab:finding4-native-set-counts} and
Figure~\ref{fig:finding4-native-tails} show that extending evaluation
from the common intersection to each method's native evaluable set changes
both the assessed scope and the observed quality distribution.
These changes are method- and metric-dependent: typical and tail errors can
shift differently, while spatial and temporal quality can even move in
opposite directions.
Thus, broader evaluability should not be interpreted as uniformly easier or
harder data, nor should conditional quality be used as a proxy for operating
scope.
Native evaluation instead complements shared-ID comparisons by revealing
performance on videos excluded from the common intersection, while comparisons
across different native sets do not constitute controlled cross-method
rankings.
Further analysis is provided in
Appendix~\ref{app:finding4-native-success}.

\noindent\textbf{Human judgments support spatial components, but not an SR-Q advantage.}
Table~\ref{tab:finding5-human-complete} shows substantially stronger human--metric agreement for Restoration and Preservation than for Temporal fidelity, where comparisons are also more frequently undecided.
This suggests that the spatial components capture perceptually meaningful differences more reliably, while temporal perceptual validation remains less conclusive.
SR-Q is comparable to the preregistered Overall-region baselines but shows no consistent advantage over them.
We therefore retain the three-dimensional raw profile as the primary diagnostic representation and use SR-Q only as a secondary summary, without post-hoc refitting.
Further details are provided in Appendix~\ref{app:finding5-human-qualitative}.

\section{Conclusion}


In this work, we presented \textbf{MFVSR}, a benchmark for systematic evaluation of video subtitle removal. MFVSR combines paired videos with controlled multilingual subtitle variations and an auxiliary collection of naturally subtitled videos. Its protocol distinguishes blind from mask-guided removal and separately evaluates restoration, preservation, temporal fidelity, and delivery coverage. Experiments on six representative methods reveal quality trade-offs and gaps between conditional performance and complete-video delivery. The held-out human study shows stronger agreement for spatial than temporal components, leaving temporal perceptual validation open. We hope that MFVSR and its analyses support reproducible comparison and the development of more reliable video subtitle removal methods.

\section*{AI Use Statement}
Generative AI tools assisted with code implementation and debugging, evaluation-protocol review, manuscript organization, and language editing. The authors reviewed the resulting text, verified the reported numerical results against the analysis outputs, and tested the evaluation code used in the study. The authors take responsibility for all claims, code, and artifacts in this submission.

\section*{Ethics Statement}

MFVSR uses videos from OpenVidHD, Panda-70M, Pexels,
VBench-2.0, and Rapidata, subtitle text from FLORES, and fonts from Google Fonts. These resources remain subject to their respective licenses and upstream restrictions. The auxiliary collection contains 953 naturally subtitled videos collected from Bilibili, Douyin, Xiaohongshu, Kuaishou, and YouTube using a web crawler. No additional privacy screening or de-identification was applied to these videos. This collection has no clean targets and is excluded from
paired quantitative evaluation. We plan to redistribute only assets for which redistribution is permitted, and otherwise provide permissible provenance information, annotations, and reconstruction instructions.

We recruited students from relevant academic disciplines for the human evaluation. Participants were informed of the research purpose and agreed to participate before the study. The study interface
used coded participant identifiers, and no additional
identifying information was retained with the study responses. Participants compared subtitle-removal outputs using a reference-based interface with concealed model identities. Human-evaluation results are reported only in aggregate in the manuscript.

\section*{Reproducibility Statement}
Sections~\ref{sec:benchmark-construction} and~\ref{sec:evaluation-protocol} describe the benchmark construction, model information boundaries, and evaluation design, while Appendix~\ref{app:evaluation-details} specifies spatial supports, metric regionalization, frozen frame and pair sampling, record reuse, aggregation, comparison sets, human evaluation, and fixed-budget validation. Across the six baselines, the retained inference logs cover a total of 10{,}137.4 GPU-hours (422.4 GPU-days). Appendix~\ref{app:computational_workload} details the logging coverage and method-specific timing boundaries. Subject to upstream licensing constraints, we plan to release the renderer, evaluation code, frozen identifiers, model configurations, analysis tables, and per-instance provenance records.

\bibliographystyle{iclr2027_conference}
\bibliography{iclr2027_conference}

\appendix
\definecolor{AppHeader}{RGB}{232,240,247}
\definecolor{AppBand}{RGB}{246,248,250}
\definecolor{AppInk}{RGB}{49,72,91}
\definecolor{AppMuted}{RGB}{90,100,109}
\renewcommand{\floatpagefraction}{0.85}
\setcounter{topnumber}{4}
\setcounter{bottomnumber}{3}
\setcounter{totalnumber}{6}
\setlength{\aboverulesep}{0pt}
\setlength{\belowrulesep}{0pt}
\setlength{\extrarowheight}{2pt}

\clearpage
\section*{Appendix Contents}

\begingroup
\small
\setlength{\parindent}{0pt}
\setlength{\parskip}{0pt}

\newcommand{\MFVSRTOCEntry}[3]{%
    \noindent\hspace*{#1}%
    \hyperref[#2]{%
        \makebox[2.7em][l]{\ref*{#2}}#3%
    }%
    \nobreak\dotfill
    \hyperref[#2]{\pageref*{#2}}%
    \par
}

\newcommand{\MFVSRTOCSection}[2]{%
    \par\addvspace{0.6em}%
    {\bfseries\MFVSRTOCEntry{0pt}{#1}{#2}}%
}

\newcommand{\MFVSRTOCSubsection}[2]{%
    \MFVSRTOCEntry{1.3em}{#1}{#2}%
}

\MFVSRTOCSection
    {sec:related-work}
    {Related Work}

\MFVSRTOCSection
    {app:evaluation-details}
    {Benchmark and Evaluation Details}
\MFVSRTOCSubsection
    {app:dataset-qa}
    {Dataset Validation}
\MFVSRTOCSubsection
    {app:supports-sampling}
    {Evaluation Supports, Sampling, and Aggregation}
\MFVSRTOCSubsection
    {app:metric-conventions}
    {Benchmark-Specific Metric Conventions}
\MFVSRTOCSubsection
    {app:human-protocol}
    {Human-Evaluation Protocol}
\MFVSRTOCSubsection
    {app:sr-q}
    {SR-Q: Definition and Validation Scope}

\MFVSRTOCSection
    {app:detailed-findings}
    {Supplementary Results}
\MFVSRTOCSubsection
    {app:audit300-complete-results}
    {Complete Audit300 Results}
\MFVSRTOCSubsection
    {app:finding1-details}
    {Decoupled Quality}
\MFVSRTOCSubsection
    {app:finding2-stratified-diagnostics}
    {Dataset and Subtitle Diagnostics}
\MFVSRTOCSubsection
    {app:finding3-delivery}
    {Full-Set Delivery and Coverage}
\MFVSRTOCSubsection
    {app:finding4-native-success}
    {Evaluable Scope and Conditional Quality}
\MFVSRTOCSubsection
    {app:finding5-human-qualitative}
    {Human Alignment and Qualitative Evidence}

\MFVSRTOCSection
    {app:computational_workload}
    {Computational Workload}

\MFVSRTOCSection
    {app:case-gallery}
    {Qualitative Cases and Diagnostic Insights}
\MFVSRTOCSubsection
    {app:cases-restoration}
    {Restoration: Text Removal and Scene Reconstruction}
\MFVSRTOCSubsection
    {app:cases-preservation}
    {Preservation: Changes beyond the Subtitle}
\MFVSRTOCSubsection
    {app:cases-temporal}
    {Temporal Fidelity: Changes Relative to the Target}

\endgroup
\medskip

\section{Related Work}
\label{sec:related-work}

\noindent\textbf{Video Subtitle Removal.}
Video subtitle removal combines localization, completion, and temporal
stabilization. The ChaLearn 2018 video-decaptioning task
~\citep{chalearn2018decaptioning} provided paired corrupted and clean clips,
while \citet{kim2019deep} introduced a blind encoder--decoder that
aggregates neighboring frames with recurrent feedback without requiring an
input mask. Recent diffusion-based methods revisit end-to-end mask-free
removal at higher resolution. CLEAR~\citep{he2026clear} separates
subtitle-prior extraction from generative refinement and performs mask-free
inference, whereas SEDiT~\citep{hui2026sedit} formulates subtitle erasure as
localized one-step video generation with chunk-wise streaming. These methods
strengthen the case for a dedicated task, but their paper-specific
400-sample evaluations do not provide a shared, fine-grained protocol for
diagnosing residual text, local reconstruction, distant modification, and
failure coverage.
The distinction between mask-free and mask-guided inference is methodological
rather than cosmetic. A blind method must infer both where and how to edit,
whereas a mask-guided method is given spatial side information. Combining both
settings in one ranking conflates localization and restoration. MFVSR
therefore uses the same paired targets and diagnostic views for both settings
but reports them in separate leaderboard blocks.

\noindent\textbf{Video Inpainting and Object Removal.}
Most video-inpainting methods assume a binary hole mask and focus on
reconstructing spatially plausible and temporally coherent content.
Flow-guided methods such as E2FGVI~\citep{li2022e2fgvi} propagate visible
evidence across frames, while later transformer- and diffusion-based
approaches expand temporal context and generative capacity.
ProPainter~\citep{zhou2023propainter} combines dual-domain propagation with a
mask-guided sparse video transformer, while
DiffuEraser~\citep{li2025diffueraser} incorporates diffusion priors for masked
video completion. Recent video-object-removal systems further target large
objects and associated visual effects. MiniMax-Remover~\citep{zi2026minimax}
reduces sampling cost through a distilled removal model;
OmnimatteZero~\citep{samuel2025omnimattezero} uses pretrained video diffusion
models to remove objects together with their footprints; and
EffectErase~\citep{fu2026effecterase} jointly learns removal and insertion to
better handle shadows, reflections, and deformation.
These methods are relevant baselines because the same canonical BBox-derived
subtitle annotation can be deterministically adapted to their released
interfaces. They are not subtitle localizers, however, and their mask-guided
results are not presented as evidence of blind subtitle-removal ability.

\noindent\textbf{Diagnostic Benchmarking and Perceptual Evaluation.}
Generic video-inpainting evaluations commonly repurpose
DAVIS~\citep{perazzi2016benchmark} or
YouTube-VOS~\citep{xu2018youtubevos} with synthetic masks, but aggregate
scores reveal little about the conditions under which a method fails.
DEVIL~\citep{szeto2022devil} addresses this limitation by labeling video and
mask attributes and evaluating controlled slices with reconstruction,
realism, and temporal-consistency criteria.
BVINet~\citep{wu2025bvinet} defines a blind video-inpainting setting and
evaluates on 508 YouTube-VOS and 60 DAVIS clips with synthesized corruptions.
These resources motivate controlled and diagnostic analysis, but their
corruptions are not designed around multilingual subtitle typography, motion,
animation, or language layout.

MFVSR also follows the broader move from one-number evaluation toward
dimension-specific evidence. VBench~\citep{huang2024vbench} decomposes
video-generation quality into fine-grained dimensions and validates them
against human preferences. For paired restoration, PSNR measures pixel-level
distortion, SSIM~\citep{wang2004ssim} measures local structural similarity,
LPIPS~\citep{zhang2018lpips} provides a learned perceptual distance, and
DISTS~\citep{ding2020dists} compares perceptual structure and texture.
Temporal warping error~\citep{lai2018temporal} measures motion-compensated
inconsistency. MFVSR retains the released metric definitions and weights where
possible, changing only the spatial support or controlled input used for
regional diagnosis. Because DISTS is an image-level score rather than a dense
map, its regional use is explicitly defined as \emph{conditioned DISTS} in
Appendix~\ref{app:evaluation-details}, rather than presented as an official
DISTS variant.

\section{Benchmark and Evaluation Details}
\label{app:evaluation-details}

\subsection{Dataset Validation}
\label{app:dataset-qa}
Following the screening of the base videos (Section~\ref{sec:base-video-filtering}), we validate decoding, target resolution, and frame alignment across videos, glyph masks, and annotations. A sample is rejected if any glyph pixel lies outside the rasterized BBox union. This check applies to the glyph core only; shadows, glow, outlines, and subtitle boxes may extend beyond it. Glyph masks must span the target timeline. Mask and prediction frame rates must match the target within 0.01 fps, using FFprobe’s rational frame-rate values. Prediction lengths are validated against the method’s temporal contract, and verified prefixes are recorded as partial outputs without padding.


\begin{table}[!htbp]
    \centering
    \caption{\textbf{Frozen inference configurations in MFVSR.}
    Mpx: per-frame pixel budget; AR: aspect ratio retained;
    $a$: dimension alignment; NN: nearest-neighbor resizing.
    All video masks are lossless.}
    \label{tab:inference-configurations}
    \begingroup
    \scriptsize
    \setlength{\tabcolsep}{3.5pt}
    \renewcommand{\arraystretch}{1.20}
    \setlength{\aboverulesep}{0pt}
    \setlength{\belowrulesep}{0pt}
    \setlength{\extrarowheight}{2pt}
    \definecolor{MFVSRConfigHeader}{RGB}{226,235,245}
    \definecolor{MFVSRConfigStripe}{RGB}{246,248,250}
    \begin{tabularx}{\linewidth}{
        >{\raggedright\arraybackslash}p{0.13\linewidth}
        >{\hsize=0.97\hsize\linewidth=\hsize\raggedright\arraybackslash}X
        >{\hsize=1.04\hsize\linewidth=\hsize\raggedright\arraybackslash}X
        >{\hsize=1.08\hsize\linewidth=\hsize\raggedright\arraybackslash}X
        >{\hsize=1.22\hsize\linewidth=\hsize\raggedright\arraybackslash}X
        >{\hsize=0.69\hsize\linewidth=\hsize\raggedright\arraybackslash}X
        }
        \toprule
        \rowcolor{MFVSRConfigHeader}
        \textbf{Method} & \textbf{Mask adapter} & \textbf{Condition}
        & \textbf{Spatial} & \textbf{Temporal / sampling}
        & \textbf{Numerics} \\
        \midrule

        \textbf{CLEAR}
        & None
        & Fixed removal prompt
        & $0.92$ Mpx; $a=16$\newline
          Scale-to-cover; center crop
        & Window 81; overlap 16\newline
          Denoising steps: 5; CFG 1.0
        & Precision logged; seed unset; tiled VAE \\

        \rowcolor{MFVSRConfigStripe}
        \textbf{ProPainter}
        & PNG sequence; NN; dilation 4
        & None
        & $2.07$ Mpx\newline AR; $a=8$
        & RAFT: 20 iter.; subvideo 80\newline
          Neighbor 10; ref.\ stride 10
        & FP16 \\

        \textbf{DiffuEraser}
        & Video; binary; dilation 8
        & ProPainter prior from $\mathbf{X},\mathbf{Q}$
        & $2.07$ Mpx; $a=8$\newline Long side $\leq1920$
        & Prior: subvideo 50, neighbor 10, ref.\ stride 10\newline
          2-Step checkpoint
        & FP16 \\

        \rowcolor{MFVSRConfigStripe}
        \textbf{MiniMax-\newline Remover}
        & Video; threshold $>20$; NN
        & No text
        & $2.07$ Mpx\newline AR; $a=16$
        & Full supplied clip\newline
          Sampling: 12 steps, 6 iter.
        & FP16; seed unset \\

        \textbf{OmnimatteZero}
        & Video; threshold 128
        & Fixed \emph{Empty} + negative prompts
        & $2.07$ Mpx\newline AR; $a=32$
        & Full supplied clip\newline Denoising steps: 30
        & BF16; seed 1; tiled VAE \\

        \rowcolor{MFVSRConfigStripe}
        \textbf{EffectErase}
        & Video; bilinear resizing
        & Fixed prompts; $224{\times}224$ crop\textsuperscript{*}
        & $2.07$ Mpx\newline AR; $a=16$
        & $\leq81$ frames; output $4n+1$\newline
          Denoising steps: 50; CFG 1.0
        & BF16; seed 2025; tiled inference \\
        \bottomrule
    \end{tabularx}
    \vspace{2pt}
    \begin{minipage}{\linewidth}
        \scriptsize
        Temporal window, overlap, subvideo, and neighbor lengths are in frames.
        \textsuperscript{*}Crop from the first active mask frame.
    \end{minipage}
    \endgroup
\end{table}

\subsection{Evaluation Supports, Sampling, and Aggregation}
\label{app:supports-sampling}

\noindent\textbf{Spatial support and correspondence.}
For frame $t$, let $G_t\in\{0,1\}^{H\times W}$ be the thresholded glyph
mask and let $B_t$ be the union of all rasterized subtitle bounding boxes.
MFVSR defines three atomic supports:
\begin{align}
    R_t^{\mathrm{Sub}} &= G_t, \\
    R_t^{\mathrm{BBox\text{-}NG}} &= B_t\wedge \neg G_t, \\
    R_t^{\mathrm{Outside}} &= \neg B_t.
\end{align}
The evaluator requires $G_t\subseteq B_t$ and aborts a sample if any glyph
pixel lies outside its BBox under the strict configuration. The three atomic
supports are therefore mutually exclusive and exhaustive. The aggregate
supports are
\begin{equation}
    R_t^{\mathrm{BBox}}
    =
    R_t^{\mathrm{Sub}}
    \cup
    R_t^{\mathrm{BBox\text{-}NG}},
    \qquad
    R_t^{\mathrm{Overall}}=\Omega.
\end{equation}
These definitions induce the five views used throughout MFVSR: Overall,
BBox, Subtitle, BBox-NG, and Outside.

A regional measurement uses $R_t\cap V_t$, with coverage
$|R_t\cap V_t|/|R_t|$, and empty supports remain missing rather than an artificial zero. Notably, BBox-NG can contain rendered effects and is not a pure background region. 

In addition, resolution changes are inverted using the frozen bicubic transform, whereas CLEAR’s recorded scale-to-cover and center-crop operations are inverted only where correspondence is valid. Exact-length contracts require all target frames, while verified-prefix contracts allow shorter outputs with temporal coverage $T_m/T$, without padding or duplication. Frame correspondence is determined using verified indices.

\noindent\textbf{Frame budget.}
Lite3000 selects at most 48 spatial frames: 32 evenly spaced subtitle-active
frames and 16 full-timeline frames. Its 32 adjacent pairs comprise 24 touching
an active frame and eight spanning the timeline. Selection covers endpoints,
removes duplicates, and fills unused allocation from the full timeline.
Activity means a nonempty benchmark BBox list. Selected indices are stored
and depend only on annotations and video length. Audit300 reuses this support,
adding SSIM, conditioned DISTS, and TWE-Excess to Lite3000's PSNR, LPIPS, and TC.

\noindent\textbf{Audit300 selection.}
\label{app:audit300-selection}
For Audit300 selection, we started from the common-evaluable pool across the evaluated models, then removing held-out human-study videos and box-style videos, yielding the candidate set. A fixed-seed, metadata-only optimizer finally selected 300 videos to reduce marginal deficits across source, resolution, duration,
style complexity, and the four subtitle factors, plus Source$\times$Complexity
and Resolution$\times$Duration. Metrics, method identities, ranks, and human
judgments were not used. Eligibility required current records and positive verified support.
Table~\ref{tab:finding2-attribute-slices} reports the selected composition, and Figure~\ref{fig:finding2-subtitle-factors} gives the subtitle-factor group
sizes and contrasts.

\begin{table}[!htbp]
\centering\small
\caption{\textbf{Audit300 composition. }Resolution labels denote evaluation classes.}
\label{tab:finding2-attribute-slices}
\begin{minipage}{0.72\linewidth}
\renewcommand{\arraystretch}{1.14}
\begin{tabularx}{\linewidth}{lY}
\toprule
\rowcolor{AppHeader}
\textbf{Attribute} & \textbf{Groups (videos)} \\
\midrule
Source & Real-HQ (123), Real-LQ (88), AI-generated (89) \\
Resolution & 720p-class (177), 1080p-class (123) \\
Duration & $\leq10$ s (168), 10--30 s (122), $>30$ s (10) \\
Style complexity & Easy (93), Medium (97), Hard (110) \\
\bottomrule
\end{tabularx}
\end{minipage}
\end{table}

\noindent\textbf{Style complexity.}
The frozen metadata score is
$d=2\mathbb{1}[\mathrm{Dynamic}]+\mathbb{1}[\mathrm{Animated}]
+s_{\mathrm{visual}}+\mathbb{1}[\mathrm{Bilingual}]$, where
$s_{\mathrm{visual}}=0,1,2$ for Plain, Decorated, and Stylized.
Easy, Medium, and Hard denote $d\leq1$, $2\leq d\leq3$, and $d\geq4$.
These are rendering-complexity groups, not an empirically calibrated
difficulty scale or a claim about language-specific difficulty.

\noindent\textbf{Aggregation and comparison sets.}
We pool pixel-additive statistics within each video and average perceptual
and temporal values over valid sampled frames or pairs. Dataset results are
video-level macro means with fixed-seed video-bootstrap intervals.
Reliability uses the scheduled set $\mathcal D$, while conditional quality
uses each method's evaluable set $\mathcal E_m$. The common intersection
$\mathcal D_{\cap}$ and Audit300 provide shared-ID comparisons, with symmetric
validity checked separately for each metric. Shared IDs do not guarantee equal
coverage. We report blind and guided methods separately and leave quality
undefined for failed outputs.

\subsection{Benchmark-Specific Metric Conventions}
\label{app:metric-conventions}
PSNR, SSIM~\citep{wang2004ssim}, LPIPS~\citep{zhang2018lpips}, and
DISTS~\citep{ding2020dists} use their standard definitions. We specify only
the regionalization and compatibility choices needed to reproduce MFVSR.
Table~\ref{tab:metric-view-matrix} distinguishes primary measurements
from diagnostic and compatibility views.

\begin{table}[!htbp]
\centering\footnotesize
\caption{\textbf{Metric--view matrix. }Lite measurements cover Lite3000; Audit adds
measurements on Audit300. Bold entries define the fidelity components.
Overall is retained for compatibility; BBox-NG is a mixed local diagnostic.}
\label{tab:metric-view-matrix}
\setlength{\tabcolsep}{3pt}
\renewcommand{\arraystretch}{1.15}
\begin{tabularx}{\linewidth}{l*{5}{>{\centering\arraybackslash}X}}
\toprule
\rowcolor{AppHeader}
\textbf{Metric / tier} & \textbf{Overall} & \textbf{BBox} & \textbf{Subtitle} & \textbf{BBox-NG} & \textbf{Outside} \\
\midrule
PSNR / Lite & Compat. & Diag. & \textbf{Primary} & Diag. & \textbf{Primary} \\
SSIM / Audit & Compat. & Diag. & Diag. & Diag. & Diag. \\
LPIPS / Lite & Compat. & Diag. & \textbf{Primary} & Diag. & \textbf{Primary} \\
DISTS / Audit & Compat. & -- & \textbf{cDISTS} & cDISTS & -- \\
\makecell[l]{TWE / TWE-Excess\\Audit} & -- & \textbf{Primary} & -- & -- & \textbf{Primary} \\
TC / Lite & -- & \textbf{Primary} & -- & -- & \textbf{Primary} \\
\bottomrule
\end{tabularx}
\end{table}

\noindent\textbf{Regional spatial metrics.}
PSNR pools RGB squared error before conversion (range 255); exact matches
retain infinite PSNR. Regional SSIM averages the standard map after excluding
a five-pixel image boundary and is diagnostic on sparse glyph masks.
Regional LPIPS uses AlexNet v0.1 and
fractional mask weights obtained by area-resampling the support to the released
spatial map. Overall SSIM and LPIPS use their standard scalars.

\noindent\textbf{Conditioned DISTS.}
DISTS is an image-level metric. For Subtitle or BBox-NG support $R_t$, we
replace only that region in the target with the prediction:
\begin{equation}
\widetilde Y_t^{(R)}=R_t\odot\widehat Y_t+(1-R_t)\odot Y_t,
\qquad
\mathrm{cDISTS}_t^R=\mathrm{DISTS}
\bigl(\widetilde Y_t^{(R)}|_{C_t},Y_t|_{C_t}\bigr).
\end{equation}
The shared BBox-context crop $C_t$ adds a 16-pixel margin and expands to a
minimum side of 64 pixels where boundaries permit. This controlled-input
diagnostic is not an official dense DISTS variant. Overall DISTS requires
complete spatial correspondence and area-downsamples long sides above 1920
pixels to 1920; local crops are not subject to that rule. All variants use the
released weights and $[0,1]$ RGB inputs. Non-finite values remain missing
if a native-backend retry fails; backend, resize, and validity are logged.

\noindent\textbf{Temporal diagnostics.}
Both diagnostics use BBox and Outside. RAFT-Large~\citep{teed2020raft}
estimates flow only from consecutive clean targets. The flow is restored bilinearly with component scaling. Forward--backward consistency rejects occlusions and inconsistent coordinates. On the resulting shared
support, TWE is the channel-mean squared residual between a frame and its
backward-warped predecessor. We retain the signed target-baseline difference
\begin{equation}
\mathrm{TWE\text{-}Excess}^r
=\mathrm{TWE}_{\mathrm{pred}}^r-\mathrm{TWE}_{\mathrm{gt}}^r.
\end{equation}
Negative values do not establish superior fidelity or isolate model-induced
error. The complementary flow-free diagnostic is
\begin{equation}
\mathrm{TC}_t^r=
\frac{1}{3|Q_t^r|}\sum_{p\in Q_t^r}
\left\| (\widehat Y_t-\widehat Y_{t-1})_p-(Y_t-Y_{t-1})_p\right\|_1,
\end{equation}
where $Q_t^r$ is the same-coordinate intersection of consecutive valid
regional supports and RGB values lie in $[0,1]$. TC penalizes suppression of
legitimate target changes. Empty temporal support remains undefined.

\subsection{Human-Evaluation Protocol}
\label{app:human-protocol}
The held-out study contains 100 distinct videos covering all 15 method pairs,
disjoint from Audit300 and the pilot. Selection uses availability, verified
support, pair coverage, and frozen metadata, without metrics or anticipated
quality. Each comparison uses the shorter verified prefix and intersected
valid spatial rectangles; all displayed subtitle BBoxes must lie inside that
support. Thus verified partial outputs remain eligible.

\begin{figure*}[!htbp]
    \centering
    
    \begin{minipage}[t]{0.49\textwidth}
        \centering
        \includegraphics[width=\linewidth]
            {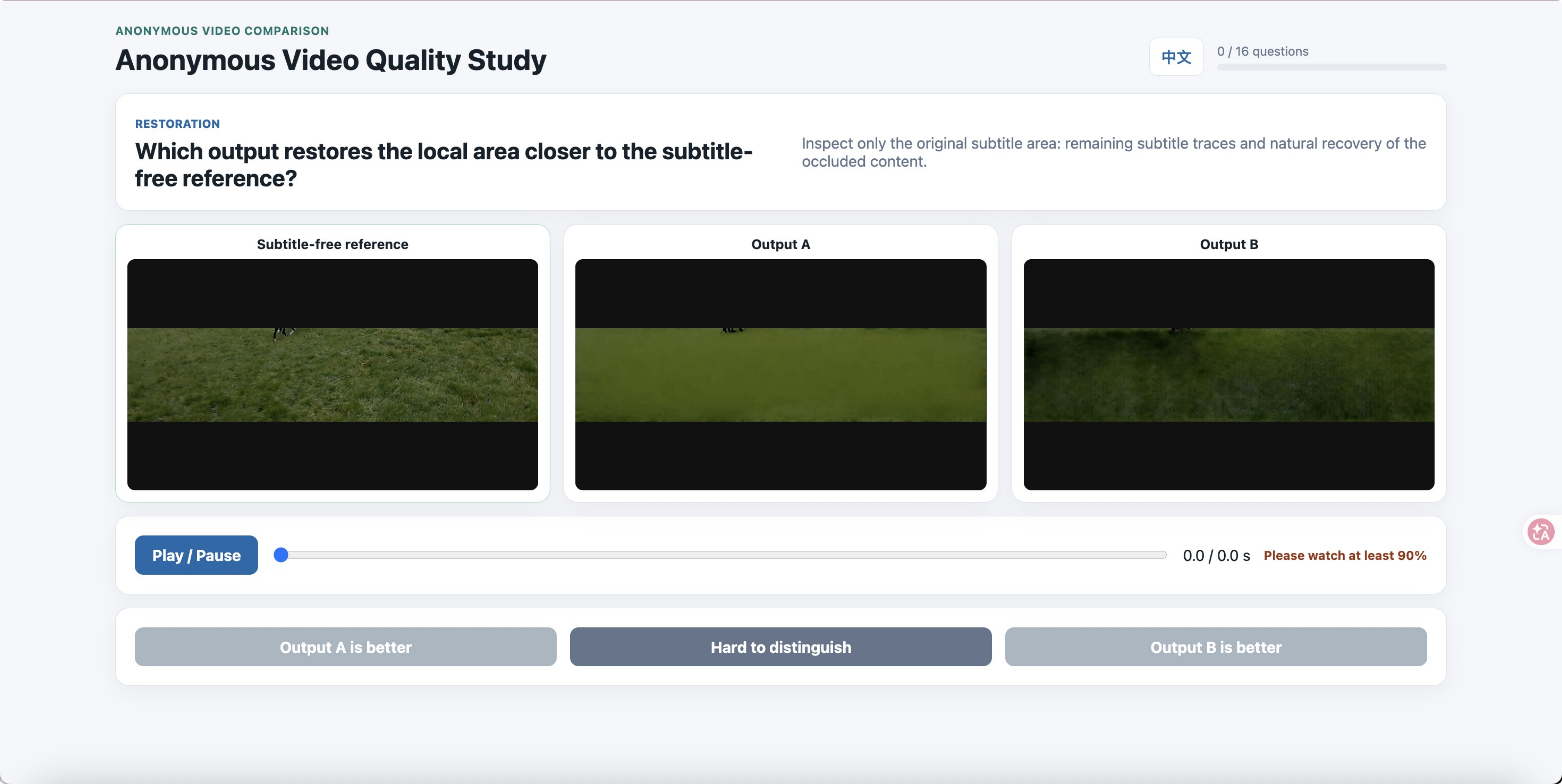}
        \vspace{1mm}
        \centerline{\small (a) Restoration Fidelity}
    \end{minipage}
    \hfill
    \begin{minipage}[t]{0.49\textwidth}
        \centering
        \includegraphics[width=\linewidth]
            {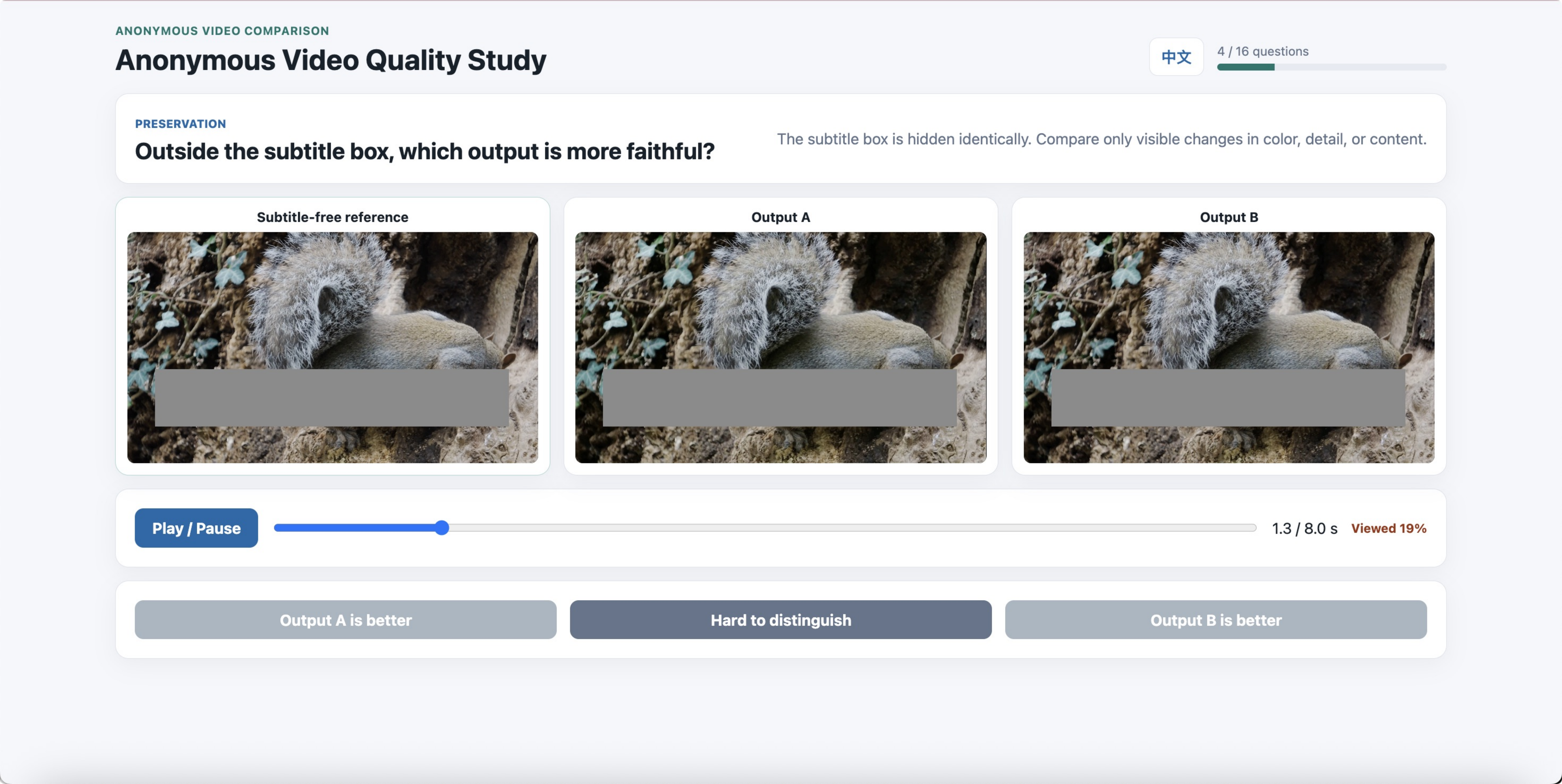}
        \vspace{1mm}
        \centerline{\small (b) Preservation Fidelity}
    \end{minipage}

    \vspace{2mm}

    \begin{minipage}[t]{0.49\textwidth}
        \centering
        \includegraphics[width=\linewidth]
            {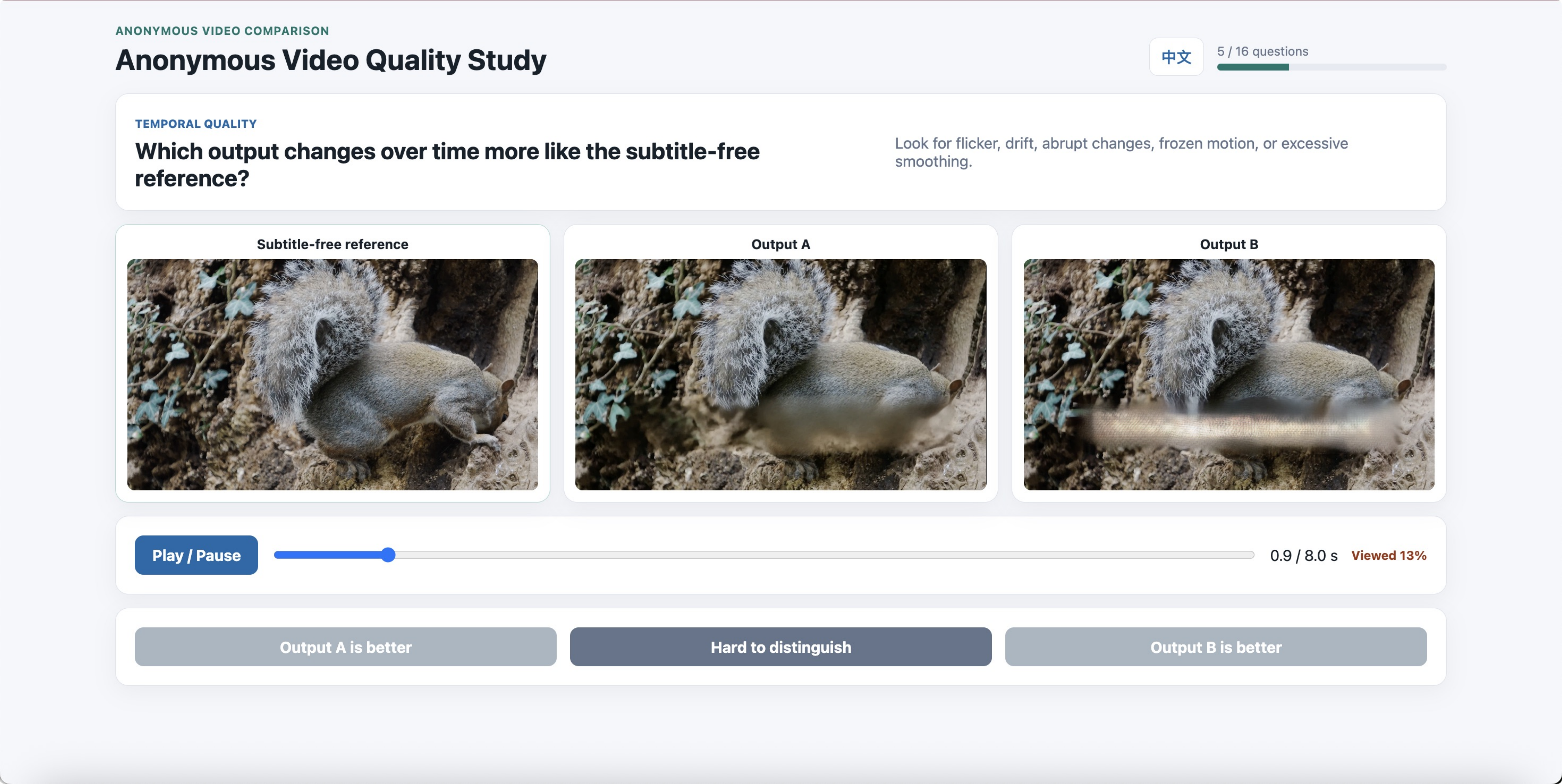}
        \vspace{1mm}
        \centerline{\small (c) Temporal Fidelity}
    \end{minipage}
    \hfill
    \begin{minipage}[t]{0.49\textwidth}
        \centering
        \includegraphics[width=\linewidth]
            {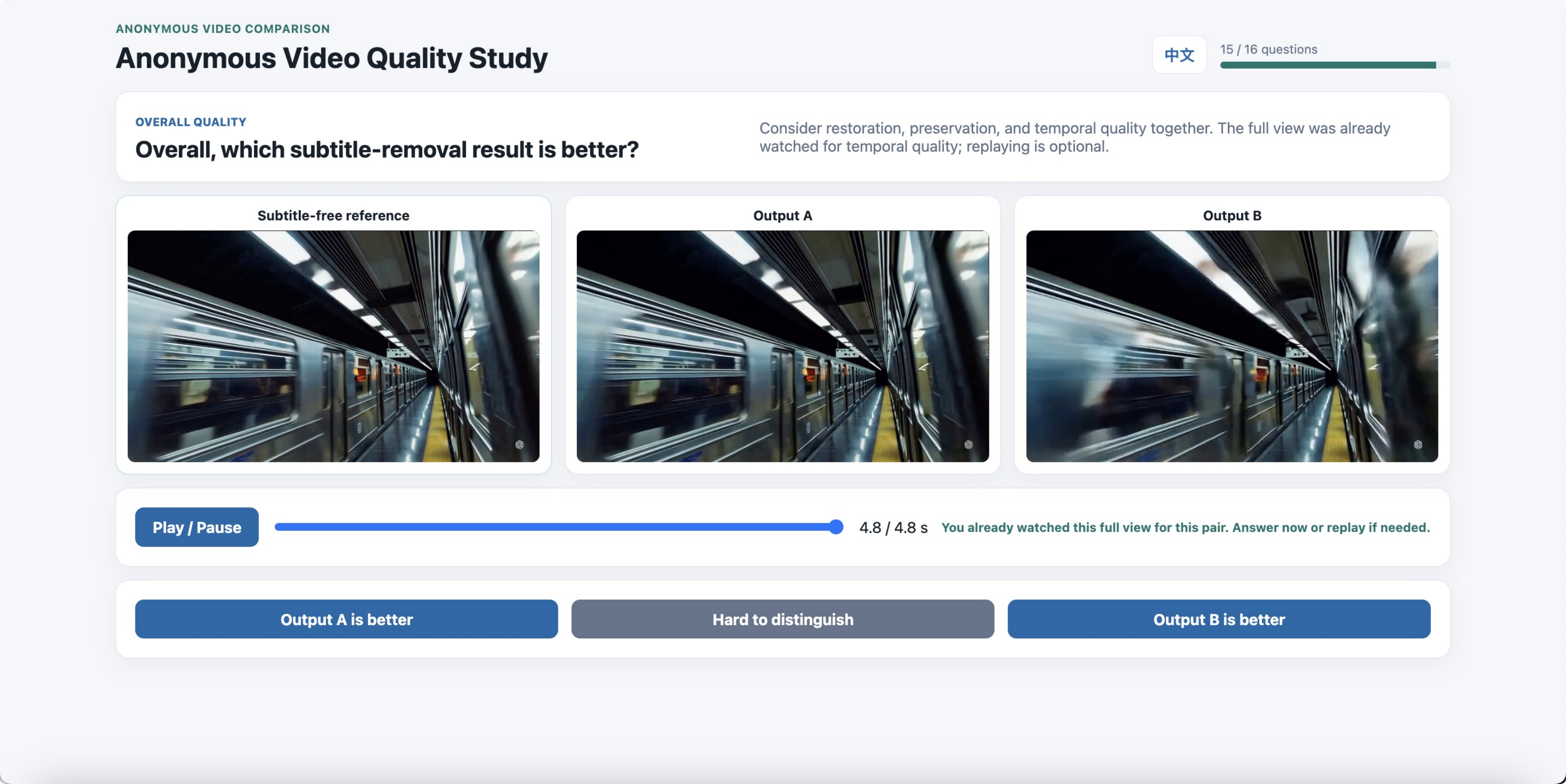}
        \vspace{1mm}
        \centerline{\small (d) Overall preference}
    \end{minipage}
    \vspace{-5pt}
    \caption{
    \textbf{Human-evaluation interface for held-out protocol validation.}
    }
    \label{fig:human-evaluation-interface}
\end{figure*}

Figure~\ref{fig:human-evaluation-interface} shows the four-task interface.
Two anonymized predictions accompany a synchronized clean reference.
Left--right assignment is randomized and held fixed across the four tasks
for each annotator--pair. Restoration assesses residual text, surrounding
effects, and reconstruction. Preservation masks the operational BBox
identically in all three views to isolate Outside content. Temporal Fidelity
assesses coherence relative to target dynamics, and Overall asks for an
integrated preference on the same assets. Each task offers left, right, or
\emph{hard to distinguish}; responses unlock after 90\% of every required
view has been watched.

After preregistered quality control, 16 of 19 annotators contribute 944 valid
judgments. For each task, 64 pairs have two ratings and 36 have three.
A unique plurality determines the pair label, and exact ties remain undecided.
All analyses treat the video pair as the independent unit. We report
variable-item-size Fleiss' agreement, tie rates, accuracy on decisive pairs
with Wilson intervals, and Kendall's $\tau_b$.

Each component is compared with its corresponding human task. SR-Q and the
four Overall metrics are compared on the same decisive Overall pairs, using
paired-bootstrap accuracy differences and exact McNemar tests with Holm
correction. Table~\ref{tab:finding5-human-complete} includes every
preregistered candidate.

\subsection{SR-Q: Definition and Validation Scope}
\label{app:sr-q}
SR-Q was preregistered as a scalar summary of the fidelity profile. It did
not improve on conventional Overall metrics in the held-out study, so we
exclude it from the primary leaderboard and stratified analysis. Its
definition is retained here to make that comparison reproducible.

For each metric $k$, P5/P95 anchors $(a_k,b_k)$ are frozen from the six-model
Audit300 grid. We compute $z_k=\mathrm{clip}((x_k-a_k)/(b_k-a_k),0,1)$ and
set $q_k=z_k$ for higher-is-better metrics and $q_k=1-z_k$ otherwise. Only this normalization
clips TWE-Excess at zero. Anchors use no human labels and are not refit for new
methods. The components and candidate are
\begin{align}
S_{\mathrm{rest}}&=\operatorname{mean}
(q_{\mathrm{Sub,PSNR}},q_{\mathrm{Sub,LPIPS}},q_{\mathrm{Sub,cDISTS}}),\\
S_{\mathrm{pres}}&=\operatorname{mean}(q_{\mathrm{Out,PSNR}},q_{\mathrm{Out,LPIPS}}),\\
S_{\mathrm{temp},r}&=\operatorname{mean}(q_{r,\mathrm{TWE\text{-}Excess}},q_{r,\mathrm{TC}}),
\quad r\in\{\mathrm{BBox},\mathrm{Out}\},\\
S_{\mathrm{temp}}&=\sqrt{S_{\mathrm{temp,BBox}}S_{\mathrm{temp,Out}}},\qquad
\mathrm{SR\text{-}Q}=100(S_{\mathrm{rest}}S_{\mathrm{pres}}S_{\mathrm{temp}})^{1/3}.
\end{align}
Audit results require common valid support. A metric may be dropped
symmetrically only if structurally unsupported for every method on that video;
method-specific missing support invalidates the comparison. No component is
imputed. Human-study videos remain disjoint from anchor estimation. SR-Q is
not extrapolated to failed or unaudited outputs, multiplied by coverage, or
combined with computational cost.

\section{Supplementary Results}
\label{app:detailed-findings}

\subsection{Complete Audit300 Results}
\label{app:audit300-complete-results}
Tables~\ref{tab:audit300-spatial-ci} and~\ref{tab:audit300-temporal-ci}
add video-bootstrap 95\% intervals and Subtitle SSIM to the main results.
All entries use 300 videos except BBox TWE-Excess, whose symmetric support
contains 299. Values remain conditional on each method's verified output.

\begin{table}[!htbp]
\centering\footnotesize
\caption{\textbf{Audit300 spatial results:} mean above video-bootstrap 95\% CI. PSNR is in dB; LPIPS and cDISTS are lower-is-better. CLEAR is blind; other methods are mask-guided.}
\label{tab:audit300-spatial-ci}
\setlength{\tabcolsep}{3.5pt}
\renewcommand{\arraystretch}{1.08}
\begin{tabularx}{\linewidth}{l|*{4}{>{\centering\arraybackslash}X}|*{2}{>{\centering\arraybackslash}X}}
\toprule
\rowcolor{AppHeader}
\textbf{Method} & \multicolumn{4}{c|}{\textbf{Subtitle}} & \multicolumn{2}{c}{\textbf{Outside}} \\
\rowcolor{AppHeader}
& PSNR & SSIM & LPIPS & cDISTS & PSNR & LPIPS \\
\midrule
CLEAR & \makecell{11.99\\[-1pt]{\scriptsize\color{AppMuted} [11.31, 12.75]}} & \makecell{0.356\\[-1pt]{\scriptsize\color{AppMuted} [0.322, 0.391]}} & \makecell{0.621\\[-1pt]{\scriptsize\color{AppMuted} [0.594, 0.651]}} & \makecell{0.236\\[-1pt]{\scriptsize\color{AppMuted} [0.225, 0.248]}} & \makecell{21.48\\[-1pt]{\scriptsize\color{AppMuted} [21.05, 21.90]}} & \makecell{0.221\\[-1pt]{\scriptsize\color{AppMuted} [0.211, 0.232]}} \\
\midrule
\rowcolor{AppBand}ProPainter & \makecell{19.66\\[-1pt]{\scriptsize\color{AppMuted} [18.94, 20.41]}} & \makecell{0.714\\[-1pt]{\scriptsize\color{AppMuted} [0.694, 0.734]}} & \makecell{0.364\\[-1pt]{\scriptsize\color{AppMuted} [0.343, 0.390]}} & \makecell{0.138\\[-1pt]{\scriptsize\color{AppMuted} [0.130, 0.148]}} & \makecell{34.11\\[-1pt]{\scriptsize\color{AppMuted} [33.78, 34.44]}} & \makecell{0.050\\[-1pt]{\scriptsize\color{AppMuted} [0.047, 0.054]}} \\
DiffuEraser & \makecell{21.36\\[-1pt]{\scriptsize\color{AppMuted} [20.74, 21.94]}} & \makecell{0.746\\[-1pt]{\scriptsize\color{AppMuted} [0.726, 0.766]}} & \makecell{0.269\\[-1pt]{\scriptsize\color{AppMuted} [0.253, 0.287]}} & \makecell{0.116\\[-1pt]{\scriptsize\color{AppMuted} [0.110, 0.124]}} & \makecell{32.87\\[-1pt]{\scriptsize\color{AppMuted} [32.59, 33.16]}} & \makecell{0.062\\[-1pt]{\scriptsize\color{AppMuted} [0.057, 0.066]}} \\
\makecell[l]{MiniMax-\\Remover} & \makecell{20.32\\[-1pt]{\scriptsize\color{AppMuted} [19.61, 20.96]}} & \makecell{0.725\\[-1pt]{\scriptsize\color{AppMuted} [0.700, 0.748]}} & \makecell{0.321\\[-1pt]{\scriptsize\color{AppMuted} [0.297, 0.346]}} & \makecell{0.132\\[-1pt]{\scriptsize\color{AppMuted} [0.123, 0.141]}} & \makecell{30.73\\[-1pt]{\scriptsize\color{AppMuted} [30.37, 31.12]}} & \makecell{0.086\\[-1pt]{\scriptsize\color{AppMuted} [0.080, 0.092]}} \\
OmnimatteZero & \makecell{17.69\\[-1pt]{\scriptsize\color{AppMuted} [17.09, 18.31]}} & \makecell{0.651\\[-1pt]{\scriptsize\color{AppMuted} [0.628, 0.673]}} & \makecell{0.369\\[-1pt]{\scriptsize\color{AppMuted} [0.352, 0.388]}} & \makecell{0.160\\[-1pt]{\scriptsize\color{AppMuted} [0.151, 0.168]}} & \makecell{26.00\\[-1pt]{\scriptsize\color{AppMuted} [25.53, 26.47]}} & \makecell{0.131\\[-1pt]{\scriptsize\color{AppMuted} [0.124, 0.139]}} \\
\rowcolor{AppBand}EffectErase & \makecell{20.58\\[-1pt]{\scriptsize\color{AppMuted} [20.00, 21.17]}} & \makecell{0.727\\[-1pt]{\scriptsize\color{AppMuted} [0.705, 0.749]}} & \makecell{0.251\\[-1pt]{\scriptsize\color{AppMuted} [0.234, 0.270]}} & \makecell{0.138\\[-1pt]{\scriptsize\color{AppMuted} [0.130, 0.146]}} & \makecell{25.77\\[-1pt]{\scriptsize\color{AppMuted} [25.26, 26.27]}} & \makecell{0.120\\[-1pt]{\scriptsize\color{AppMuted} [0.112, 0.127]}} \\
\bottomrule
\end{tabularx}
\end{table}

\begin{table}[!htbp]
\centering\footnotesize
\caption{\textbf{Audit300 temporal results:} mean above video-bootstrap 95\% CI. TC is scaled by $10^2$ and TWE-X by $10^3$. CLEAR is blind; other methods are mask-guided.}
\label{tab:audit300-temporal-ci}
\setlength{\tabcolsep}{3.5pt}
\renewcommand{\arraystretch}{1.08}
\begin{tabularx}{\linewidth}{l|*{2}{>{\centering\arraybackslash}X}|*{2}{>{\centering\arraybackslash}X}|>{\centering\arraybackslash}X}
\toprule
\rowcolor{AppHeader}
\textbf{Method} & \multicolumn{2}{c|}{\textbf{BBox}} & \multicolumn{2}{c|}{\textbf{Outside}} & \textbf{Joint} \\
\rowcolor{AppHeader}
& TC & TWE-X & TC & TWE-X & \textbf{coverage (\%)} \\
\midrule
CLEAR & \makecell{5.00\\[-1pt]{\scriptsize\color{AppMuted} [4.54, 5.43]}} & \makecell{15.042\\[-1pt]{\scriptsize\color{AppMuted} [12.605, 17.707]}} & \makecell{1.77\\[-1pt]{\scriptsize\color{AppMuted} [1.59, 1.98]}} & \makecell{0.076\\[-1pt]{\scriptsize\color{AppMuted} [$-0.013$, 0.175]}} & \makecell{85.4\\[-1pt]{\scriptsize\color{AppMuted} [83.9, 86.8]}} \\
\midrule
\rowcolor{AppBand}ProPainter & \makecell{2.57\\[-1pt]{\scriptsize\color{AppMuted} [2.36, 2.81]}} & \makecell{0.683\\[-1pt]{\scriptsize\color{AppMuted} [0.481, 0.891]}} & \makecell{0.70\\[-1pt]{\scriptsize\color{AppMuted} [0.65, 0.75]}} & \makecell{0.072\\[-1pt]{\scriptsize\color{AppMuted} [0.017, 0.160]}} & \makecell{100.0\\[-1pt]{\scriptsize\color{AppMuted} [100.0, 100.0]}} \\
DiffuEraser & \makecell{2.53\\[-1pt]{\scriptsize\color{AppMuted} [2.28, 2.79]}} & \makecell{0.686\\[-1pt]{\scriptsize\color{AppMuted} [0.491, 0.923]}} & \makecell{0.73\\[-1pt]{\scriptsize\color{AppMuted} [0.68, 0.79]}} & \makecell{0.068\\[-1pt]{\scriptsize\color{AppMuted} [0.012, 0.152]}} & \makecell{99.9\\[-1pt]{\scriptsize\color{AppMuted} [99.8, 99.9]}} \\
\makecell[l]{MiniMax-\\Remover} & \makecell{2.52\\[-1pt]{\scriptsize\color{AppMuted} [2.25, 2.85]}} & \makecell{1.142\\[-1pt]{\scriptsize\color{AppMuted} [0.508, 2.058]}} & \makecell{1.12\\[-1pt]{\scriptsize\color{AppMuted} [1.01, 1.24]}} & \makecell{0.028\\[-1pt]{\scriptsize\color{AppMuted} [$-0.041$, 0.117]}} & \makecell{99.4\\[-1pt]{\scriptsize\color{AppMuted} [99.3, 99.5]}} \\
OmnimatteZero & \makecell{2.77\\[-1pt]{\scriptsize\color{AppMuted} [2.49, 3.10]}} & \makecell{0.653\\[-1pt]{\scriptsize\color{AppMuted} [0.429, 0.899]}} & \makecell{1.51\\[-1pt]{\scriptsize\color{AppMuted} [1.34, 1.70]}} & \makecell{0.048\\[-1pt]{\scriptsize\color{AppMuted} [$-0.032$, 0.119]}} & \makecell{98.5\\[-1pt]{\scriptsize\color{AppMuted} [98.3, 98.7]}} \\
\rowcolor{AppBand}EffectErase & \makecell{2.39\\[-1pt]{\scriptsize\color{AppMuted} [2.12, 2.67]}} & \makecell{0.634\\[-1pt]{\scriptsize\color{AppMuted} [0.420, 0.908]}} & \makecell{1.51\\[-1pt]{\scriptsize\color{AppMuted} [1.36, 1.67]}} & \makecell{0.129\\[-1pt]{\scriptsize\color{AppMuted} [0.062, 0.198]}} & \makecell{43.1\\[-1pt]{\scriptsize\color{AppMuted} [40.4, 45.5]}} \\
\bottomrule
\end{tabularx}
\end{table}

\subsection{Decoupled Quality}
\label{app:finding1-details}

\noindent\textbf{Spatial trade-offs.}
Among guided methods, EffectErase has the lowest Subtitle LPIPS, whereas
ProPainter has the lowest Outside LPIPS. These different strengths lead to
substantial ranking reversals (Table~\ref{tab:finding1-rank-concordance}).
Overall LPIPS tracks Outside LPIPS more closely than Subtitle LPIPS, so a
full-frame score can hide errors in the repaired region.

\noindent\textbf{Sensitivity to partial outputs.}
Removing EffectErase still leaves spatial and temporal ranking reversals
(2/6 and 6/6 pairs), so its short outputs cannot fully explain the disagreement.
This check is descriptive because it does not align temporal support.
The temporal metrics also capture different behavior: low TWE-Excess can
coexist with a mismatch in target changes measured by TC. Reporting both
preserves a distinction that a single score would obscure.

\begin{table}[!htbp]
    \centering
    \caption{\textbf{Pairwise rank agreement across mask-guided methods.} Metric-induced orderings are compared over five methods (10 pairs), including prefix-conditioned EffectErase; reversals indicate opposite
pairwise preferences}
    \label{tab:finding1-rank-concordance}
    \footnotesize
    \setlength{\tabcolsep}{4.5pt}
    \renewcommand{\arraystretch}{1.14}
    \begin{minipage}{0.7\linewidth}
\begin{tabularx}{\linewidth}{Ycc}
        \toprule
\rowcolor{AppHeader}
        \textbf{Metric relation} & \textbf{Reversed pairs}
        & \textbf{Kendall $\tau_b$} \\
        \midrule
        Subtitle LPIPS vs.\ Outside LPIPS & 5/10 & 0.00 \\
        BBox TWE-Excess vs.\ BBox TC & 6/10 & $-0.20$ \\
        Overall LPIPS vs.\ Subtitle LPIPS & 4/10 & 0.20 \\
        Overall LPIPS vs.\ Outside LPIPS & 1/10 & 0.80 \\
        \bottomrule
    \end{tabularx}
\end{minipage}
\end{table}

\subsection{Dataset and Subtitle Diagnostics}
\label{app:finding2-stratified-diagnostics}
Table~\ref{tab:finding2-attribute-slices} gives the Audit300 groups used in
the main-paper Figure~\ref{fig:finding2-attributes}.

\noindent\textbf{Dataset slices.}
Real-HQ has the highest Subtitle LPIPS for every method, whereas AI-generated
videos have the lowest Outside LPIPS. Temporal ordering does not simply
follow these spatial trends. Medium-duration clips also have higher
Subtitle LPIPS than Short clips for all guided methods. These associations
identify conditions worth inspecting, but source and resolution cannot be
separated here: Real-HQ is entirely 1080p-class, while Real-LQ and
AI-generated are 720p-class. The Long slice has only ten Real-LQ videos.

The Hard style-complexity group is easier for guided methods on Subtitle
LPIPS but harder for CLEAR. The metadata score therefore measures rendering
complexity, not a universal difficulty scale. In particular, access to a
mask changes which subtitle attributes create a challenge.

\begin{figure}[!htbp]
\centering
\includegraphics[width=\textwidth]{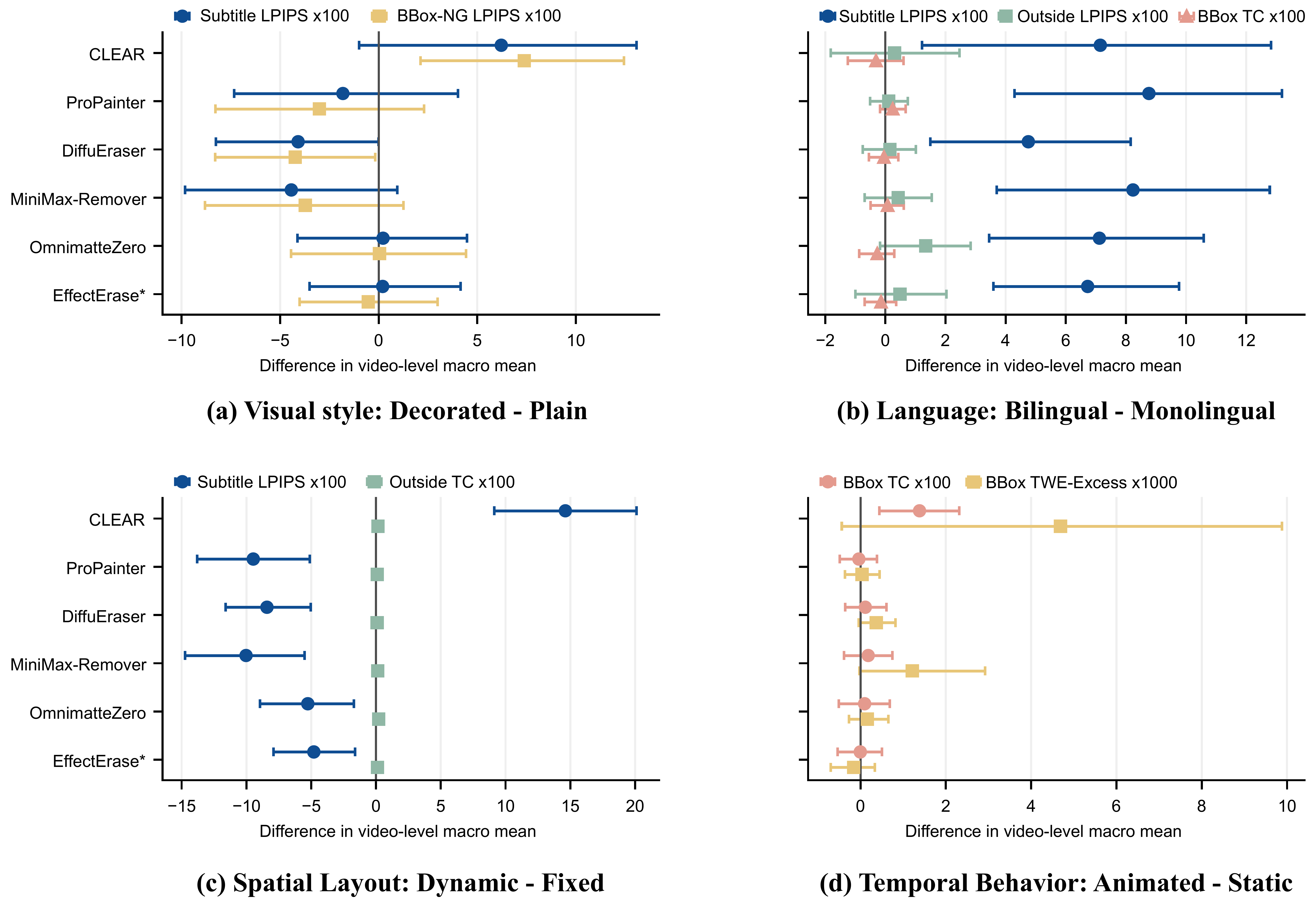}
\vspace{-20pt}
\caption{\textbf{Subtitle-factor contrasts on Audit300.} Decorated--Plain (96/107),
Bilingual--Monolingual (150/150), Dynamic--Fixed (145/155), and
Animated--Static (150/150). Positive differences indicate higher error in
the first group; bars are 95\% bootstrap intervals. EffectErase uses its
verified prefix.}
\label{fig:finding2-subtitle-factors}
\end{figure}

\noindent\textbf{Subtitle factors.}
Figure~\ref{fig:finding2-subtitle-factors} contrasts independent groups of
base videos. Differences are first minus second group means, with 4,000
within-group video-bootstrap resamples. 
Bilingual subtitles have
higher Subtitle LPIPS for every method, with intervals excluding zero,
while their Outside LPIPS and BBox TC contrasts are inconclusive. Dynamic
layouts associate with higher Subtitle LPIPS for CLEAR but lower error for
all guided methods. Decorated styles likewise have method-dependent effects:
DiffuEraser improves on both Subtitle and BBox-NG LPIPS, while CLEAR's
BBox-NG error increases. The differing responses reinforce the value of separating blind and guided
methods and evaluating local views. Because the groups contain different
base videos, these contrasts describe associations rather than causal
effects of individual subtitle attributes.

\begin{figure}[!htbp]
\centering
\includegraphics[width=0.84\textwidth]{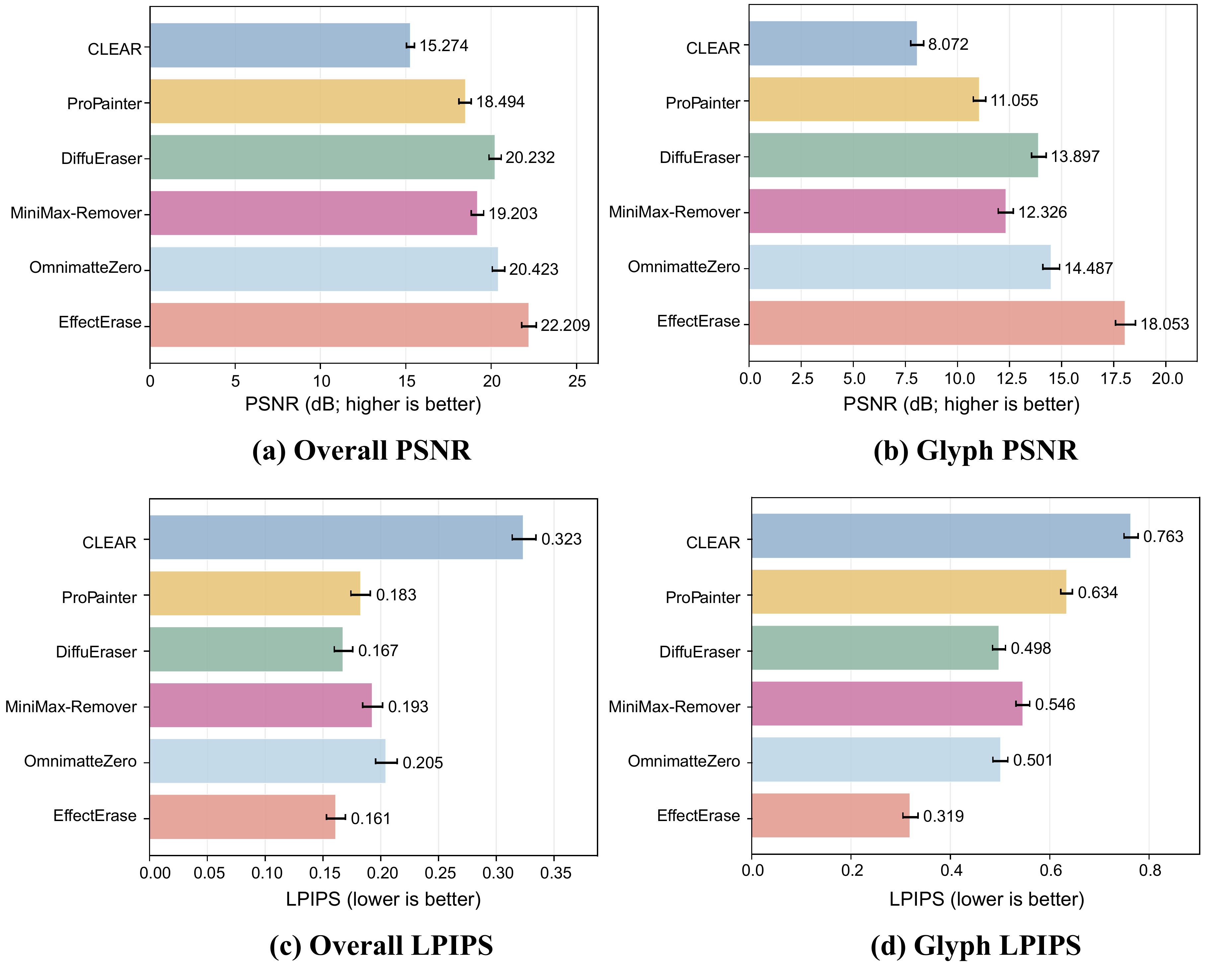}
\vspace{-12pt}
\caption{\textbf{Box-style stress results with video-bootstrap 95\% intervals.}
Each measurement uses 464 videos except EffectErase's Subtitle metrics
($n=463$, owing to one empty glyph support). Quality uses verified outputs,
including partial prefixes; missing values are not imputed.}
\label{fig:box-style-stress}
\end{figure}

\noindent\textbf{Box-style stress set.}
The final common-evaluable set contains 464 Boxed/Subtitle-box videos.
Their text-tight BBoxes need not cover the rendered box, so Outside cannot
identify unaffected content. Figure~\ref{fig:box-style-stress} therefore
reports only Overall and Subtitle PSNR/LPIPS, omitting Preservation, BBox
temporal scores, and SR-Q. Good glyph-core fidelity alone cannot establish recovery of the rendered
box: changes outside a text-tight BBox may be necessary to remove it.
The stress set makes this annotation boundary explicit.

\subsection{Full-Set Delivery and Coverage}
\label{app:finding3-delivery}
All delivery rates use 3,000 scheduled videos per method. Complete outputs
have full spatial and temporal coverage; verified partial outputs have
positive but incomplete support. Inference failures, missing artifacts, and
evaluation failures are distinct terminal states
(Table~\ref{tab:finding3-terminal-counts}). A missing artifact alone does not
establish an inference failure. The six evaluation failures concern the
same video.

\begin{table}[!htbp]
    \centering

\centering
    \caption{\textbf{Terminal-state counts: }complete (C), verified partial (P), inference failure (IF), missing artifact (MA), and evaluation failure (EF). Each row sums to 3,000.}
    \label{tab:finding3-terminal-counts}
    \small
    \setlength{\tabcolsep}{4pt}
    \renewcommand{\arraystretch}{1.08}
    \renewcommand{\arraystretch}{1.14}
    \begin{minipage}{0.83\linewidth}
\begin{tabularx}{\linewidth}{l*{5}{>{\centering\arraybackslash}X}}
        \toprule
\rowcolor{AppHeader}
        \textbf{Method} & \textbf{C} & \textbf{P}
        & \textbf{IF} & \textbf{MA} & \textbf{EF} \\
        \midrule
        CLEAR & 168 & 2,831 & 0 & 0 & 1 \\
        \rowcolor{AppBand}ProPainter & 2,949 & 0 & 50 & 0 & 1 \\
        DiffuEraser & 2,171 & 799 & 29 & 0 & 1 \\
        \rowcolor{AppBand}MiniMax-Remover & 924 & 1,947 & 128 & 0 & 1 \\
        OmnimatteZero & 630 & 2,255 & 91 & 23 & 1 \\
        \rowcolor{AppBand}EffectErase & 117 & 2,810 & 72 & 0 & 1 \\
        \bottomrule
    \end{tabularx}
\end{minipage}
\end{table}

\begin{figure}[!htbp]
\centering
\includegraphics[width=0.75\textwidth]{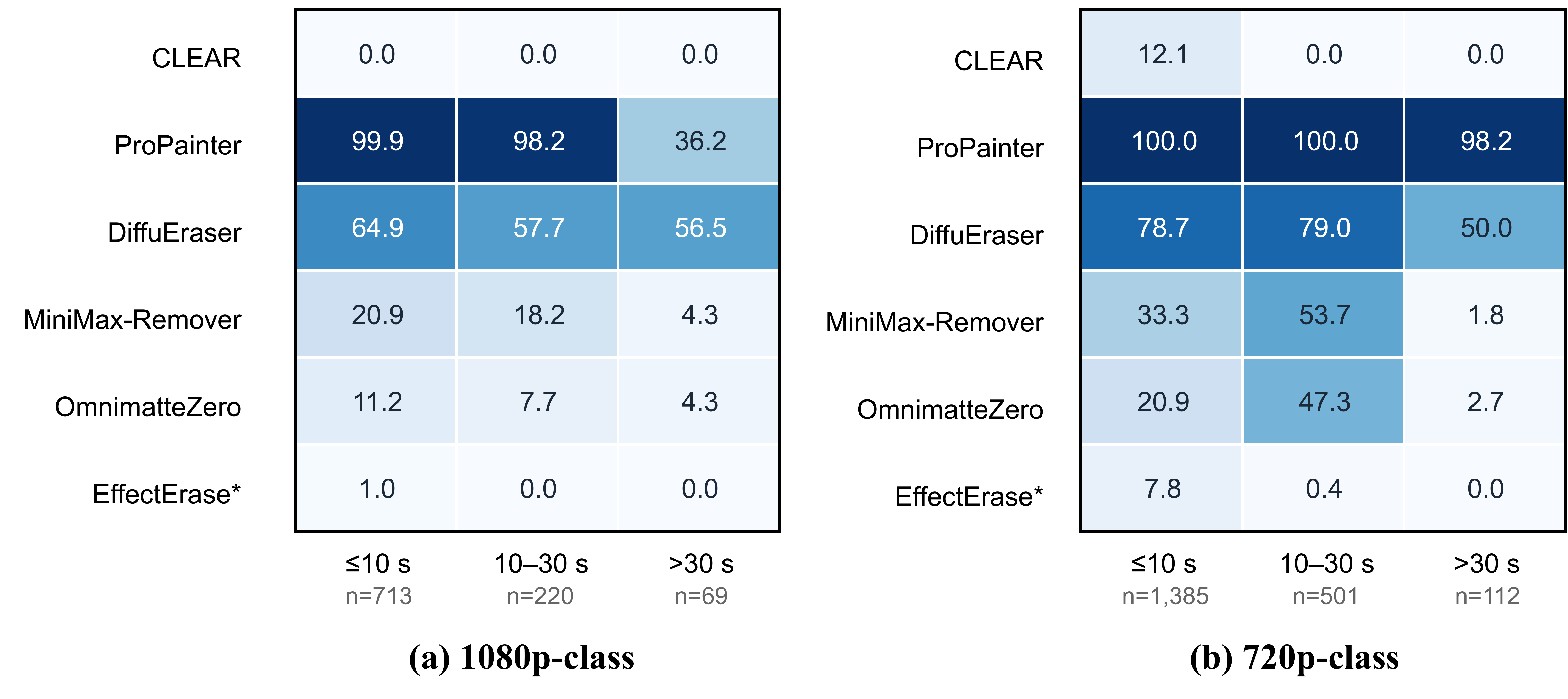}
\vspace{-12pt}
\caption{\textbf{Delivery across resolution and duration groups. }Rates use all
scheduled videos in each group, including failed and missing outputs.
Terminal-state mixtures distinguish verified truncation from failure.}
\label{fig:finding3-delivery}
\end{figure}

\noindent\textbf{Resolution and duration.}
Figure~\ref{fig:finding3-delivery} breaks down delivery by target resolution
and duration. ProPainter usually completes videos, with reduced coverage for
long, high-resolution inputs. CLEAR and EffectErase predominantly return
partial outputs under their frozen temporal contracts. This distinguishes
contractual truncation from inference failure and makes delivery a property
of the evaluated configuration as well as the model.

\begin{figure}[!htbp]
\centering
\includegraphics[width=0.8\textwidth]{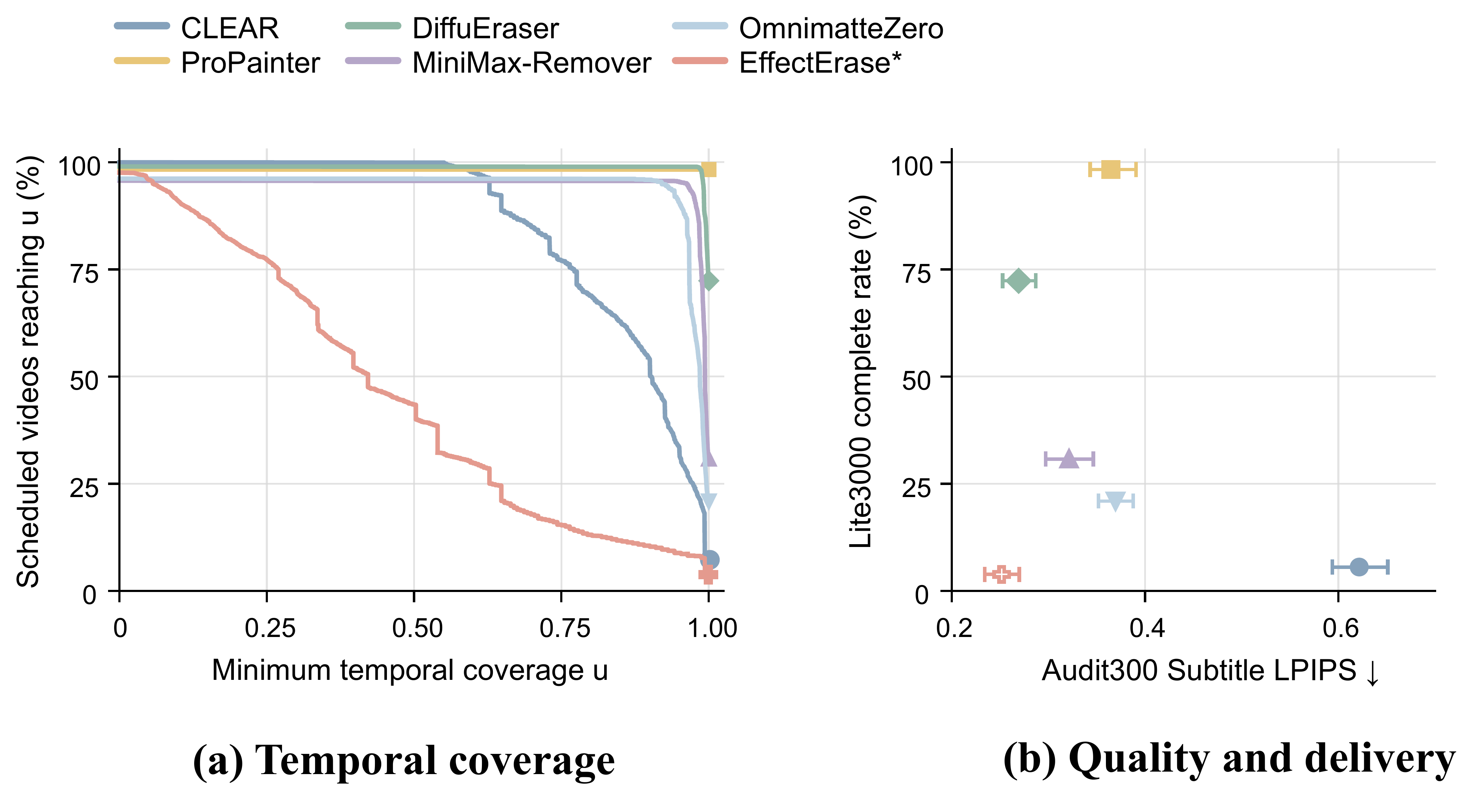}
\vspace{-15pt}
\caption{\textbf{Quality--coverage relationship across evaluation scopes.}
Left: temporal coverage over all scheduled videos.
Right: Audit300 Subtitle LPIPS versus Lite3000 complete-delivery rate.
Quality intervals are video-bootstrap 95\% CIs.
EffectErase quality is computed on verified prefixes, while delivery is
assessed against full targets.}
\label{fig:finding3-coverage-bridge}
\end{figure}

\noindent\textbf{Coverage and usable output.} Figure~\ref{fig:finding3-coverage-bridge} distinguishes near-complete from
severely truncated outputs. Its coverage curves use all scheduled videos,
assigning non-evaluable records zero coverage only for this plot; their
quality stays missing. Full temporal coverage does not imply full spatial
coverage. Quality and delivery also use different sets and are reported
separately. The strict six-method complete non-box intersection contains
only 80 short, 720p-class videos, nearly all AI-generated. Restricting
evaluation to this set would substantially change the benchmark population,
so it cannot support a representative full-video efficiency comparison.

\subsection{Evaluable Scope and Conditional Quality}
\label{app:finding4-native-success}
\begin{figure}[!htbp]
\centering
\includegraphics[width=0.9\textwidth]{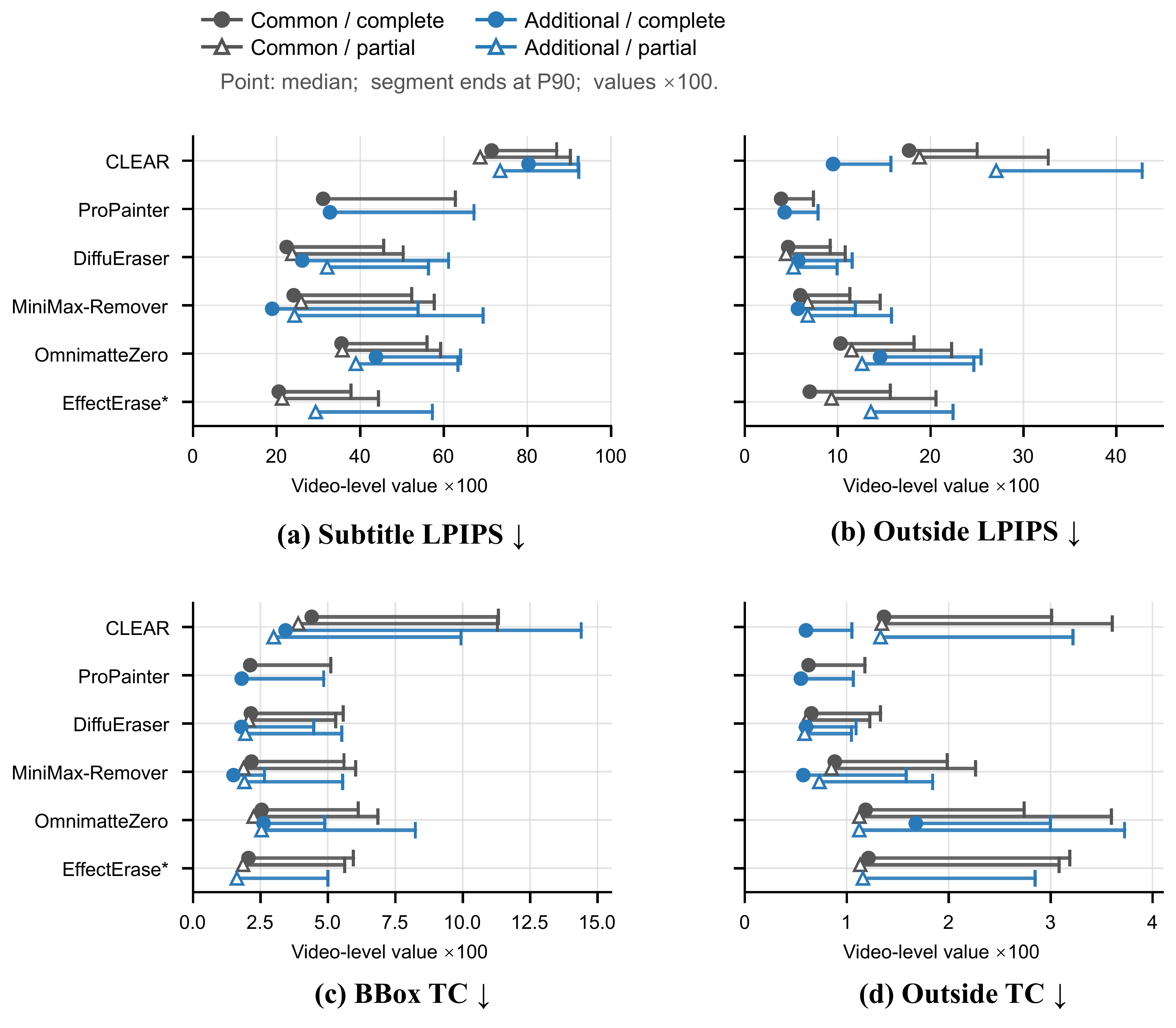}
\vspace{-13pt}
\caption{\textbf{Conditional quality on common and additional videos. }Points are
medians, and segments extend to the 90th percentile, not confidence limits.
Values are multiplied by 100.}
\label{fig:finding4-native-tails}
\end{figure}

The final non-box primary pool has 2,499 videos; its six-model common
intersection has 2,298. Table~\ref{tab:finding4-native-set-counts} gives each
method's additional evaluable scope. These sets can overlap and are dominated
by Real-LQ videos; all AI-generated primary-pool videos already lie in the
common intersection. 
Figure~\ref{fig:finding4-native-tails} separates complete and partial outputs
and distinguishes typical error from the upper tail. ProPainter's added
videos increase both median and tail Subtitle LPIPS. MiniMax-Remover instead
has a lower median but a worse tail, which a mean alone would conceal.
For DiffuEraser and EffectErase, worse restoration on added videos coexists
with lower BBox TC. For example, DiffuEraser's median Subtitle LPIPS rises
from 0.228 to 0.285, while median BBox TC falls from 0.0212 to 0.0187.
MiniMax-Remover's median Subtitle LPIPS falls from 0.253 to 0.219, while its
90th percentile rises from 0.558 to 0.671. These summaries pool complete and
partial outputs with valid metrics. Lower temporal discrepancy or a lower median can
therefore conceal a deterioration in another aspect of quality.

The additional sets are selected by method evaluability and differ in
composition. These shifts describe each method's expanded operating scope,
not the effect of a controlled increase in difficulty. Common-set results
support shared-ID comparisons; native results reveal what those comparisons
leave out and do not support cross-method rankings on different IDs.

\begin{table}[!htbp]
\centering

\centering\small
\caption{\textbf{Method-native evaluable scope in the non-box primary pool.}
All methods share 2,298 videos; native sets include verified partial outputs.}
\label{tab:finding4-native-set-counts}
\renewcommand{\arraystretch}{1.14}
\begin{minipage}{0.65\linewidth}
\begin{tabularx}{\linewidth}{l*{2}{>{\raggedleft\arraybackslash}X}}
\toprule
\rowcolor{AppHeader}
\textbf{Method} & \textbf{Native set} & \textbf{Additional videos} \\
\midrule
CLEAR & 2,499 & 201 \\
\rowcolor{AppBand}ProPainter & 2,456 & 158 \\
DiffuEraser & 2,474 & 176 \\
\rowcolor{AppBand}MiniMax-Remover & 2,392 & 94 \\
OmnimatteZero & 2,407 & 109 \\
\rowcolor{AppBand}EffectErase & 2,436 & 138 \\
\bottomrule
\end{tabularx}
\end{minipage}
\end{table}

\subsection{Human Alignment and Qualitative Evidence}
\label{app:finding5-human-qualitative}

\begin{figure}[!htbp]
\centering
\includegraphics[width=\textwidth]{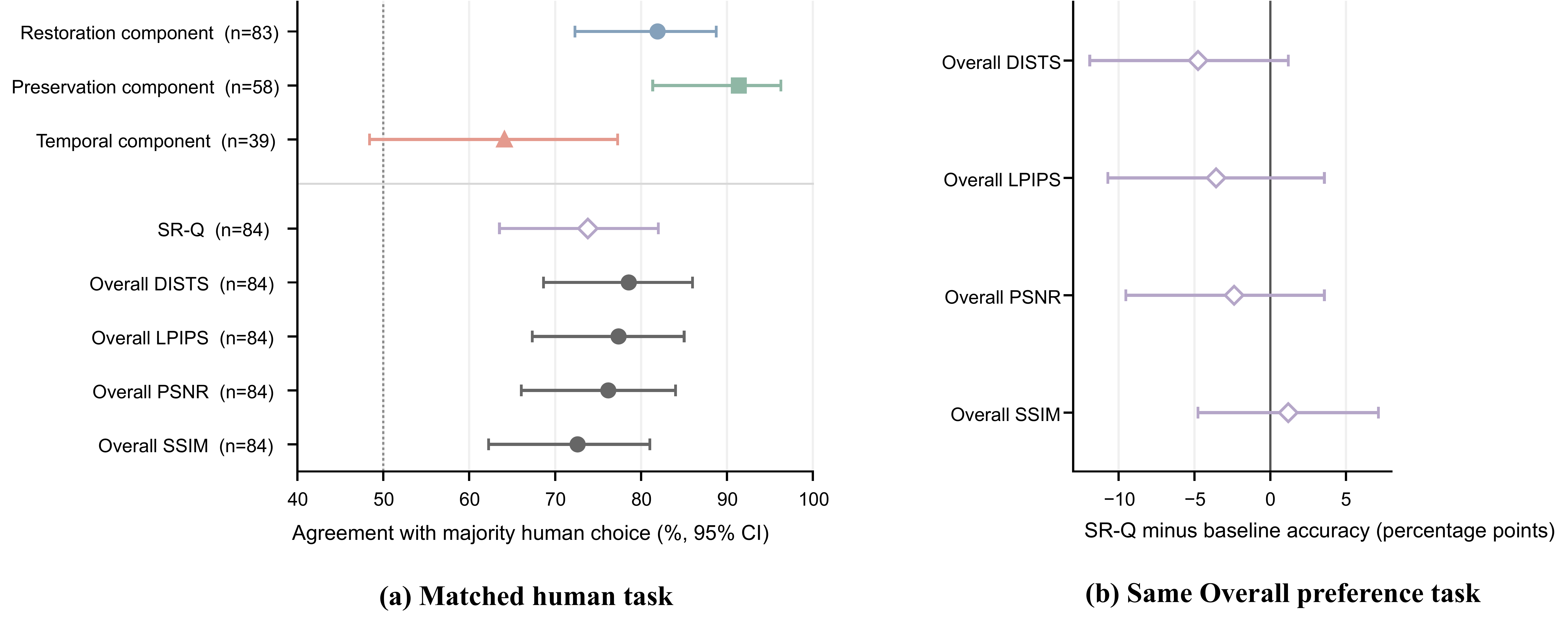}
\vspace{-22pt}
\caption{\textbf{Human alignment and scalar comparison.} (a) Matched-task agreement
with Wilson 95\% intervals; $n$ counts decisive pairs and the dashed line marks
50\%. (b) SR-Q minus baseline accuracy on the same 84 decisive Overall pairs,
with paired-bootstrap 95\% intervals. Component and Overall rows address
different judgments and should be interpreted within their tasks.}
\label{fig:finding5-human-validation}
\end{figure}

\noindent\textbf{Agreement with human preferences.}
The study protocol and interface are given in
Appendix~\ref{app:human-protocol} and
Figure~\ref{fig:human-evaluation-interface}; the cases in
Appendix~\ref{app:case-gallery} illustrate the distinctions being assessed.
Figure~\ref{fig:finding5-human-validation} summarizes alignment, with task
decisiveness and all preregistered candidates reported in
Tables~\ref{tab:finding5-human-ties} and~\ref{tab:finding5-human-complete}.
Restoration and Preservation components agree with most decisive
preferences, although neither improves observed accuracy over its LPIPS
counterpart. Human validation thus supports the regional views more clearly
than the benefit of combining their metrics.

Temporal judgments are less decisive, with most pairs undecided and low
inter-annotator agreement. The resulting small decisive subset yields wide
intervals, leaving temporal alignment uncertain. Accuracy is conditional on
decisive preferences; tie rates show how much of the full study it represents.

\begin{table}[!htbp]
    \centering

\centering
    \caption{
    \textbf{Human-response decisiveness and agreement.}
    Each task contains 100 pairs; $\kappa$ is Fleiss' inter-annotator
    agreement.
    }
    \label{tab:finding5-human-ties}
    \small
    \setlength{\tabcolsep}{4pt}
    \renewcommand{\arraystretch}{1.14}
    \begin{minipage}{0.7\linewidth}
\begin{tabularx}{\linewidth}{Y*{3}{>{\centering\arraybackslash}X}}
        \toprule
\rowcolor{AppHeader}
        \textbf{Task} & \textbf{Decisive $n$}
        & \textbf{Undecided (\%)} & \textbf{$\kappa$} \\
        \midrule
        Restoration  & 83 & 17 & 0.544 \\
        Preservation & 58 & 42 & 0.517 \\
        Temporal     & 39 & 61 & 0.123 \\
        Overall      & 84 & 16 & 0.578 \\
        \bottomrule
    \end{tabularx}
\end{minipage}
\end{table}

\begin{table}[!htbp]
    \centering
    \caption{
    \textbf{Held-out metric--human alignment.}
    Accuracy uses decisive preferences; intervals are Wilson 95\% CIs.
    Kendall's $\tau_b$ uses signed aggregate preferences.
    }
    \label{tab:finding5-human-complete}
    \label{tab:all-human-alignment}

    \footnotesize
    \setlength{\tabcolsep}{2.8pt}
    \renewcommand{\arraystretch}{1.14}

    \begin{tabularx}{\linewidth}{
        @{}
        >{\hsize=0.90\hsize}Z
        >{\hsize=1.35\hsize}Z
        >{\hsize=0.90\hsize}C
        >{\hsize=1.15\hsize}C
        >{\hsize=0.70\hsize}C
        @{}
    }
        \toprule[1.3pt]

        \rowcolor{AppHeader}
        \textbf{Human task}
        & \textbf{Candidate}
        & \makecell{\textbf{Accuracy}\\\textbf{(\%)}}
        & \textbf{95\% CI}
        & \textbf{$\tau_b$} \\

        \midrule

        \rowcolor{AppBand}
        \textbf{Restoration}
        & Restoration component
        & 81.9
        & [72.3, 88.7]
        & 0.471 \\

        {\scriptsize $(n=83)$}
        & Subtitle LPIPS
        & 88.0
        & [79.2, 93.3]
        & 0.501 \\

        & Subtitle PSNR
        & 81.9
        & [72.3, 88.7]
        & 0.443 \\

        & Subtitle cDISTS
        & 75.9
        & [65.7, 83.8]
        & 0.391 \\

        \addlinespace[2pt]

        \rowcolor{AppBand}
        \textbf{Preservation}
        & Preservation component
        & 91.4
        & [81.4, 96.3]
        & 0.559 \\

        {\scriptsize $(n=58)$}
        & Outside LPIPS
        & 91.4
        & [81.4, 96.3]
        & 0.519 \\

        & Outside PSNR
        & 89.7
        & [79.2, 95.2]
        & 0.574 \\

        \addlinespace[2pt]

        \rowcolor{AppBand}
        \textbf{Temporal}
        & Temporal component
        & 64.1
        & [48.4, 77.3]
        & 0.222 \\

        {\scriptsize $(n=39)$}
        & BBox TC
        & 61.5
        & [45.9, 75.1]
        & 0.221 \\

        & BBox TWE-Excess
        & 59.0
        & [43.4, 72.9]
        & 0.150 \\

        & Outside TC
        & 74.4
        & [58.9, 85.4]
        & 0.280 \\

        & Outside TWE-Excess
        & 41.0
        & [27.1, 56.6]
        & 0.111 \\

        \addlinespace[2pt]

        \rowcolor{AppBand}
        \textbf{Overall}
        & SR-Q
        & 73.8
        & [63.5, 82.0]
        & 0.408 \\

        {\scriptsize $(n=84)$}
        & Overall DISTS
        & 78.6
        & [68.7, 86.0]
        & 0.389 \\

        & Overall LPIPS
        & 77.4
        & [67.4, 85.0]
        & 0.429 \\

        & Overall PSNR
        & 76.2
        & [66.1, 84.0]
        & 0.409 \\

        & Overall SSIM
        & 72.6
        & [62.3, 81.0]
        & 0.367 \\

        \bottomrule[1.3pt]
    \end{tabularx}
\end{table}

\noindent\textbf{Scalar comparison.}
On the same decisive Overall pairs, SR-Q does not outperform the strongest
conventional candidate. All paired-bootstrap accuracy-difference intervals
include zero, and all Holm-adjusted McNemar $p$ values are 1.0. SR-Q therefore
fails the preregistered promotion rule. We retain the raw fidelity profile
without refitting the scalar to these judgments.

\section{Computational Workload}
\label{app:computational_workload}
Retained inference logs total 10,137.4 GPU-hours, with one unshared GPU per worker. Workload sums elapsed
runtime times assigned GPU count, including 207 timed failures and observed
runtime up to termination for partial outputs. Table~\ref{tab:computational_workload}
covers 17,814 of 18,000 scheduled records.

\begin{table}[ht]
\centering

\centering\small
\caption{Recorded inference workload. Each method has 3,000 scheduled videos.}
\label{tab:computational_workload}
\renewcommand{\arraystretch}{1.14}
\begin{minipage}{0.8\linewidth}
\begin{tabularx}{\linewidth}{l*{2}{>{\raggedleft\arraybackslash}X}}
\toprule
\rowcolor{AppHeader}
\textbf{Method} & \textbf{Timed records} & \textbf{GPU-hours} \\
\midrule
CLEAR & 3,000 & 844.6 \\
\rowcolor{AppBand}ProPainter & 3,000 & 289.6 \\
DiffuEraser & 3,000 & 738.9 \\
\rowcolor{AppBand}MiniMax-Remover & 2,909 & 4,500.0 \\
OmnimatteZero & 2,977 & 2,021.4 \\
\rowcolor{AppBand}EffectErase & 2,928 & 1,742.9 \\
\midrule
\textbf{Total} & \textbf{17,814} & \textbf{10,137.4} \\
\bottomrule
\end{tabularx}
\end{minipage}
\end{table}

The totals exclude missing or invalid timing, overwritten attempts, untimed
stages, and metric evaluation. External decoding and, where applicable, mask
preparation precede timing for CLEAR, MiniMax, OmnimatteZero, and EffectErase.
Encoding is included for DiffuEraser and EffectErase but excluded from other
successful runs. These differences in timing boundaries prevent interpreting
the totals as either project cost or a full-video efficiency comparison.

\clearpage
\section{Qualitative Cases and Diagnostic Insights}
\label{app:case-gallery}
We retain five cases per diagnostic, selected by decreasing six-model range
in Subtitle LPIPS, Outside LPIPS, or BBox TC (ties broken by video ID).
Selection uses neither SR-Q nor human labels. Only unreadable assets, invalid
common crops, or unavailable shared temporal support permit rejection.
The gallery illustrates disagreements, not failure frequencies or an
unbiased ranking. Case records retain IDs, frame indices, crops, and scores.

Restoration comparisons use local crops; Preservation comparisons retain
full-frame context. Temporal strips show three consecutive shared frames,
read left to right. Views and aspect ratios are matched across methods.
CLEAR is blind; the other methods receive BBox masks.

\newcommand{\mfcasecell}[1]{\includegraphics[width=\linewidth,height=0.065\textheight,keepaspectratio]{assets/cases/#1.pdf}}
\subsection{Restoration: Text Removal and Scene Reconstruction}
\label{app:cases-restoration}
The local crops distinguish residual text from errors in the reconstructed scene.
\par\smallskip
\begin{center}\footnotesize\setlength{\tabcolsep}{3pt}\renewcommand{\arraystretch}{1.08}
\renewcommand{\mfcasecell}[1]{\includegraphics[width=\linewidth,height=0.069\textheight,keepaspectratio]{assets/cases/#1.pdf}}
\renewcommand{\tabularxcolumn}[1]{m{#1}}
\begin{tabularx}{\textwidth}{>{\raggedright\arraybackslash}m{0.16\textwidth}>{\centering\arraybackslash}X>{\centering\arraybackslash}X>{\centering\arraybackslash}X}
\toprule
\rowcolor{AppHeader}
\textbf{Method / view} & \makecell{\textbf{R1} / Shadowed bilingual\\{\scriptsize frame 0}} & \makecell{\textbf{R2} / Animated neon\\{\scriptsize frame 31}} & \makecell{\textbf{R3} / Plain Korean\\{\scriptsize frame 0}} \\
\midrule
\rowcolor{AppBand}Input & \mfcasecell{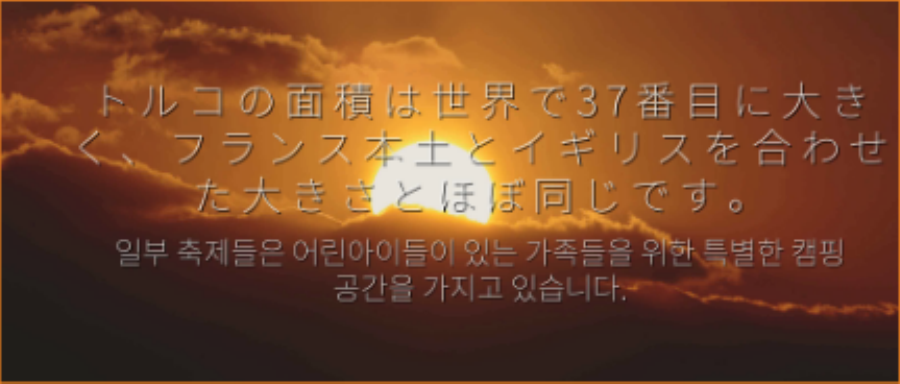} & \mfcasecell{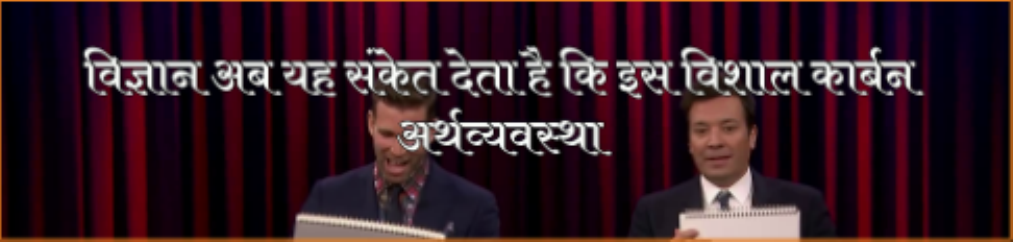} & \mfcasecell{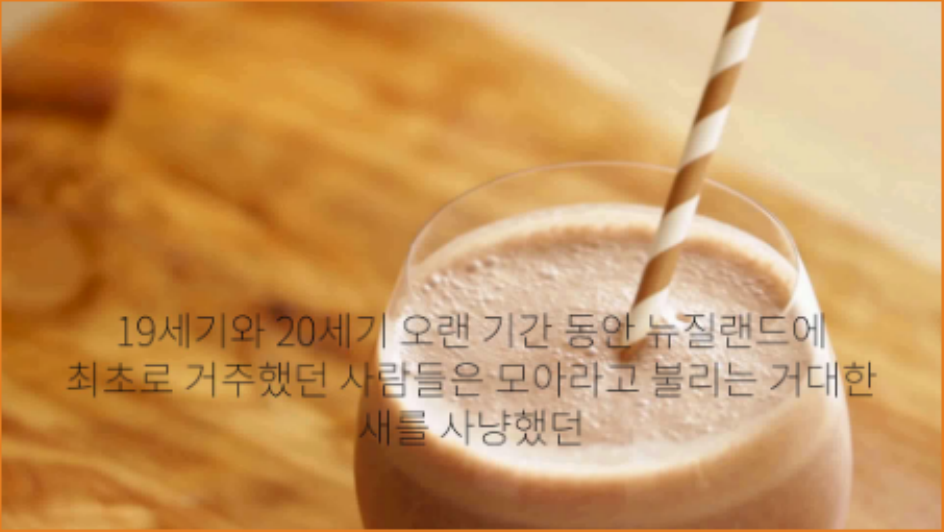} \\
Clean target & \mfcasecell{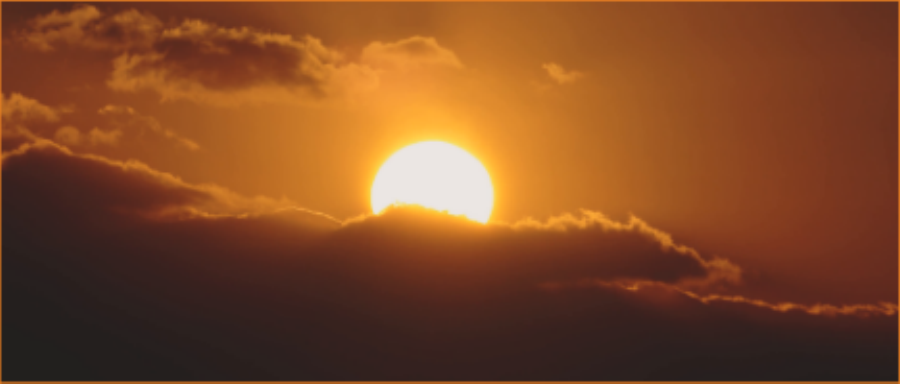} & \mfcasecell{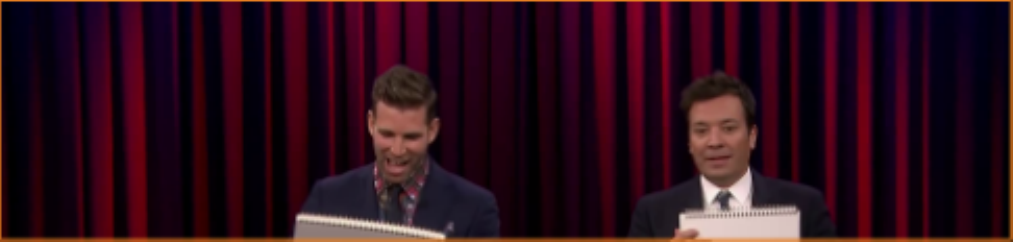} & \mfcasecell{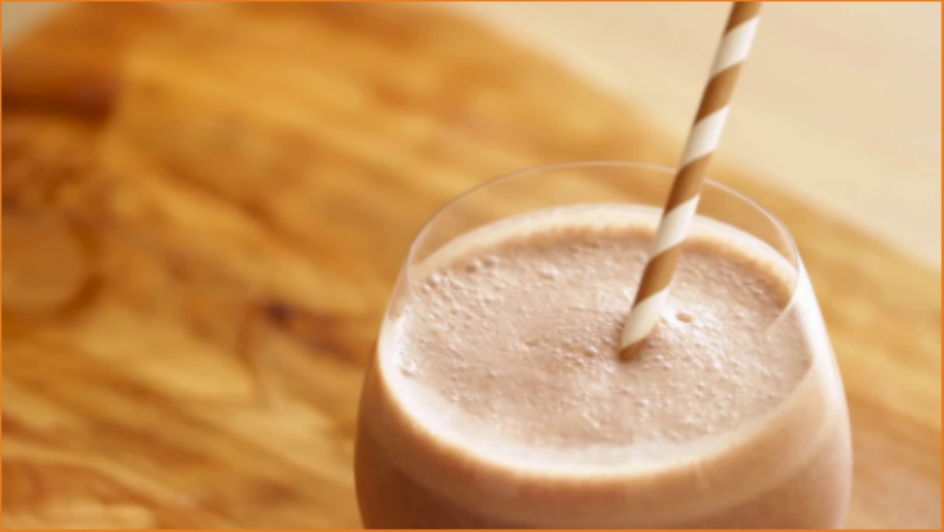} \\
\midrule
\rowcolor{AppBand}CLEAR (blind) & \mfcasecell{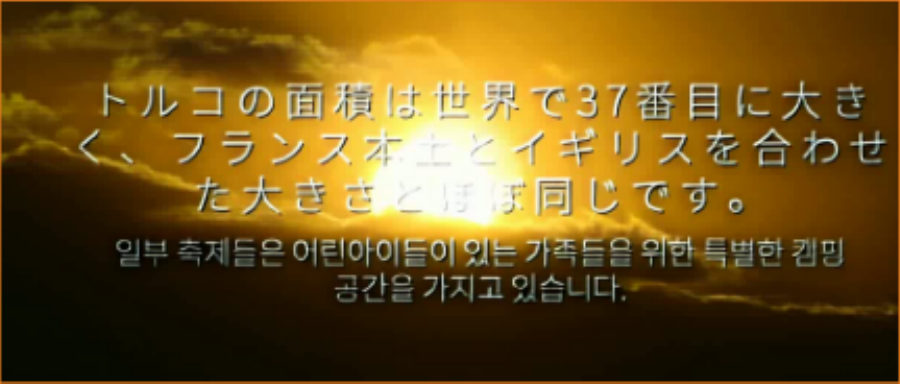} & \mfcasecell{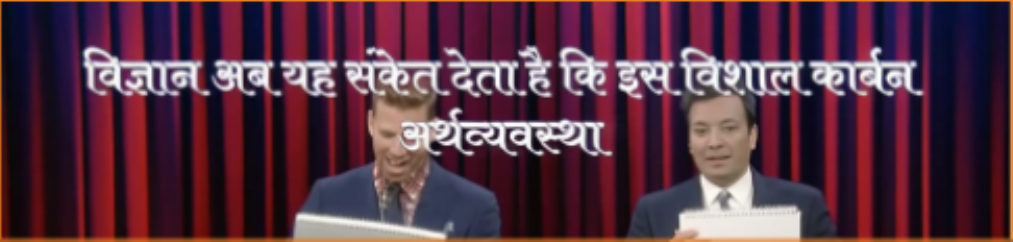} & \mfcasecell{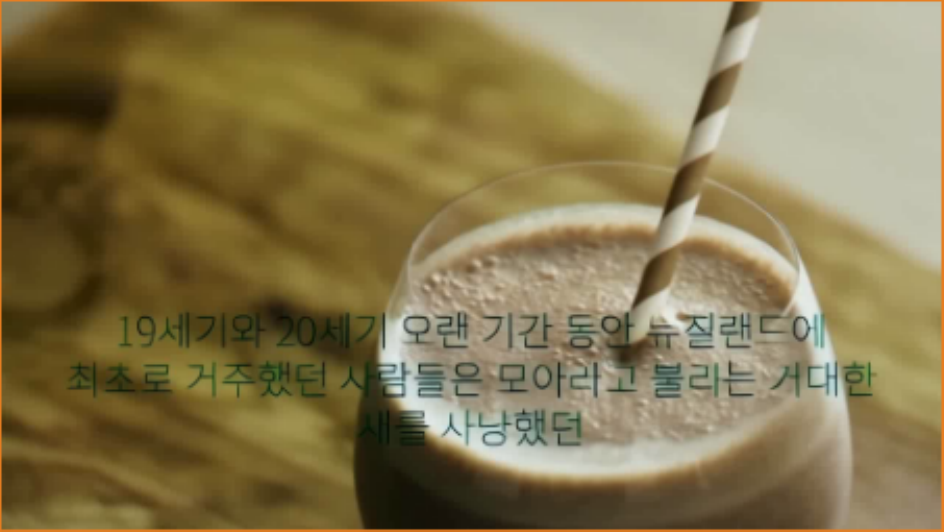} \\
ProPainter & \mfcasecell{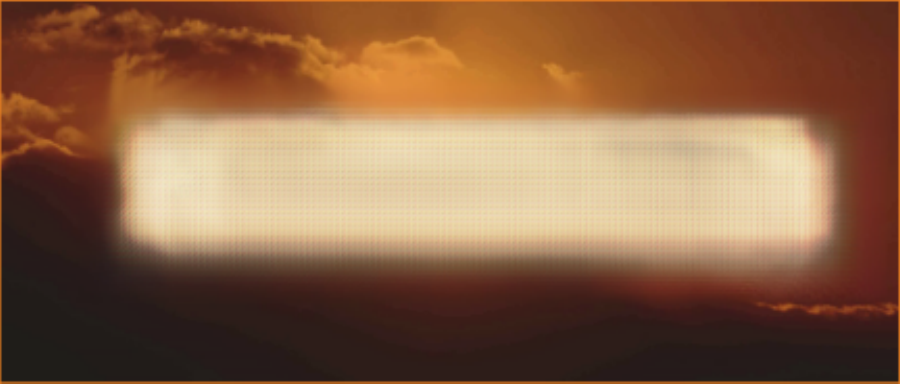} & \mfcasecell{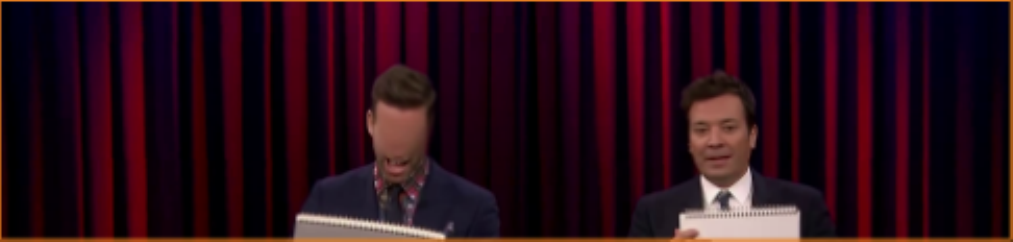} & \mfcasecell{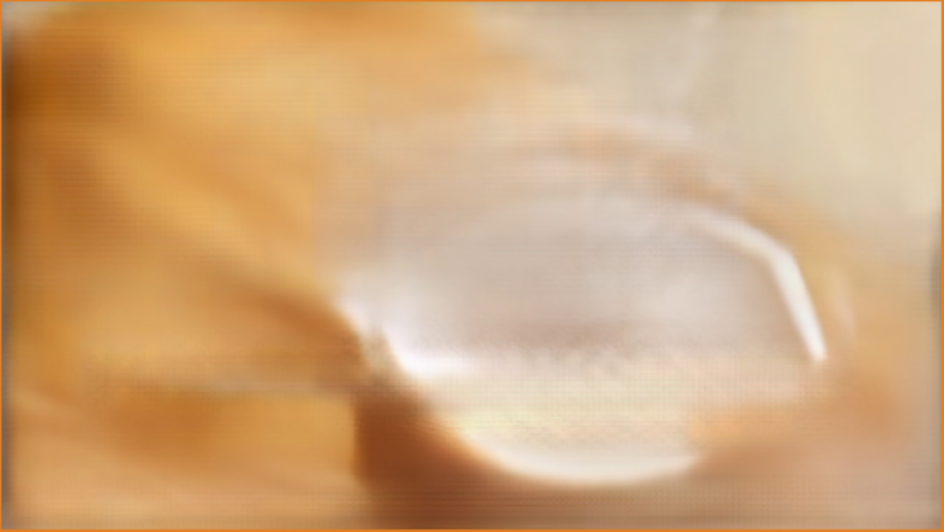} \\
\rowcolor{AppBand}DiffuEraser & \mfcasecell{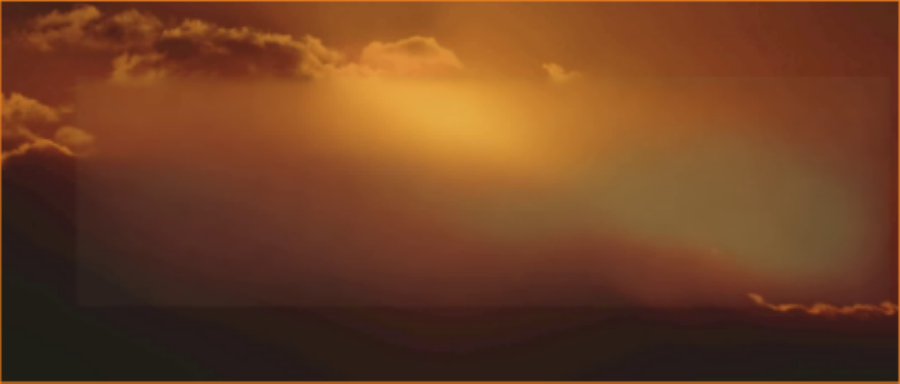} & \mfcasecell{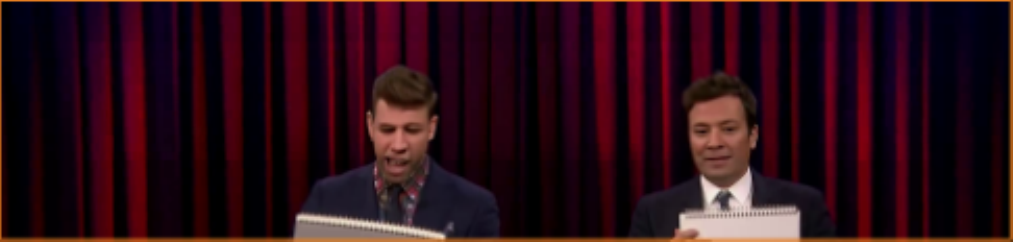} & \mfcasecell{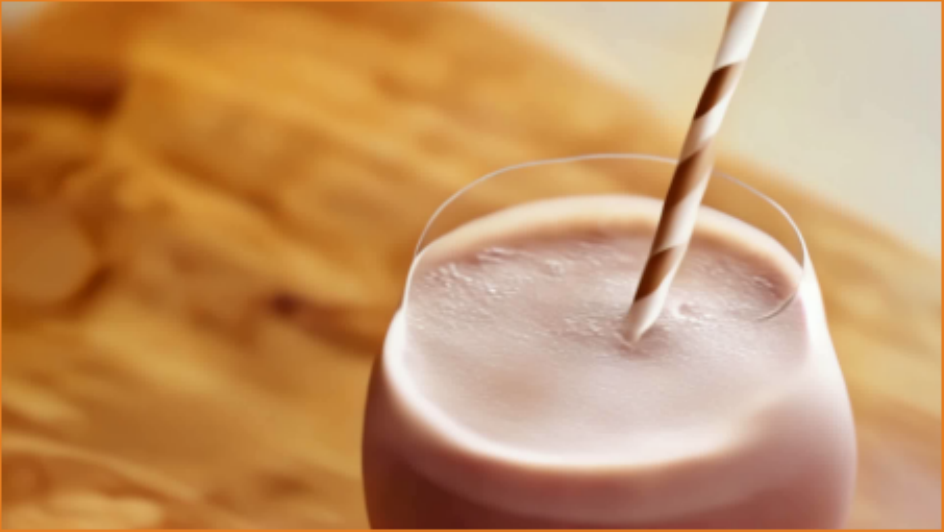} \\
MiniMax-Remover & \mfcasecell{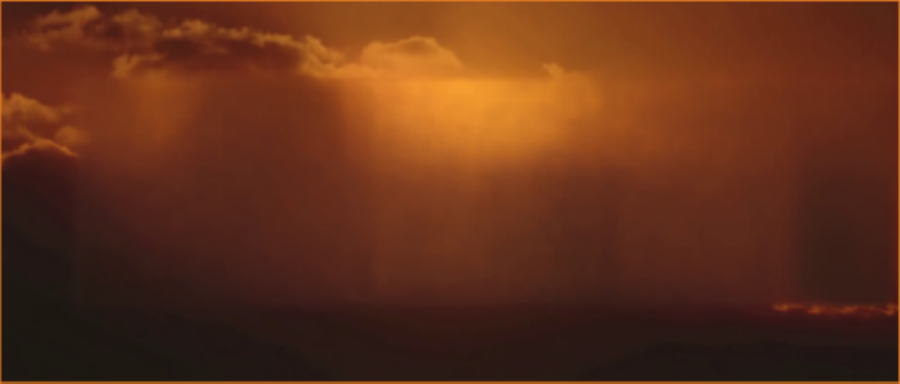} & \mfcasecell{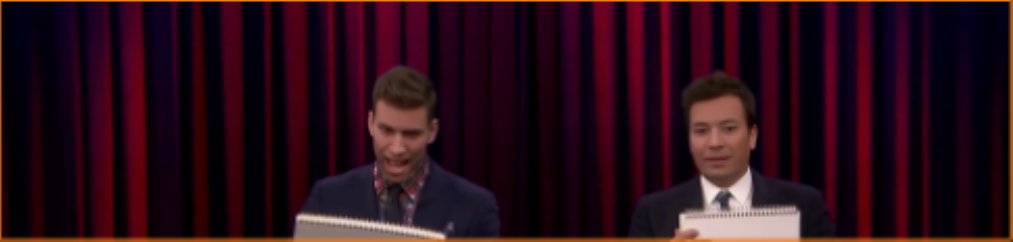} & \mfcasecell{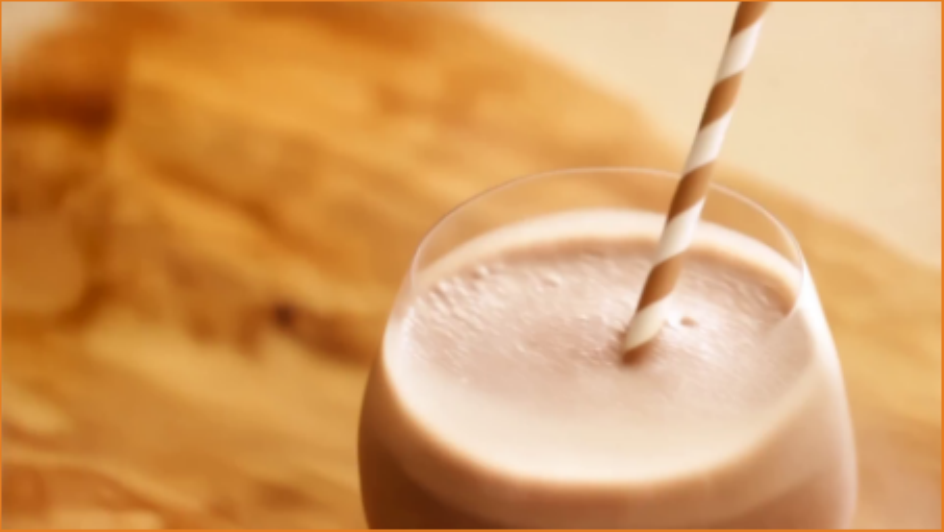} \\
\rowcolor{AppBand}OmnimatteZero & \mfcasecell{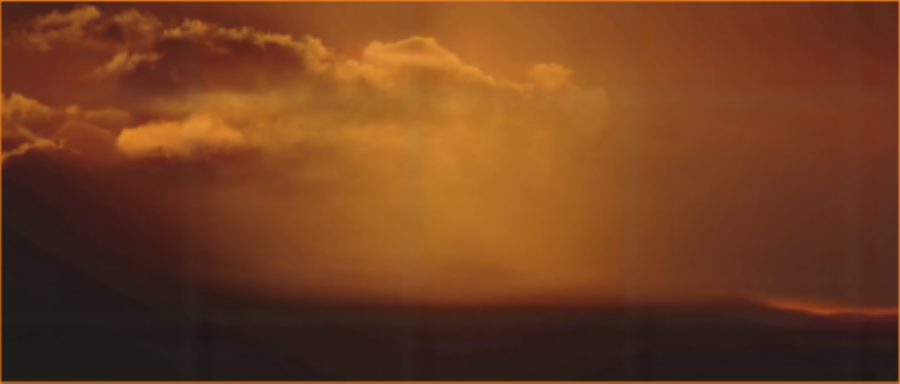} & \mfcasecell{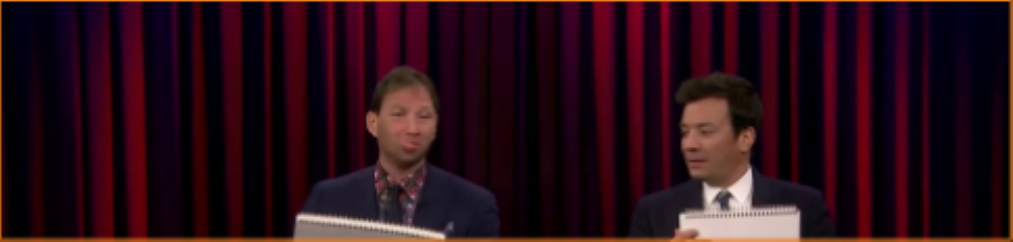} & \mfcasecell{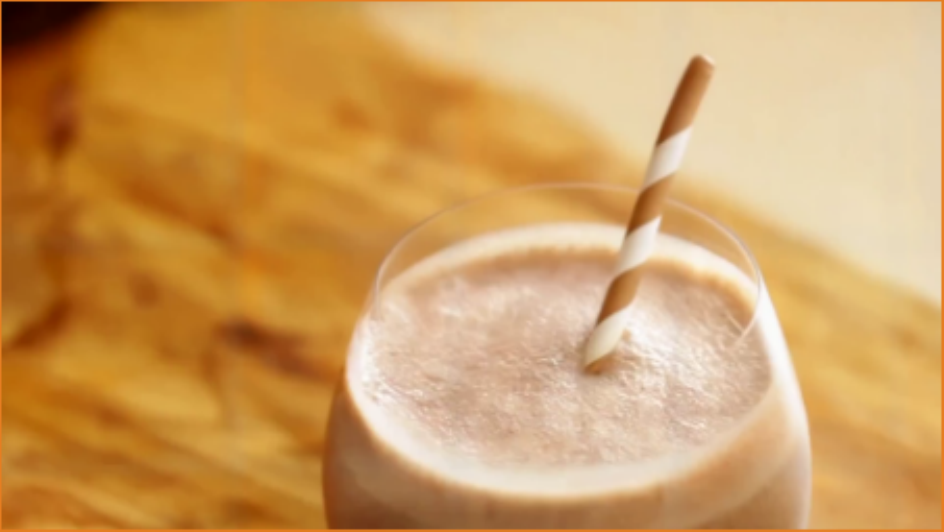} \\
EffectErase & \mfcasecell{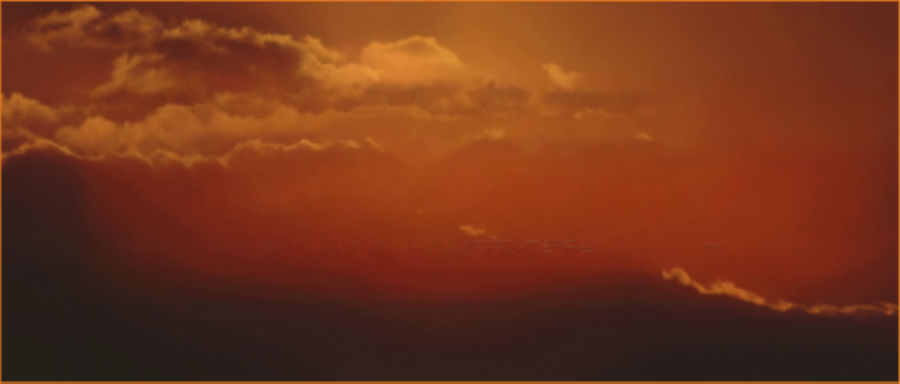} & \mfcasecell{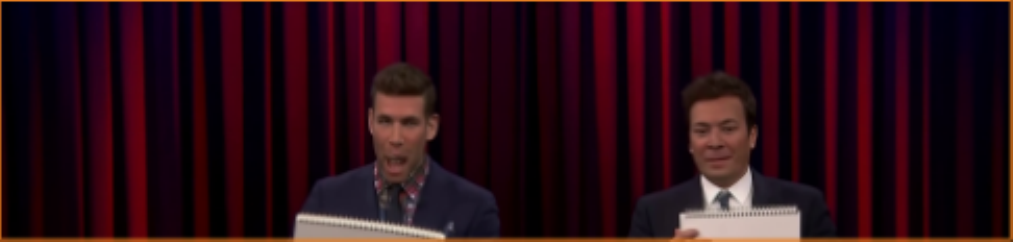} & \mfcasecell{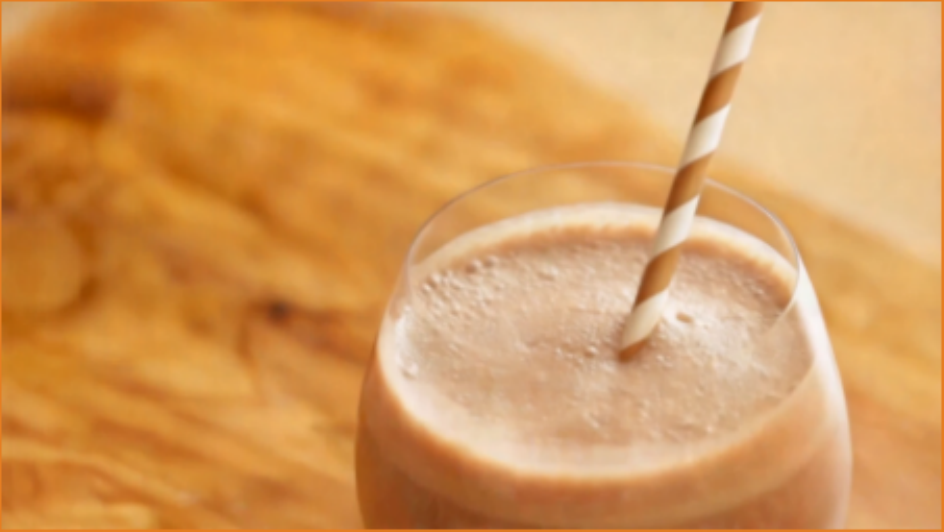} \\
\bottomrule
\end{tabularx}
\end{center}
\noindent In R1, CLEAR retains visible glyphs, ProPainter leaves a smooth
band, and generative fills change the cloud structure. R2's stage geometry
and R3's transparent glass expose further reconstruction errors. Removing
text is only part of the task: the replacement must also match the clean scene.
\clearpage
\subsection*{Restoration across Backgrounds}
The shared crop spans textured natural content in R4 and a dark background with a localized object in R5.
\par\smallskip
\begin{center}\footnotesize\setlength{\tabcolsep}{3pt}\renewcommand{\arraystretch}{1.08}
\renewcommand{\mfcasecell}[1]{\includegraphics[width=\linewidth,height=0.085\textheight,keepaspectratio]{assets/cases/#1.pdf}}
\renewcommand{\tabularxcolumn}[1]{m{#1}}
\begin{tabularx}{\textwidth}{>{\raggedright\arraybackslash}m{0.16\textwidth}>{\centering\arraybackslash}X>{\centering\arraybackslash}X}
\toprule
\rowcolor{AppHeader}
\textbf{Method / view} & \makecell{\textbf{R4} / Moving glow\\{\scriptsize frame 11}} & \makecell{\textbf{R5} / Animated Thai\\{\scriptsize frame 56}} \\
\midrule
\rowcolor{AppBand}Input & \mfcasecell{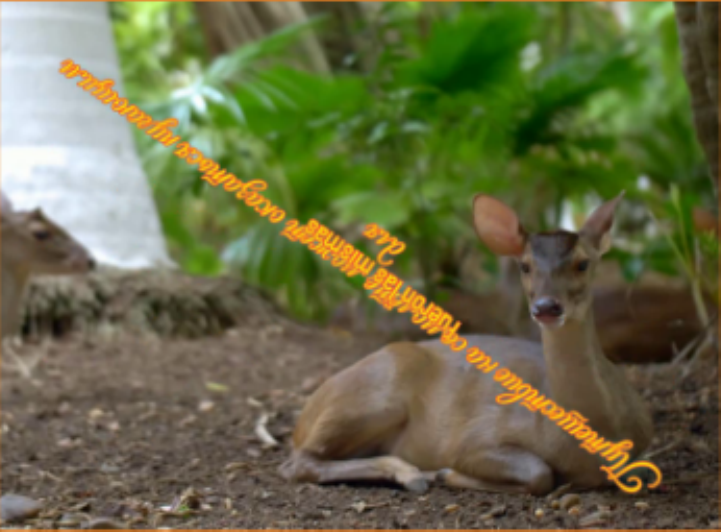} & \mfcasecell{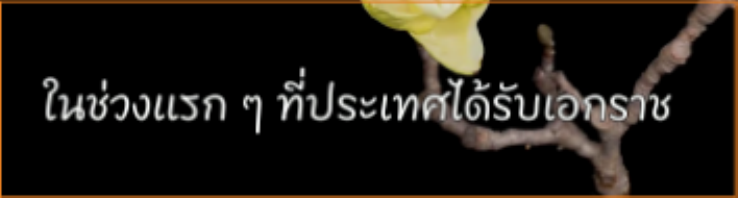} \\
Clean target & \mfcasecell{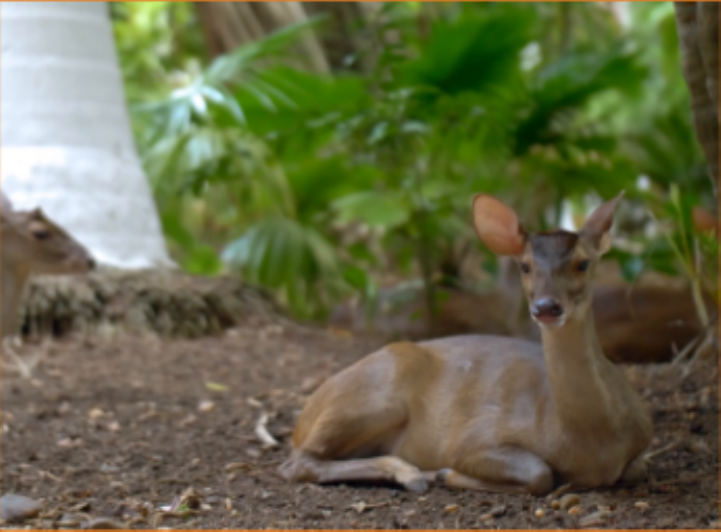} & \mfcasecell{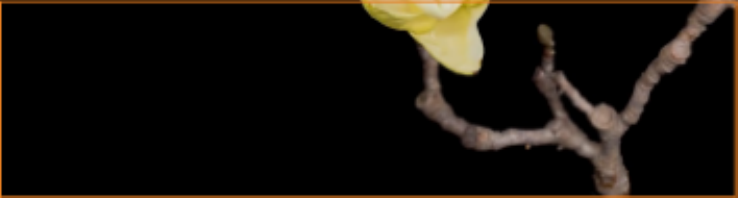} \\
\midrule
\rowcolor{AppBand}CLEAR (blind) & \mfcasecell{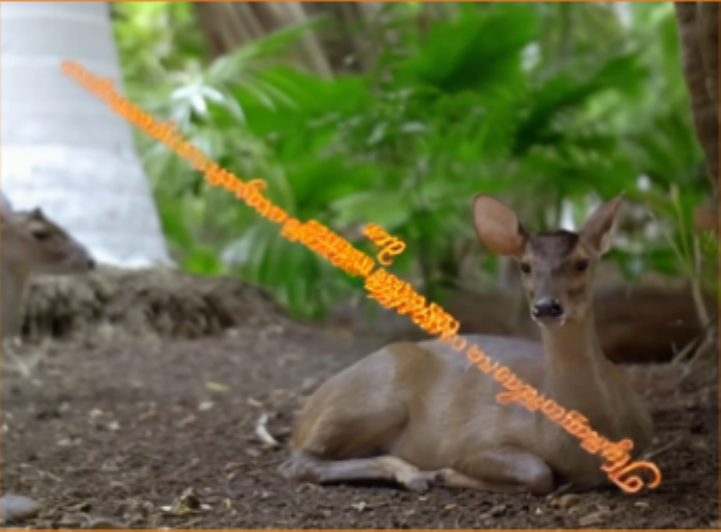} & \mfcasecell{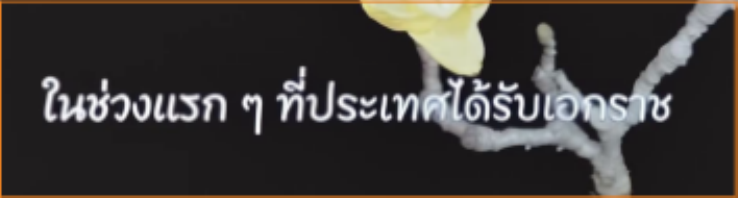} \\
ProPainter & \mfcasecell{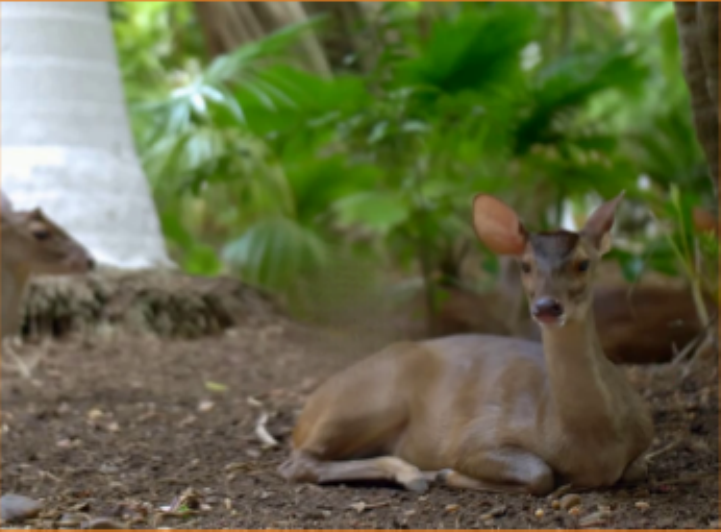} & \mfcasecell{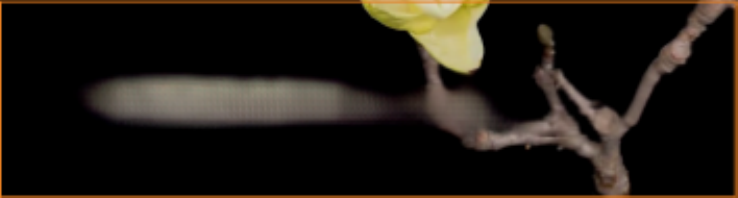} \\
\rowcolor{AppBand}DiffuEraser & \mfcasecell{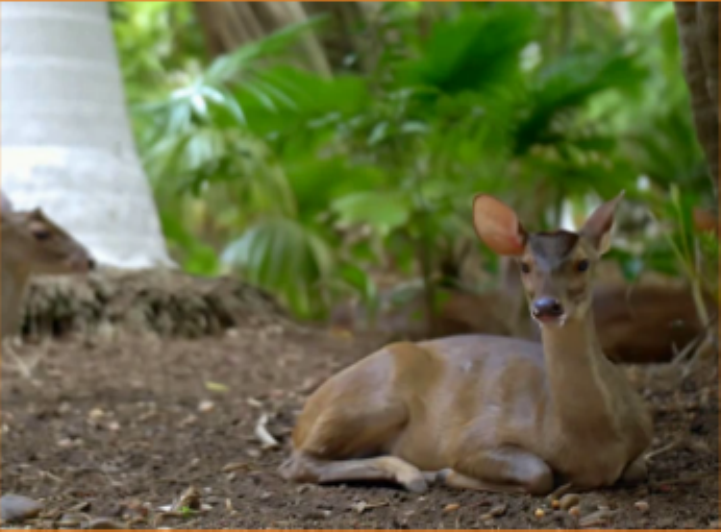} & \mfcasecell{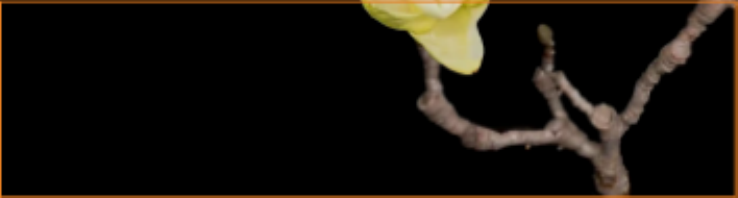} \\
MiniMax-Remover & \mfcasecell{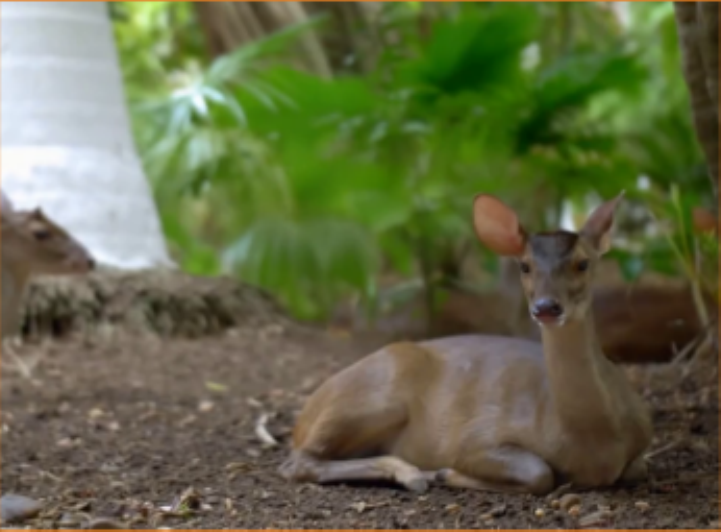} & \mfcasecell{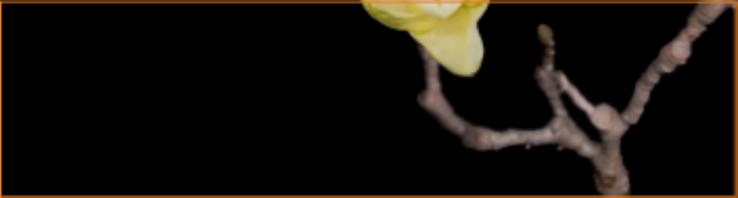} \\
\rowcolor{AppBand}OmnimatteZero & \mfcasecell{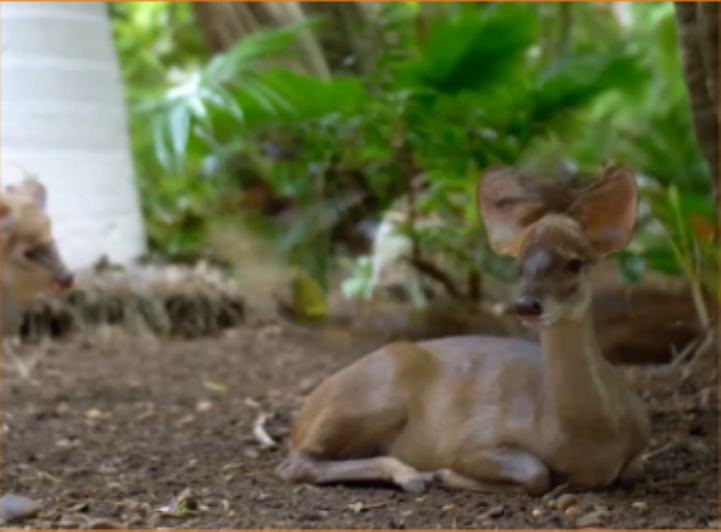} & \mfcasecell{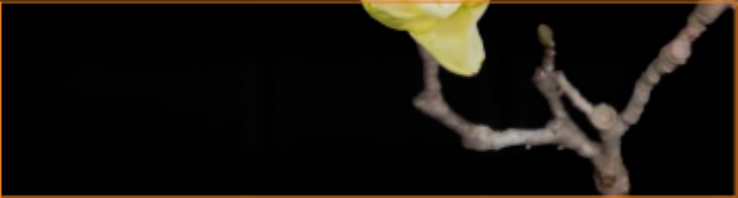} \\
EffectErase & \mfcasecell{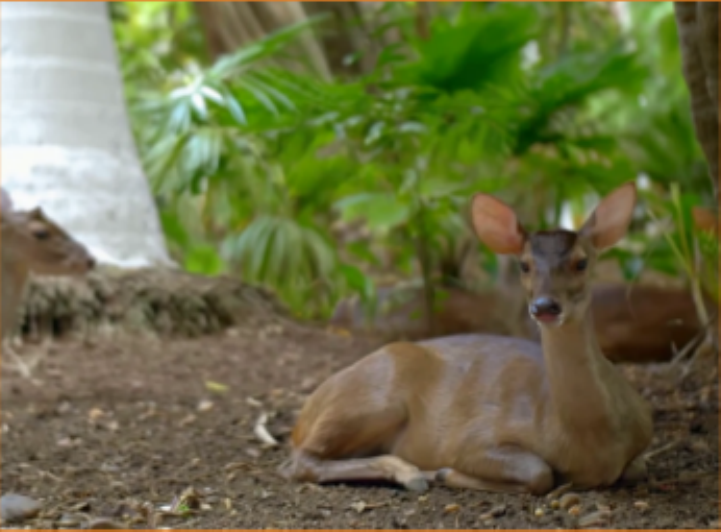} & \mfcasecell{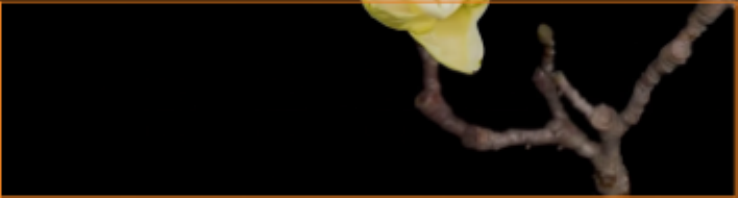} \\
\bottomrule
\end{tabularx}
\end{center}
\noindent R4 reveals losses in natural detail, while R5 makes changes to a
localized object visible against a dark background. Both illustrate the cost
of reconstructing a region that contains real scene content beneath the text.
They complement the factor-level results in
Appendix~\ref{app:finding2-stratified-diagnostics}, without attributing an
individual error to language or rendering style.
\clearpage
\subsection{Preservation: Changes beyond the Subtitle}
\label{app:cases-preservation}
Full frames expose changes beyond the subtitle region that local crops would omit.
\par\smallskip
\begin{center}\footnotesize\setlength{\tabcolsep}{3pt}\renewcommand{\arraystretch}{1.08}
\renewcommand{\mfcasecell}[1]{\includegraphics[width=\linewidth,height=0.085\textheight,keepaspectratio]{assets/cases/#1.pdf}}
\renewcommand{\tabularxcolumn}[1]{m{#1}}
\begin{tabularx}{\textwidth}{>{\raggedright\arraybackslash}m{0.16\textwidth}>{\centering\arraybackslash}X>{\centering\arraybackslash}X>{\centering\arraybackslash}X}
\toprule
\rowcolor{AppHeader}
\textbf{Method / view} & \makecell{\textbf{P1} / Abstract geometry\\{\scriptsize frame 47}} & \makecell{\textbf{P2} / Forest foliage\\{\scriptsize frame 9}} & \makecell{\textbf{P3} / Forest stream\\{\scriptsize frame 0}} \\
\midrule
\rowcolor{AppBand}Input & \mfcasecell{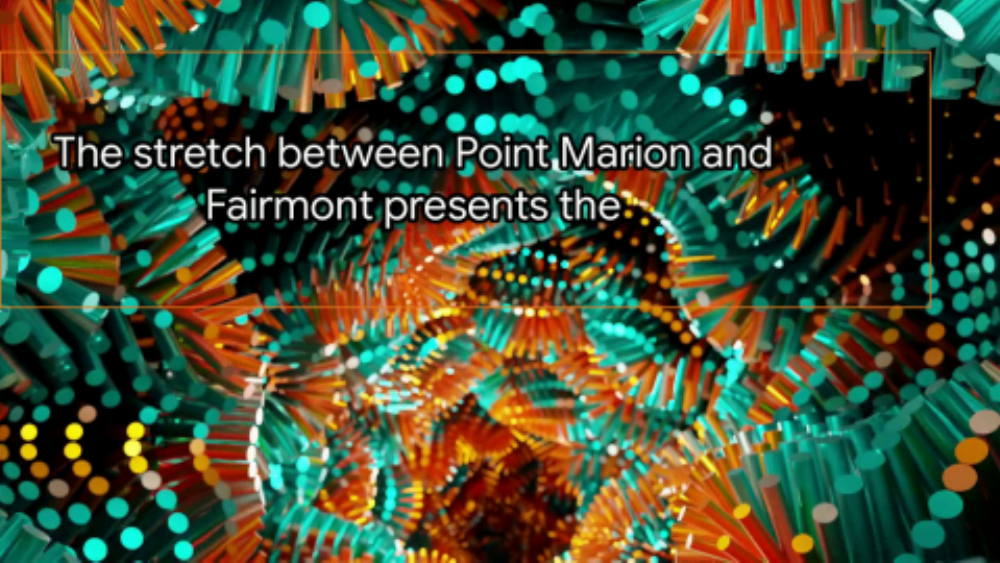} & \mfcasecell{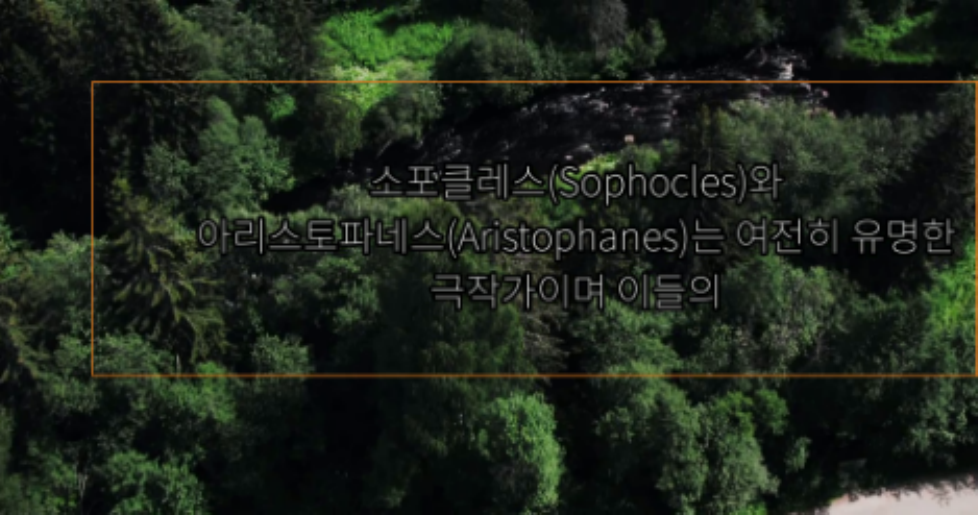} & \mfcasecell{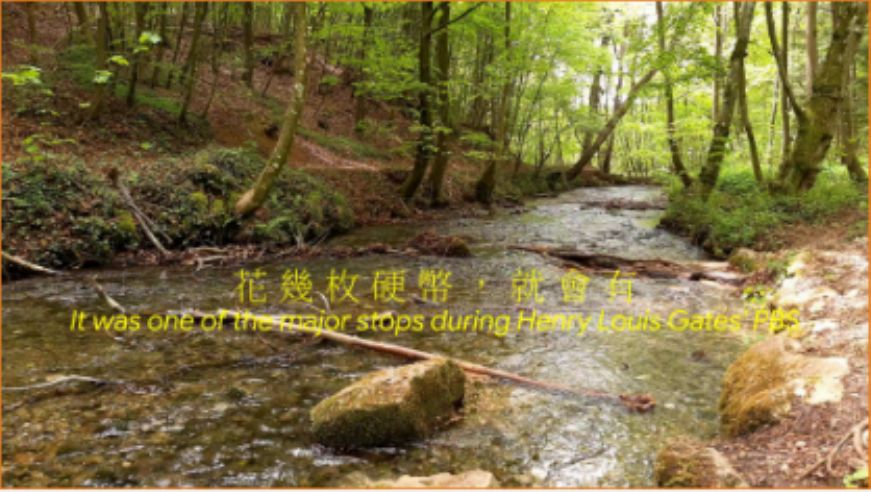} \\
Clean target & \mfcasecell{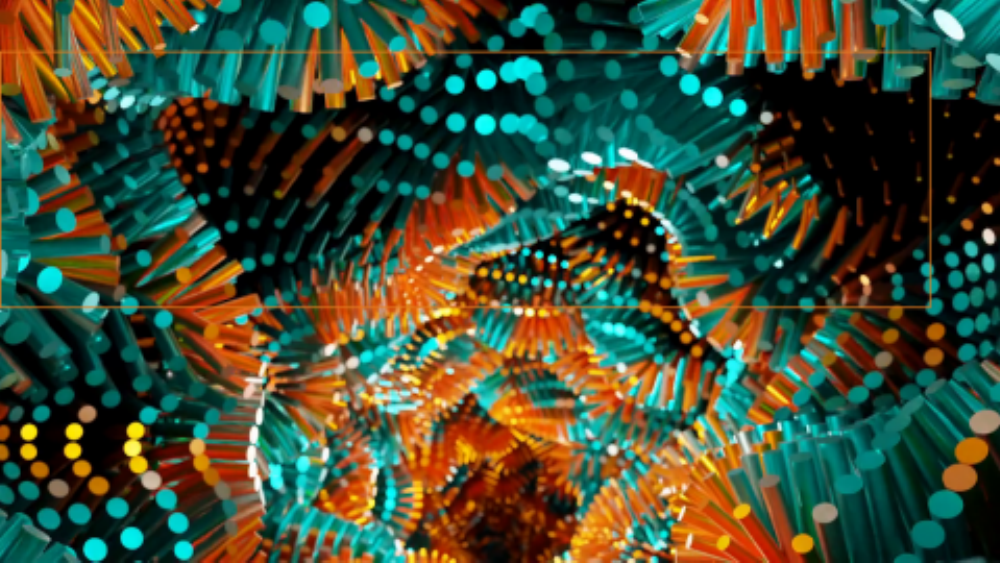} & \mfcasecell{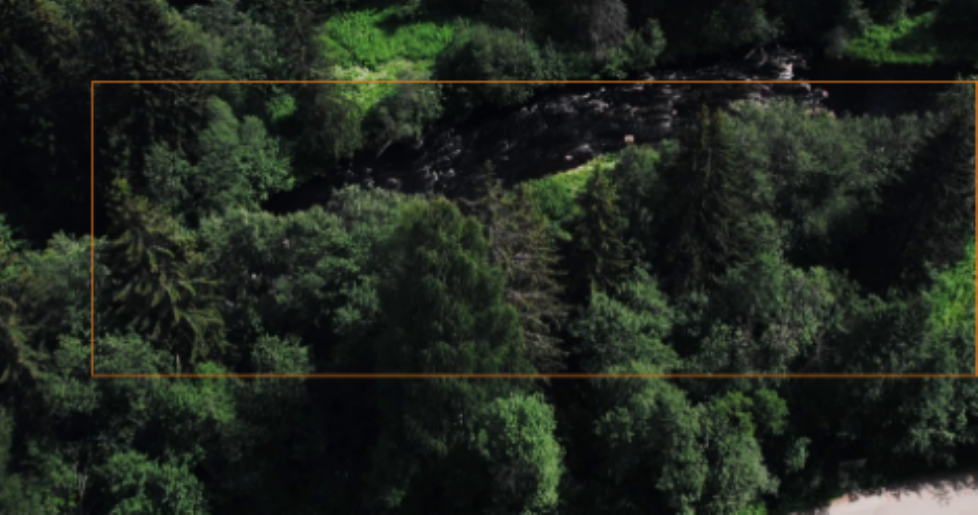} & \mfcasecell{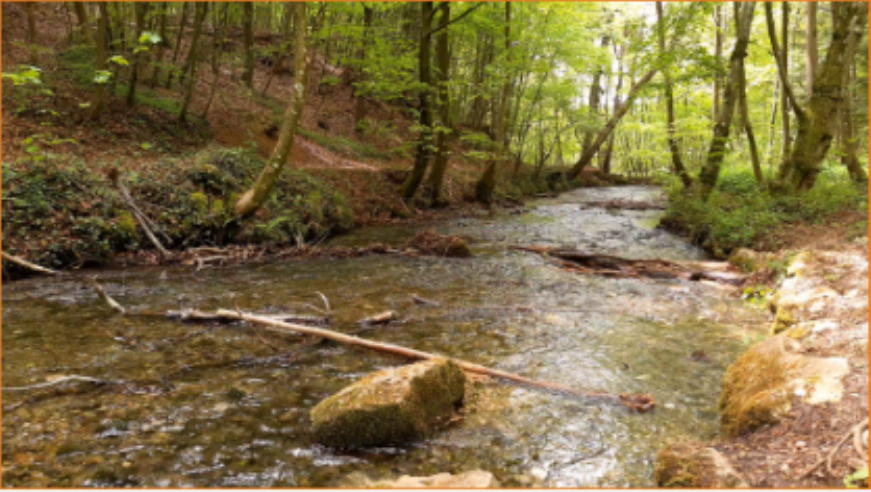} \\
\midrule
\rowcolor{AppBand}CLEAR (blind) & \mfcasecell{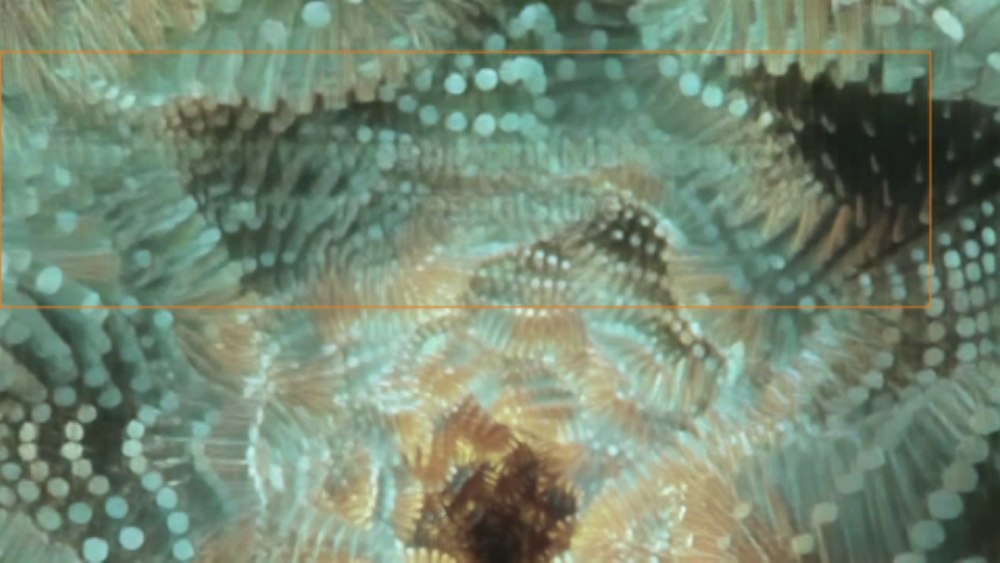} & \mfcasecell{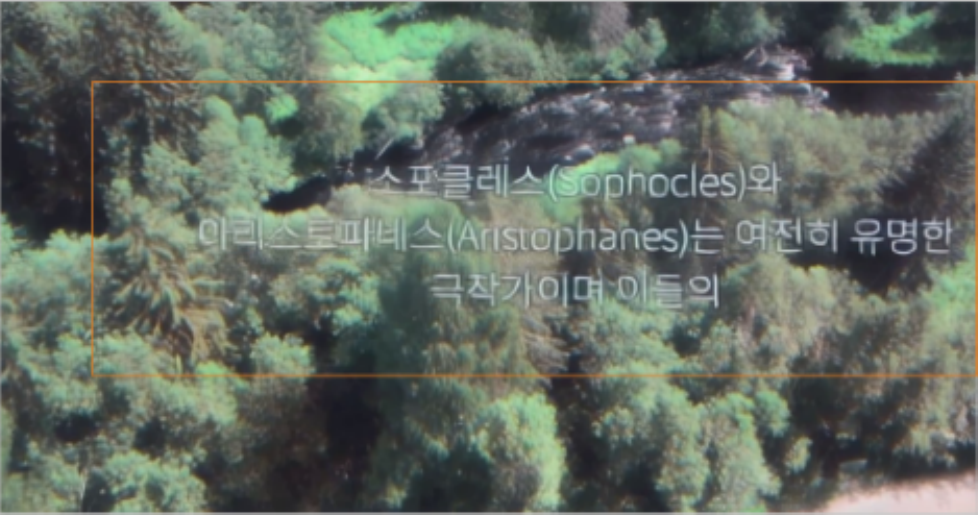} & \mfcasecell{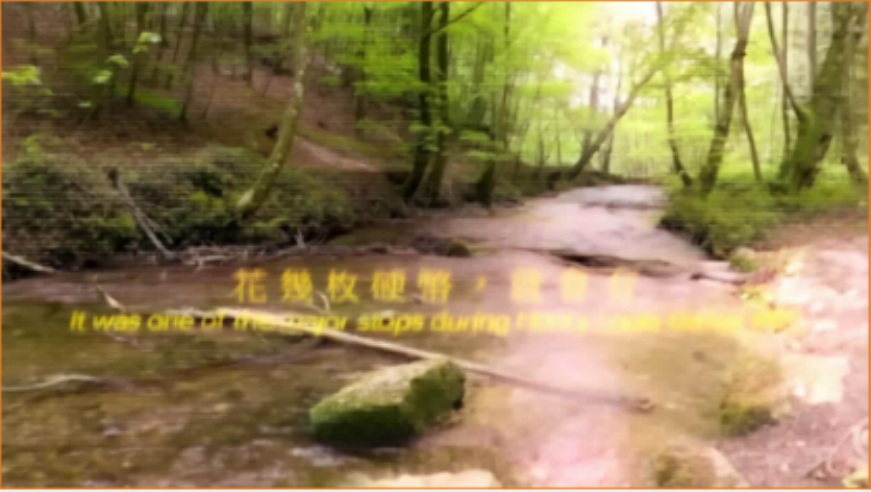} \\
ProPainter & \mfcasecell{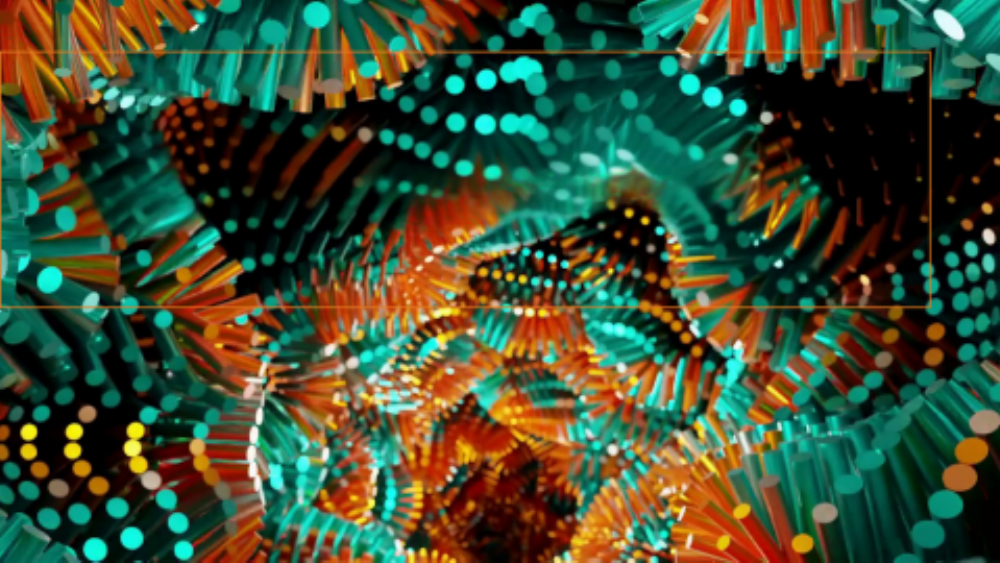} & \mfcasecell{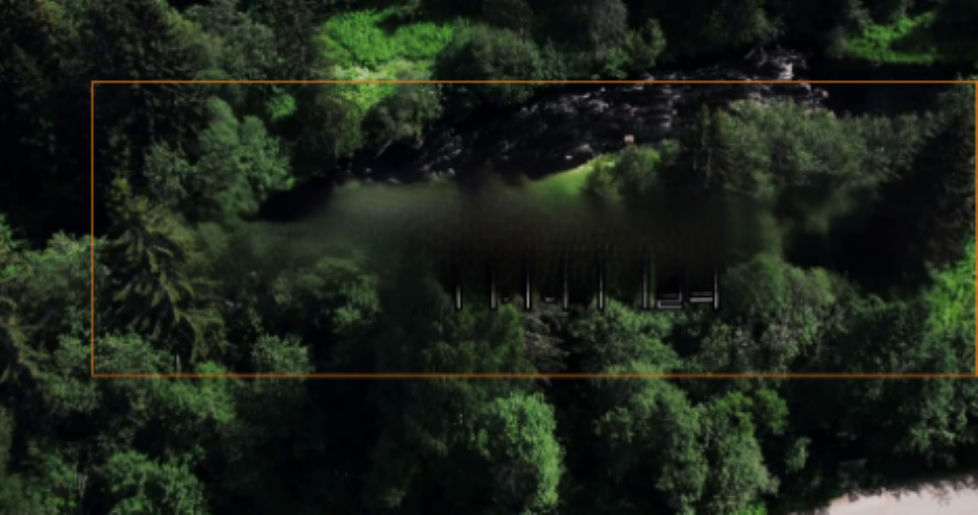} & \mfcasecell{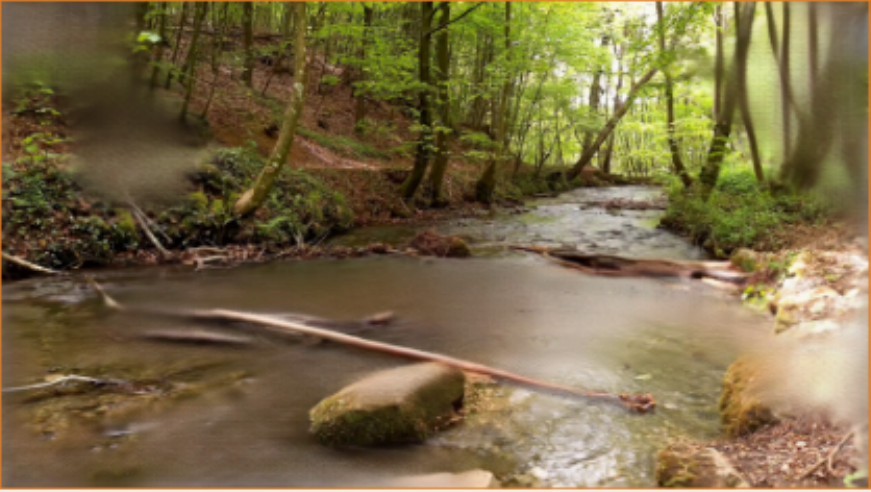} \\
\rowcolor{AppBand}DiffuEraser & \mfcasecell{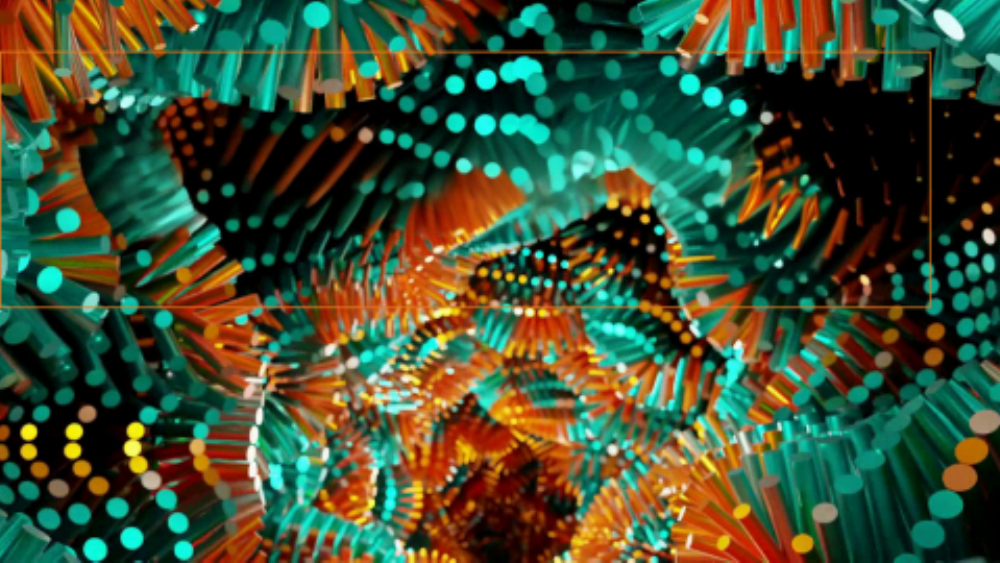} & \mfcasecell{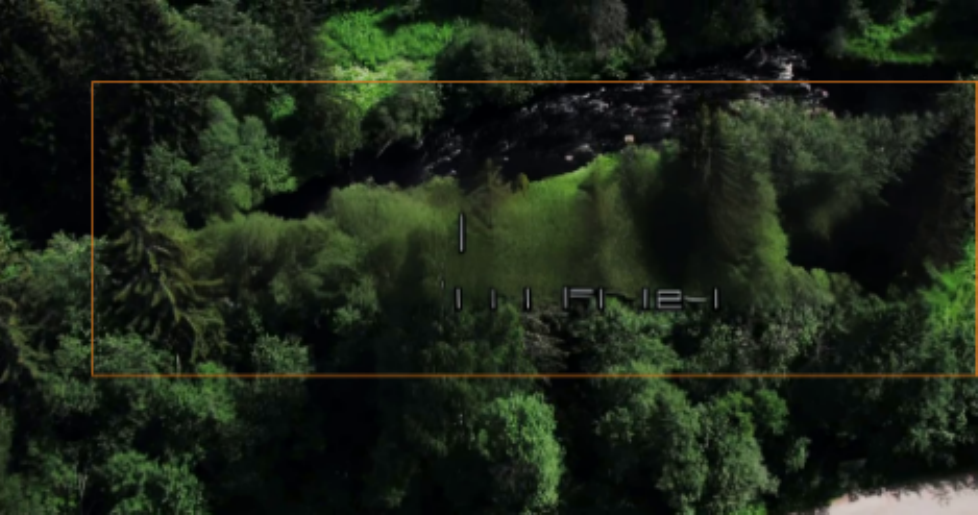} & \mfcasecell{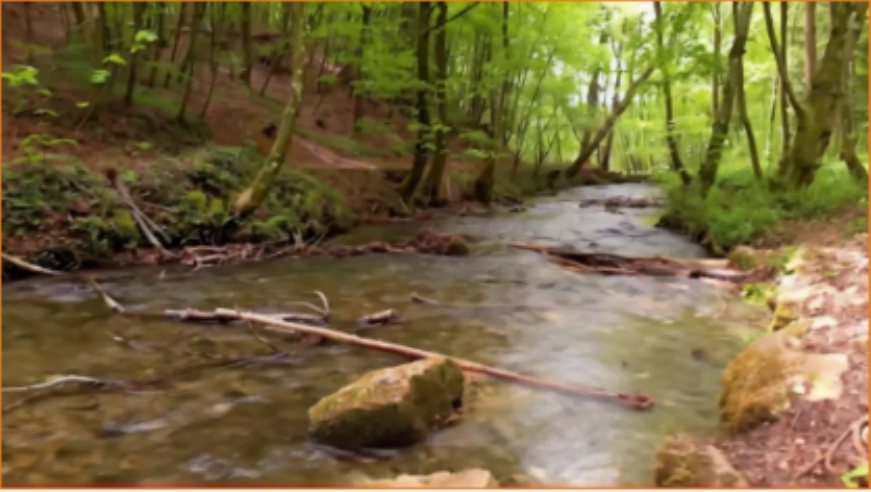} \\
MiniMax-Remover & \mfcasecell{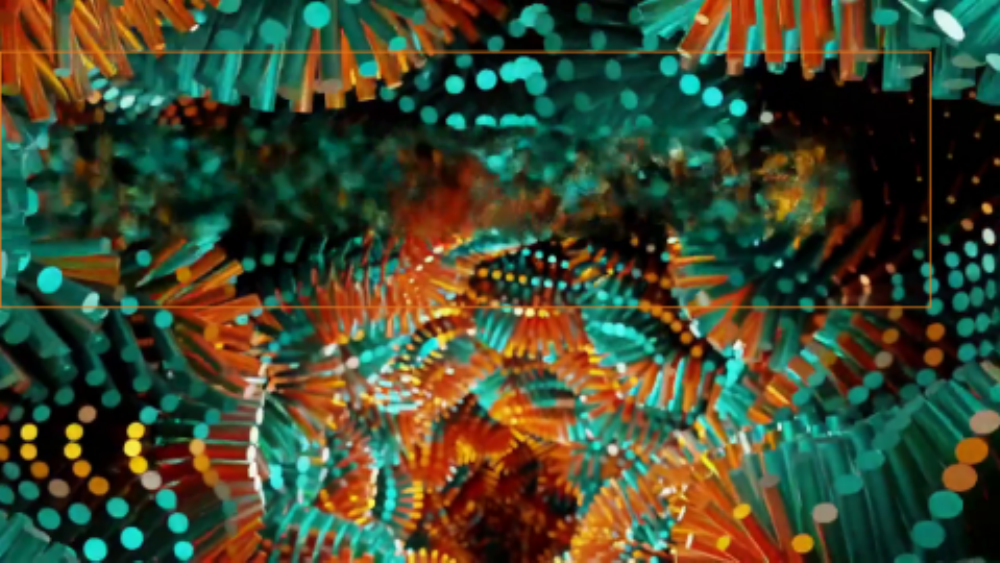} & \mfcasecell{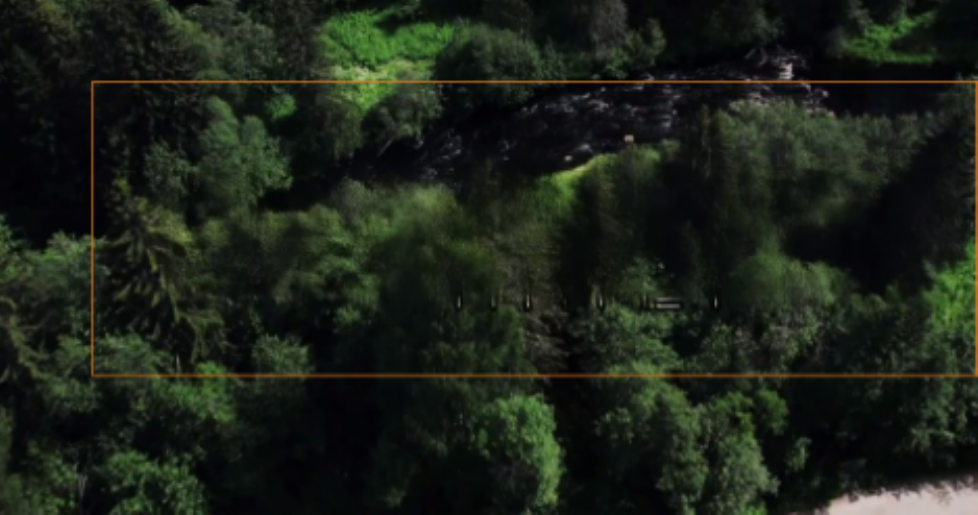} & \mfcasecell{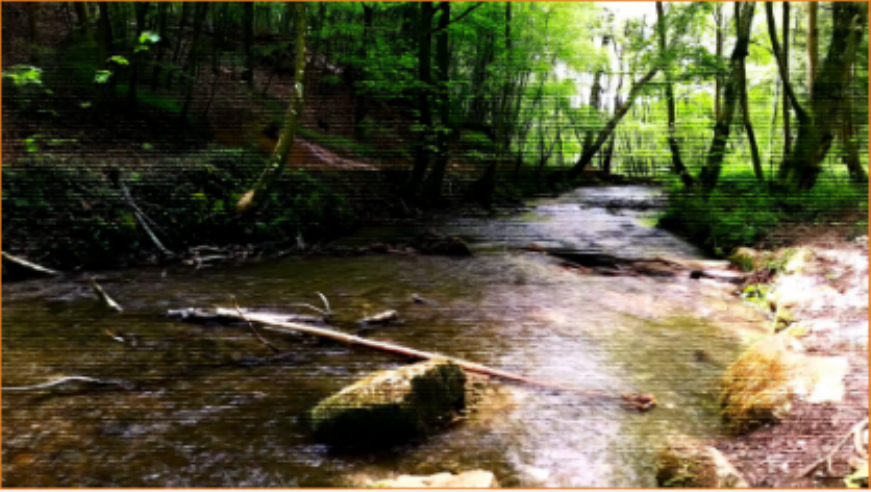} \\
\rowcolor{AppBand}OmnimatteZero & \mfcasecell{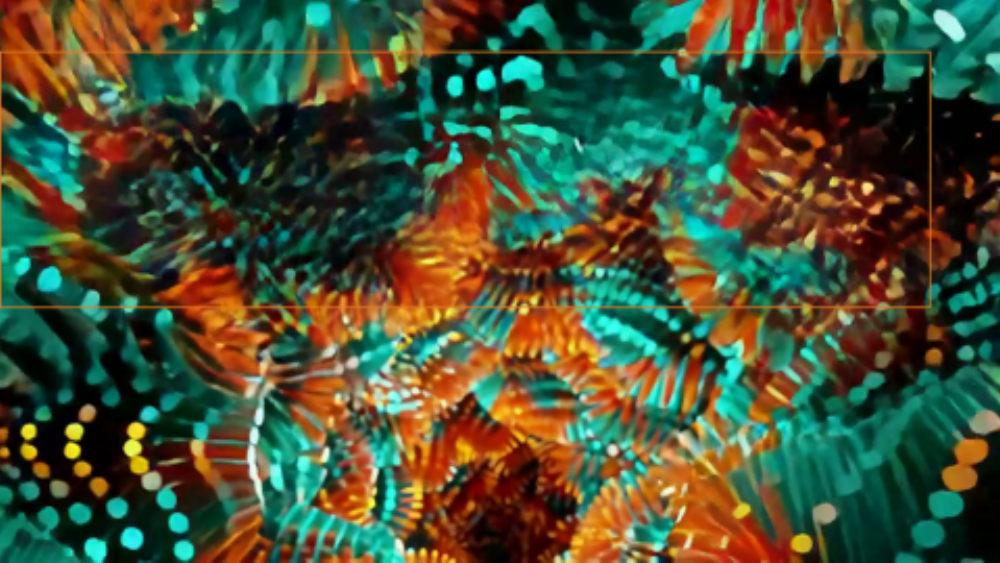} & \mfcasecell{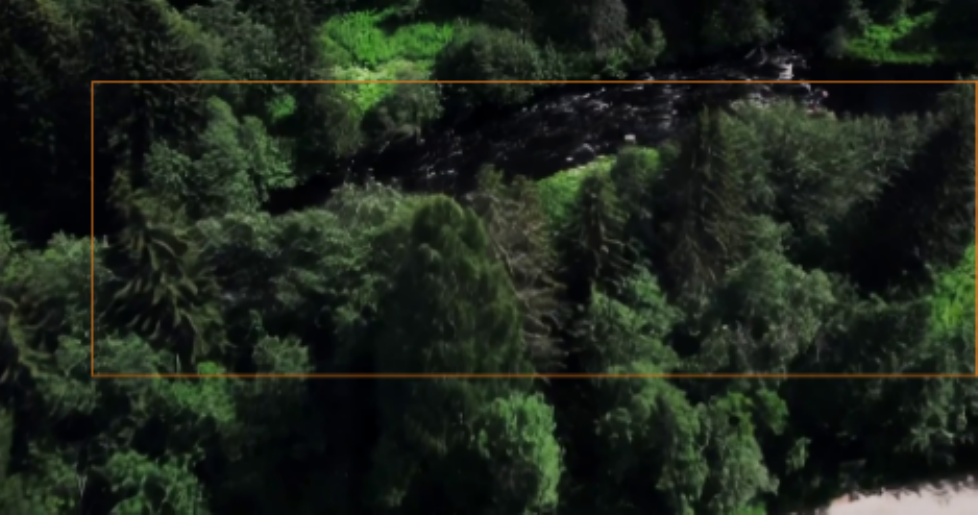} & \mfcasecell{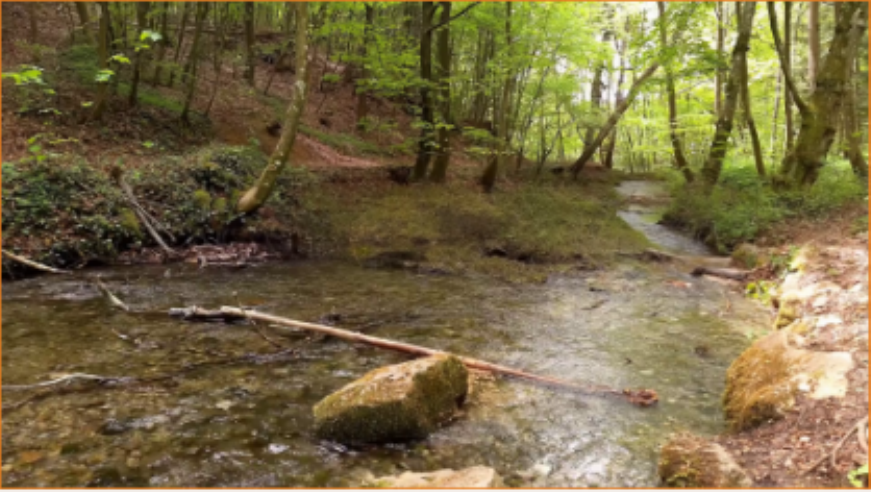} \\
EffectErase & \mfcasecell{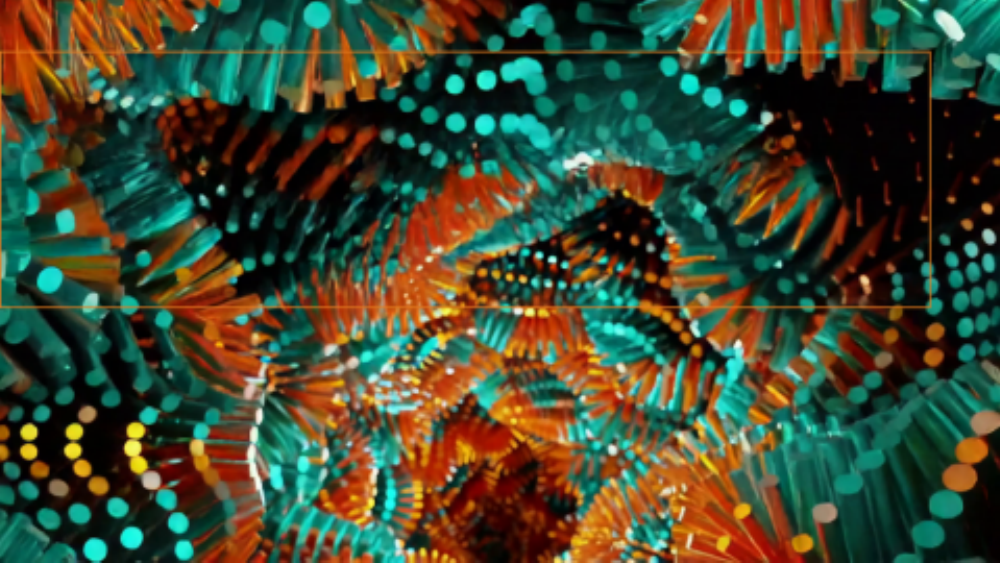} & \mfcasecell{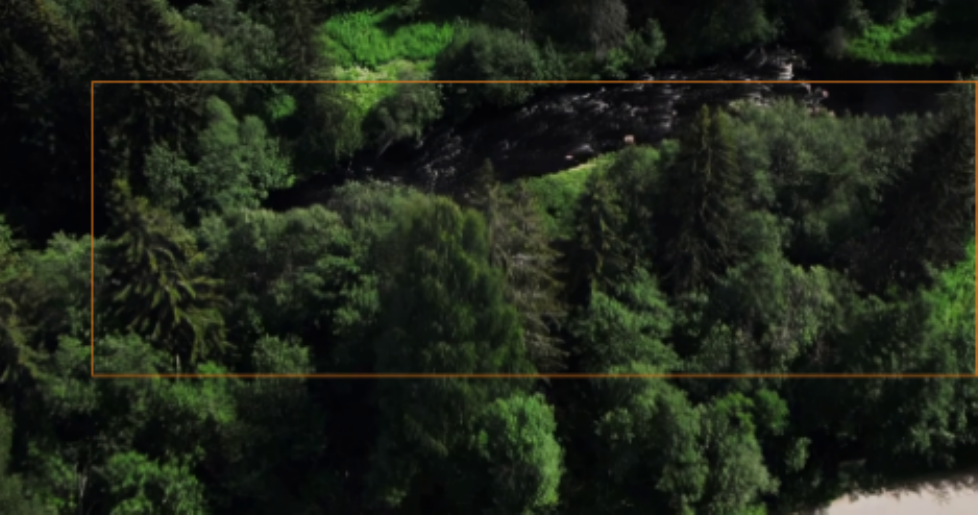} & \mfcasecell{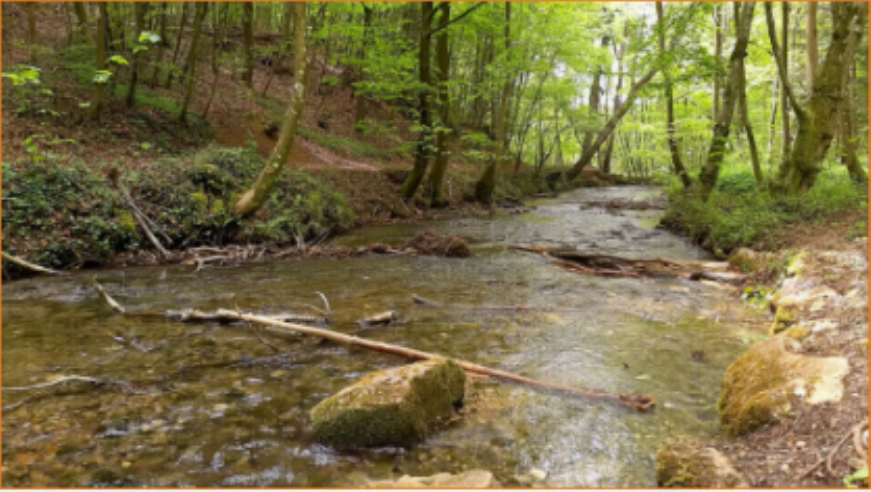} \\
\bottomrule
\end{tabularx}
\end{center}
\noindent In P1, CLEAR changes colors and geometric boundaries well beyond
the text. P2 and P3 extend the preservation comparison to foliage and flowing
water, where texture changes are easy to overlook in a local crop. Outside
fidelity records these differences independently of the repair itself.
\clearpage
\subsection*{Preservation under Moving and Animated Subtitles}
These full-frame comparisons retain grass, foliage, and scene boundaries that a subtitle crop would hide.
\par\smallskip
\begin{center}\footnotesize\setlength{\tabcolsep}{3pt}\renewcommand{\arraystretch}{1.08}
\renewcommand{\mfcasecell}[1]{\includegraphics[width=\linewidth,height=0.085\textheight,keepaspectratio]{assets/cases/#1.pdf}}
\renewcommand{\tabularxcolumn}[1]{m{#1}}
\begin{tabularx}{\textwidth}{>{\raggedright\arraybackslash}m{0.16\textwidth}>{\centering\arraybackslash}X>{\centering\arraybackslash}X}
\toprule
\rowcolor{AppHeader}
\textbf{Method / view} & \makecell{\textbf{P4} / Grass and motion\\{\scriptsize frame 74}} & \makecell{\textbf{P5} / Rotating text\\{\scriptsize frame 33}} \\
\midrule
\rowcolor{AppBand}Input & \mfcasecell{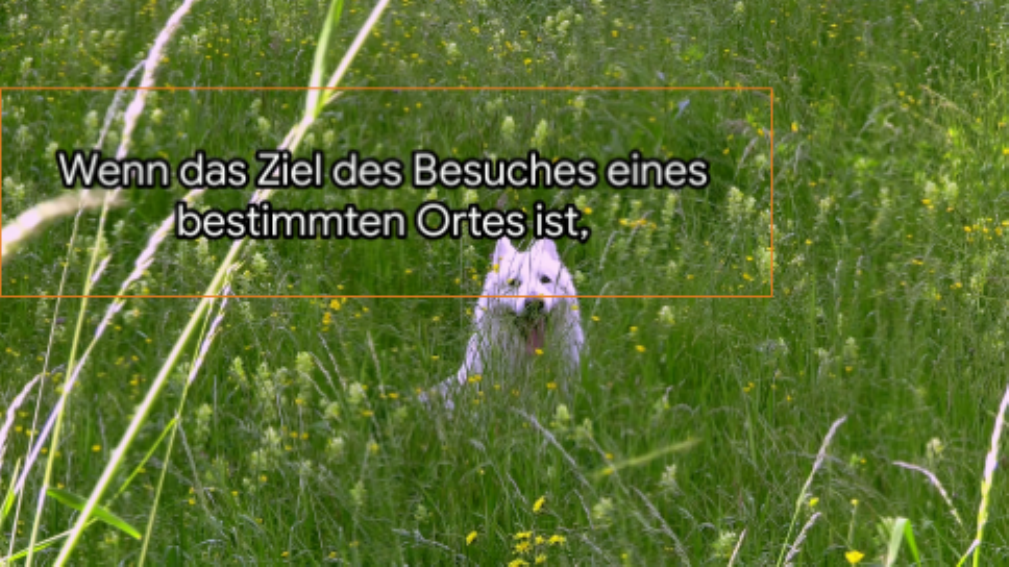} & \mfcasecell{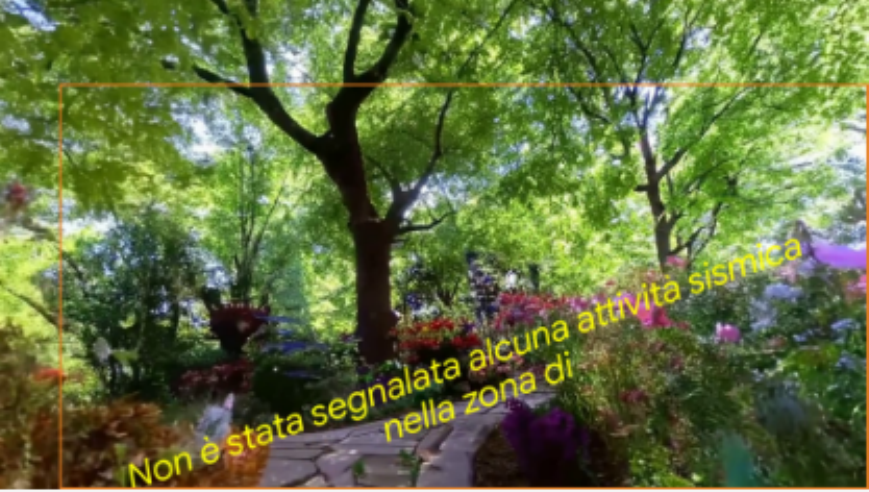} \\
Clean target & \mfcasecell{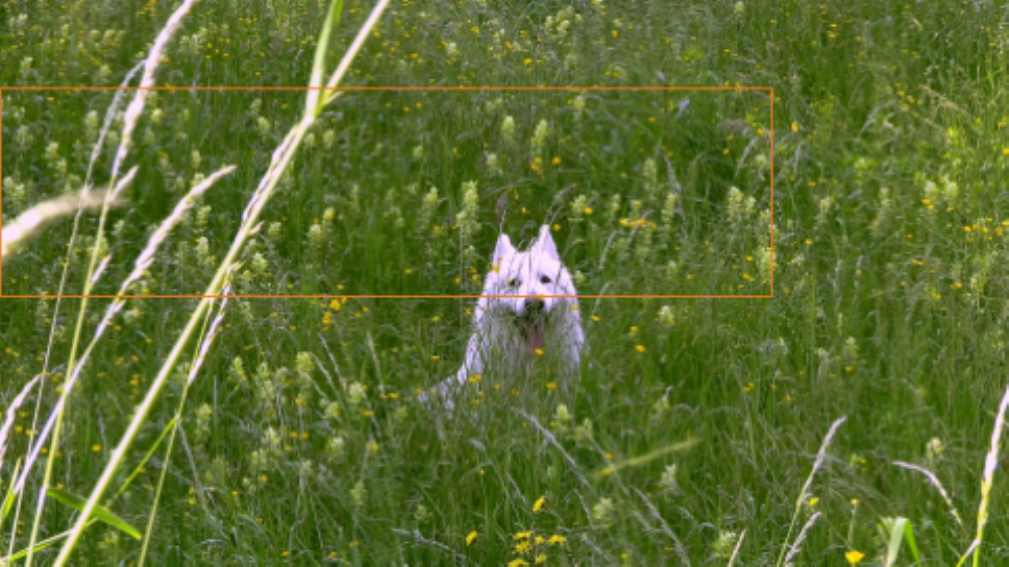} & \mfcasecell{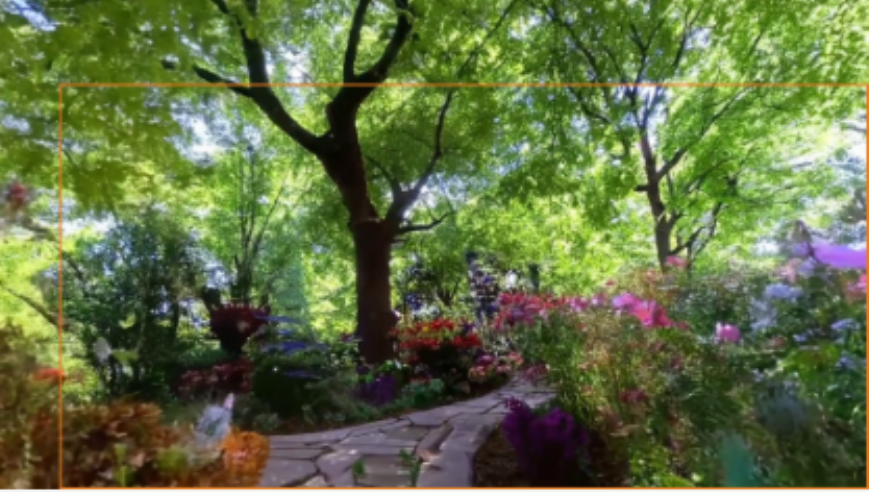} \\
\midrule
\rowcolor{AppBand}CLEAR (blind) & \mfcasecell{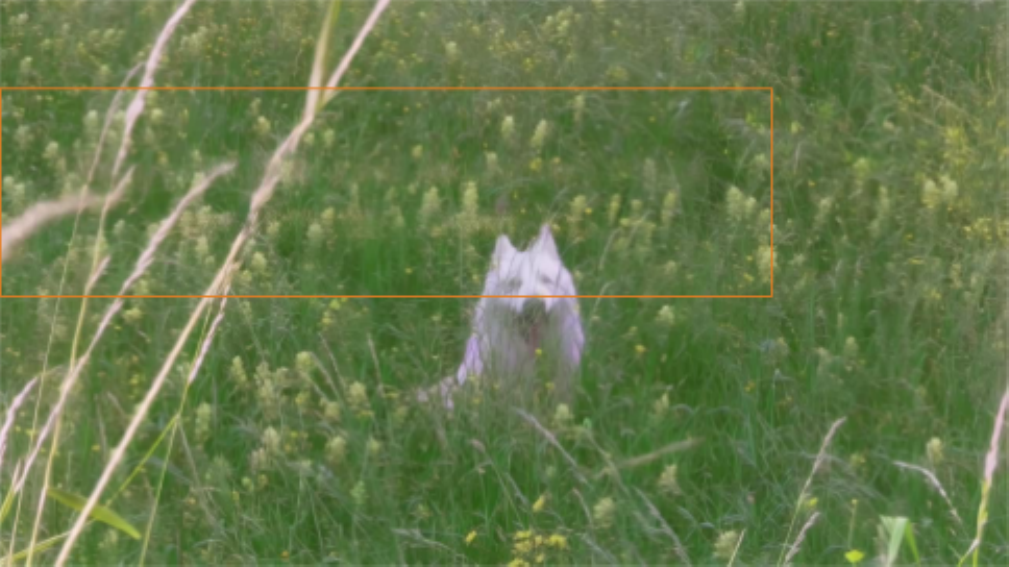} & \mfcasecell{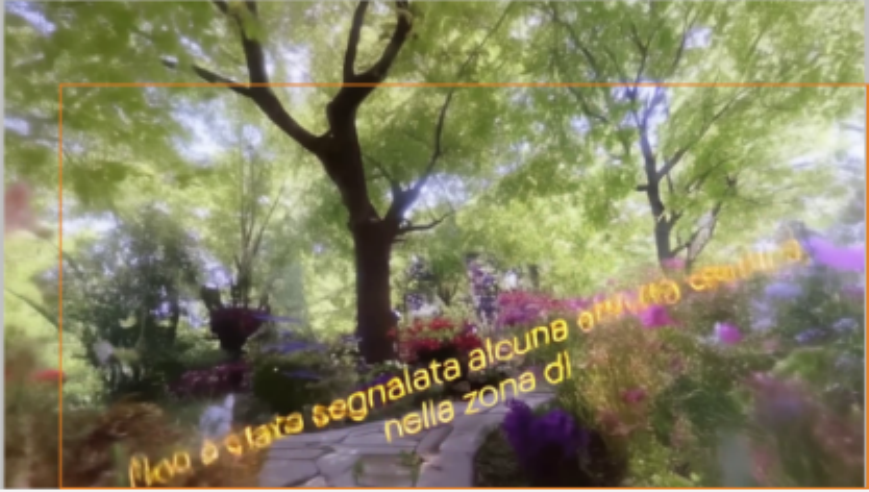} \\
ProPainter & \mfcasecell{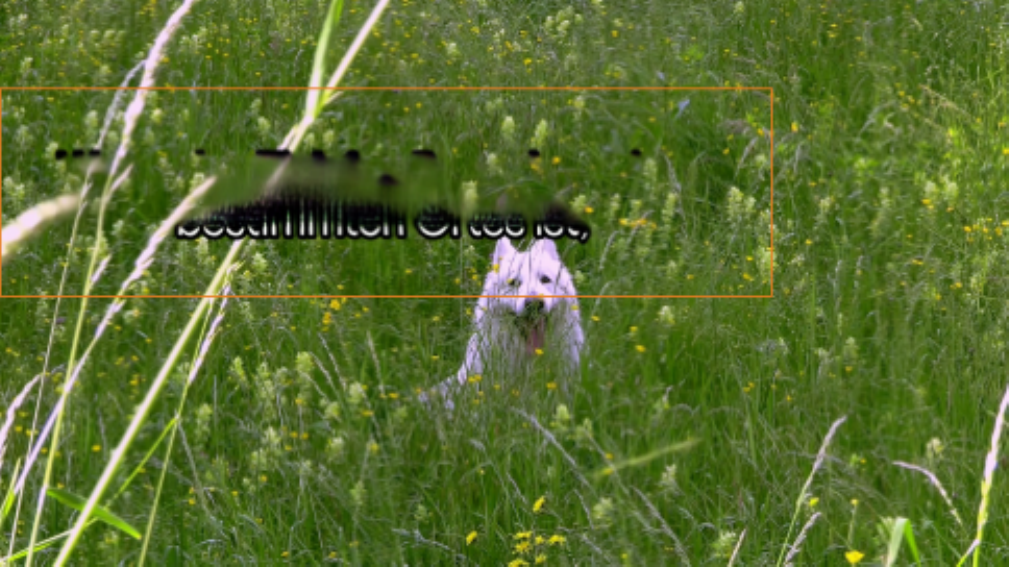} & \mfcasecell{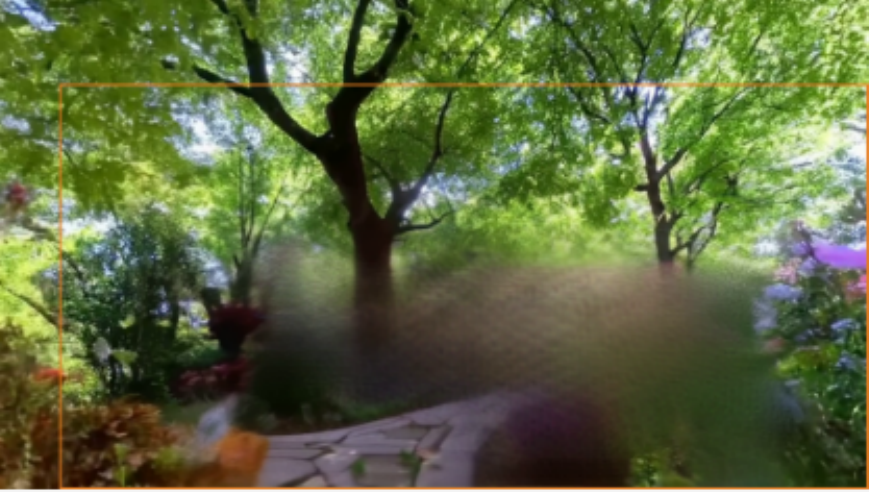} \\
\rowcolor{AppBand}DiffuEraser & \mfcasecell{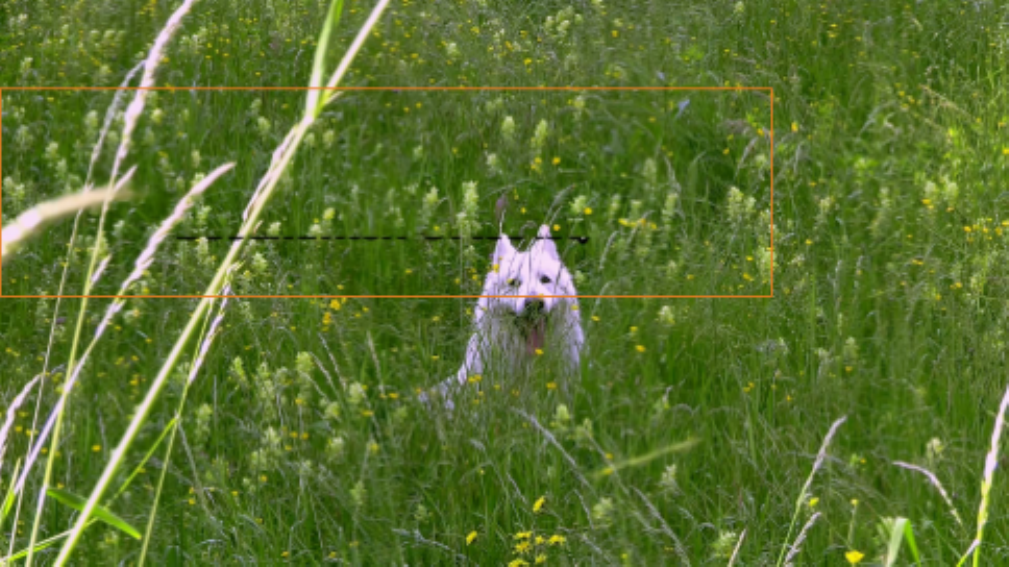} & \mfcasecell{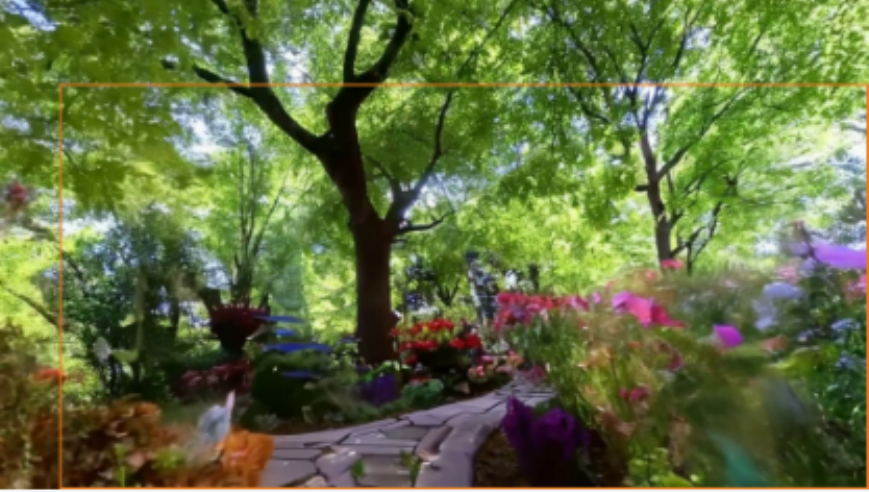} \\
MiniMax-Remover & \mfcasecell{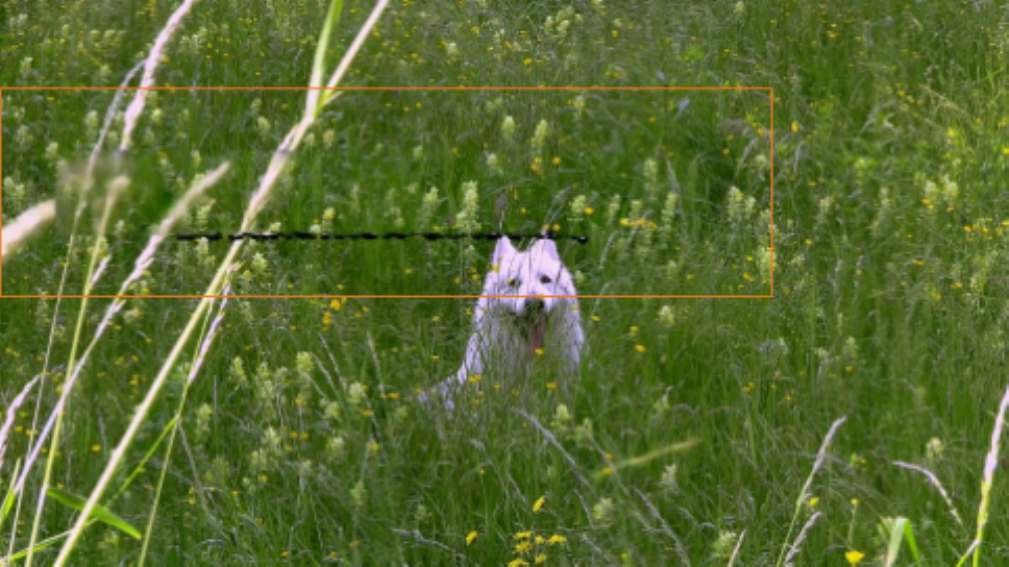} & \mfcasecell{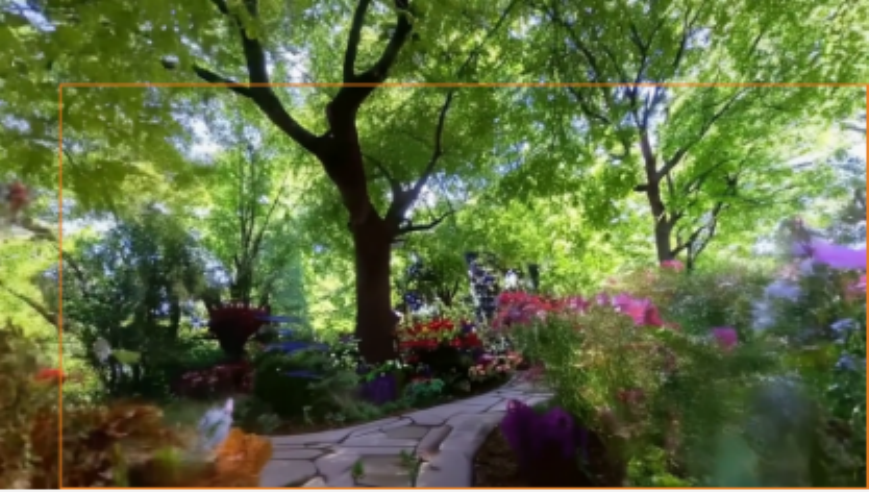} \\
\rowcolor{AppBand}OmnimatteZero & \mfcasecell{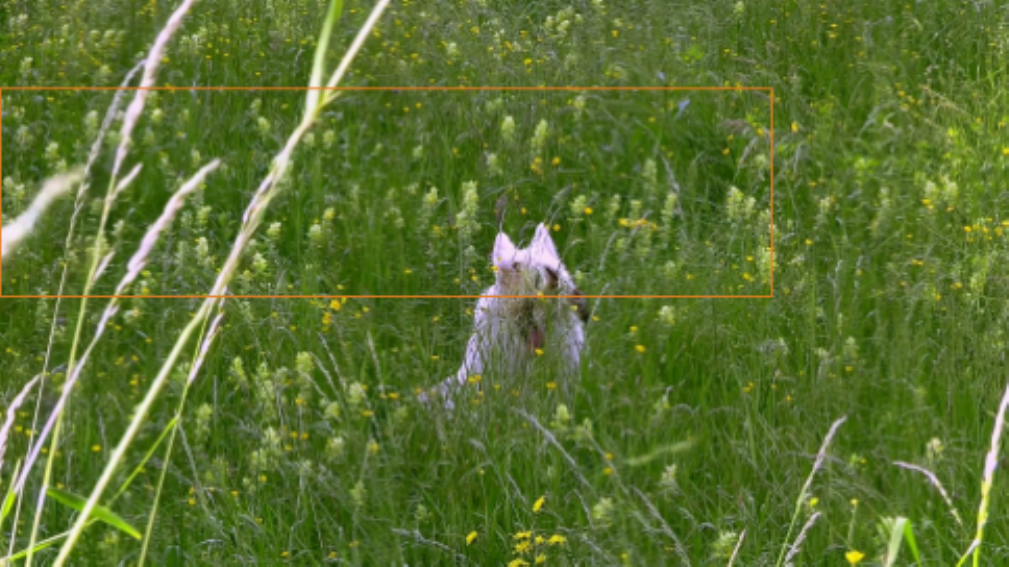} & \mfcasecell{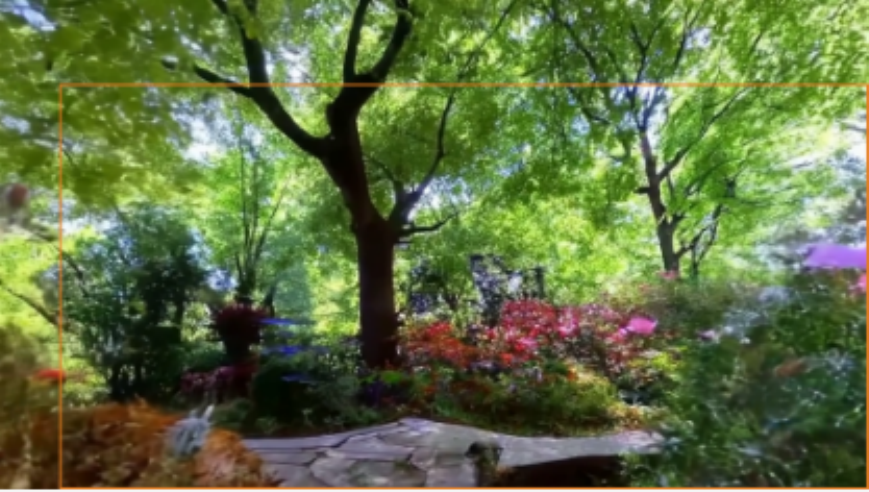} \\
EffectErase & \mfcasecell{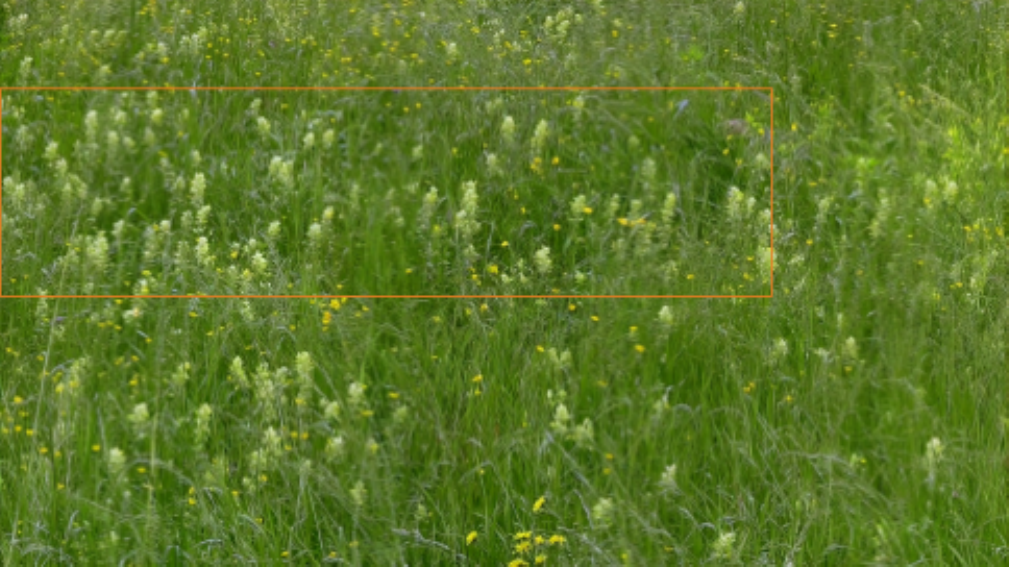} & \mfcasecell{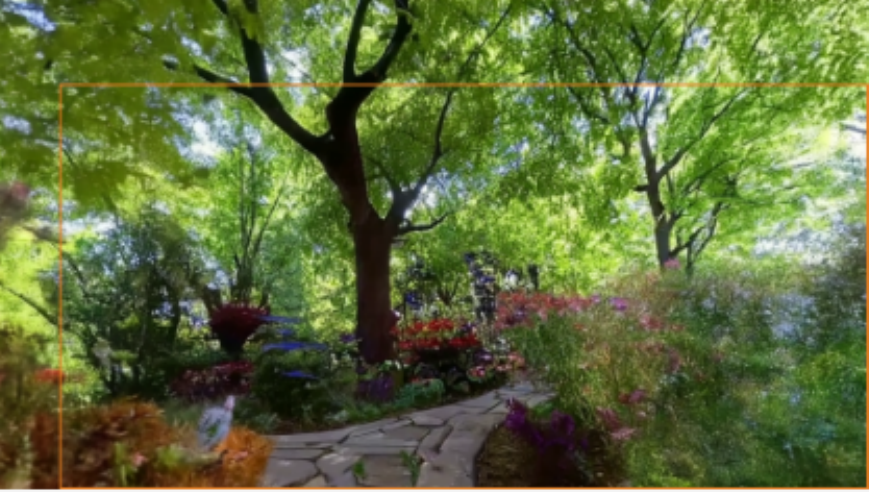} \\
\bottomrule
\end{tabularx}
\end{center}
\noindent P4 and P5 extend the comparison to moving or animated subtitles,
where the relevant background includes grass, foliage, and scene boundaries.
Such context is easy to miss when inspecting only the current text location.
Outside scores remain tied to the operational BBox: the separate box-style
stress test addresses effects that extend beyond this annotation.
\clearpage
\subsection{Temporal Fidelity: Changes Relative to the Target}
\label{app:cases-temporal}
Each strip shows frames 0--2 from a shared crop, ordered left to right.
\par\smallskip
\begingroup\footnotesize\setlength{\tabcolsep}{5pt}\setlength{\extrarowheight}{1pt}
\renewcommand{\arraystretch}{1.0}
\begin{tabularx}{\textwidth}{XX}
\toprule
\rowcolor{AppHeader}
\textbf{T1 / Bouncing neon} & \hfill Frames 0, 1, 2 $\longrightarrow$ \\
\midrule
\rowcolor{AppBand}\begin{minipage}[t]{\linewidth}\strut Input\par\vspace{2pt}\includegraphics[width=\linewidth,keepaspectratio]{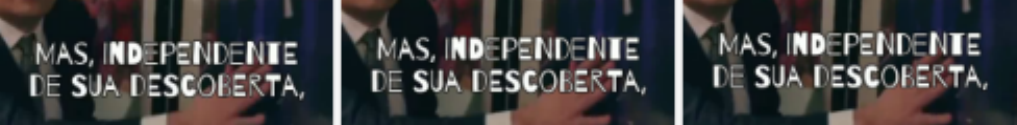}\end{minipage} & \begin{minipage}[t]{\linewidth}\strut Clean target\par\vspace{2pt}\includegraphics[width=\linewidth,keepaspectratio]{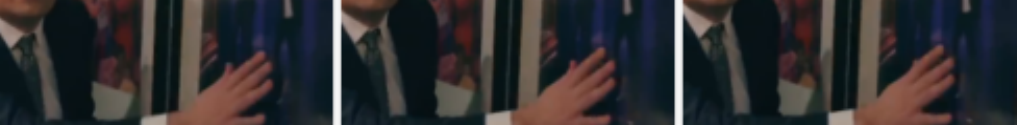}\end{minipage} \\
\begin{minipage}[t]{\linewidth}\strut CLEAR (blind)\par\vspace{2pt}\includegraphics[width=\linewidth,keepaspectratio]{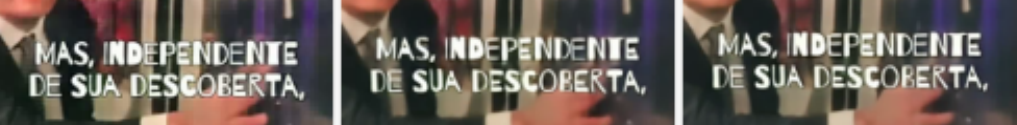}\end{minipage} & \begin{minipage}[t]{\linewidth}\strut ProPainter\par\vspace{2pt}\includegraphics[width=\linewidth,keepaspectratio]{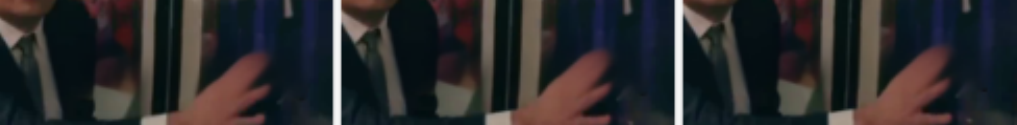}\end{minipage} \\
\rowcolor{AppBand}\begin{minipage}[t]{\linewidth}\strut DiffuEraser\par\vspace{2pt}\includegraphics[width=\linewidth,keepaspectratio]{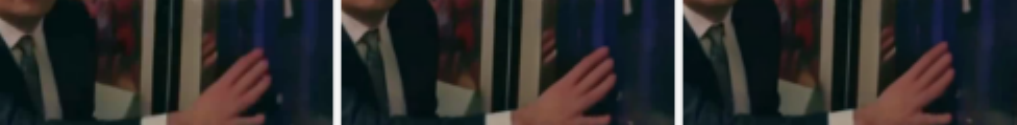}\end{minipage} & \begin{minipage}[t]{\linewidth}\strut MiniMax-Remover\par\vspace{2pt}\includegraphics[width=\linewidth,keepaspectratio]{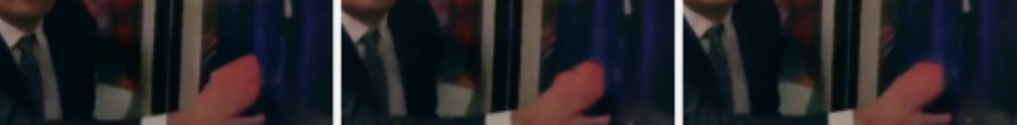}\end{minipage} \\
\begin{minipage}[t]{\linewidth}\strut OmnimatteZero\par\vspace{2pt}\includegraphics[width=\linewidth,keepaspectratio]{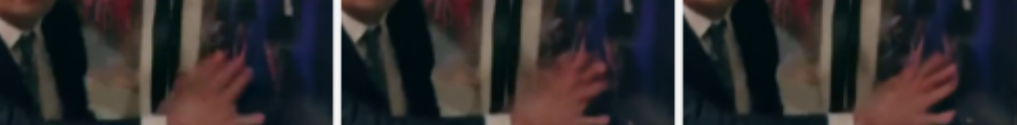}\end{minipage} & \begin{minipage}[t]{\linewidth}\strut EffectErase\par\vspace{2pt}\includegraphics[width=\linewidth,keepaspectratio]{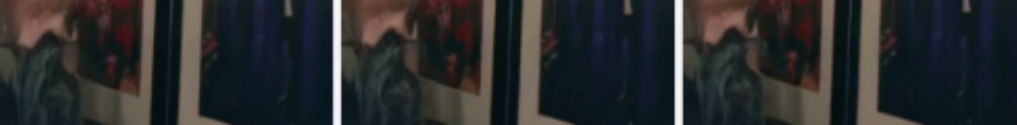}\end{minipage} \\
\bottomrule
\end{tabularx}
\endgroup
\par\smallskip
\begingroup\footnotesize\setlength{\tabcolsep}{5pt}\setlength{\extrarowheight}{1pt}
\renewcommand{\arraystretch}{1.0}
\begin{tabularx}{\textwidth}{XX}
\toprule
\rowcolor{AppHeader}
\textbf{T2 / Moving outline} & \hfill Frames 0, 1, 2 $\longrightarrow$ \\
\midrule
\rowcolor{AppBand}\begin{minipage}[t]{\linewidth}\strut Input\par\vspace{2pt}\includegraphics[width=\linewidth,keepaspectratio]{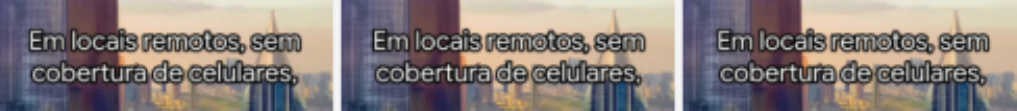}\end{minipage} & \begin{minipage}[t]{\linewidth}\strut Clean target\par\vspace{2pt}\includegraphics[width=\linewidth,keepaspectratio]{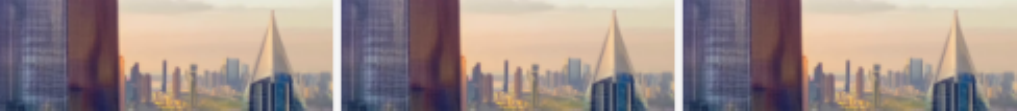}\end{minipage} \\
\begin{minipage}[t]{\linewidth}\strut CLEAR (blind)\par\vspace{2pt}\includegraphics[width=\linewidth,keepaspectratio]{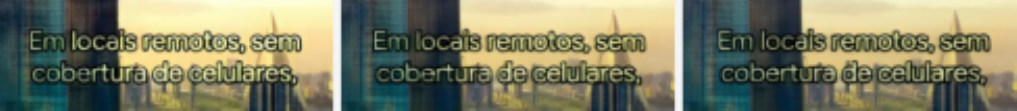}\end{minipage} & \begin{minipage}[t]{\linewidth}\strut ProPainter\par\vspace{2pt}\includegraphics[width=\linewidth,keepaspectratio]{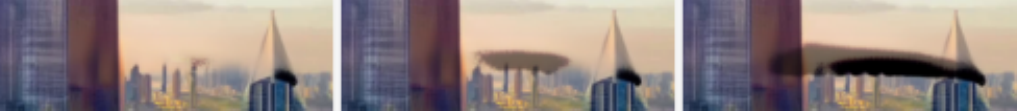}\end{minipage} \\
\rowcolor{AppBand}\begin{minipage}[t]{\linewidth}\strut DiffuEraser\par\vspace{2pt}\includegraphics[width=\linewidth,keepaspectratio]{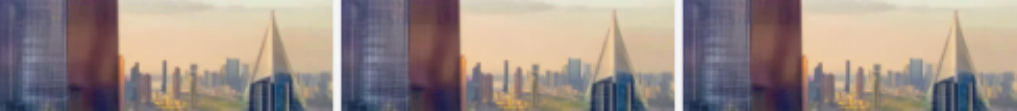}\end{minipage} & \begin{minipage}[t]{\linewidth}\strut MiniMax-Remover\par\vspace{2pt}\includegraphics[width=\linewidth,keepaspectratio]{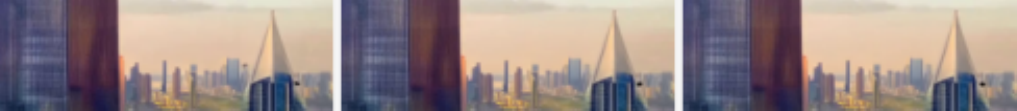}\end{minipage} \\
\begin{minipage}[t]{\linewidth}\strut OmnimatteZero\par\vspace{2pt}\includegraphics[width=\linewidth,keepaspectratio]{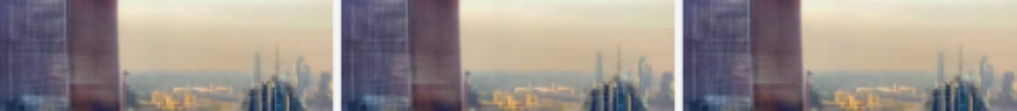}\end{minipage} & \begin{minipage}[t]{\linewidth}\strut EffectErase\par\vspace{2pt}\includegraphics[width=\linewidth,keepaspectratio]{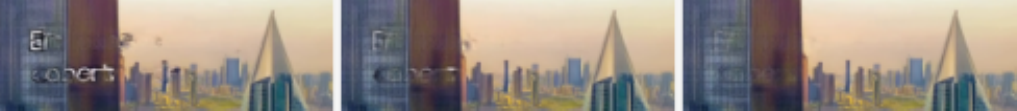}\end{minipage} \\
\bottomrule
\end{tabularx}
\endgroup
\par\smallskip
\begingroup\footnotesize\setlength{\tabcolsep}{5pt}\setlength{\extrarowheight}{1pt}
\renewcommand{\arraystretch}{1.0}
\begin{tabularx}{\textwidth}{XX}
\toprule
\rowcolor{AppHeader}
\textbf{T3 / Sliding outline} & \hfill Frames 0, 1, 2 $\longrightarrow$ \\
\midrule
\rowcolor{AppBand}\begin{minipage}[t]{\linewidth}\strut Input\par\vspace{2pt}\includegraphics[width=\linewidth,keepaspectratio]{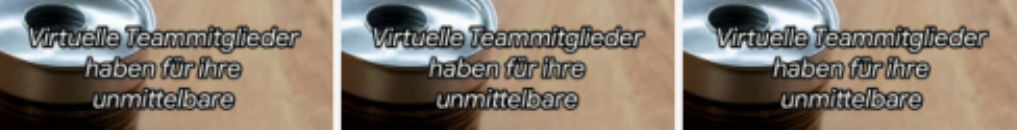}\end{minipage} & \begin{minipage}[t]{\linewidth}\strut Clean target\par\vspace{2pt}\includegraphics[width=\linewidth,keepaspectratio]{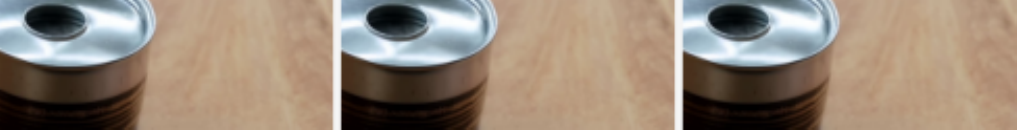}\end{minipage} \\
\begin{minipage}[t]{\linewidth}\strut CLEAR (blind)\par\vspace{2pt}\includegraphics[width=\linewidth,keepaspectratio]{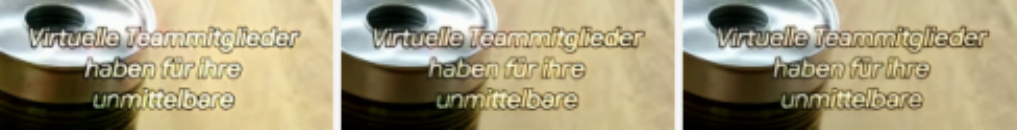}\end{minipage} & \begin{minipage}[t]{\linewidth}\strut ProPainter\par\vspace{2pt}\includegraphics[width=\linewidth,keepaspectratio]{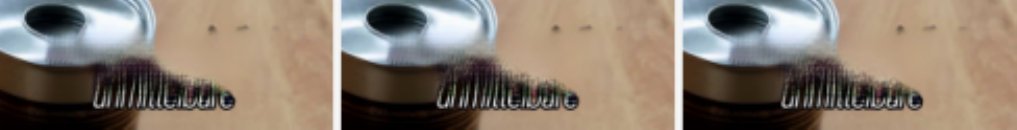}\end{minipage} \\
\rowcolor{AppBand}\begin{minipage}[t]{\linewidth}\strut DiffuEraser\par\vspace{2pt}\includegraphics[width=\linewidth,keepaspectratio]{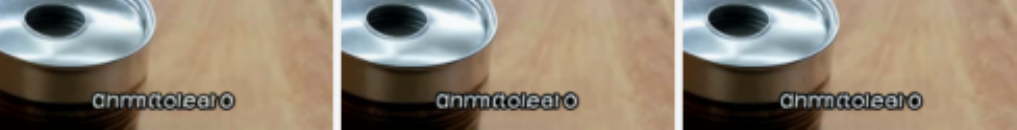}\end{minipage} & \begin{minipage}[t]{\linewidth}\strut MiniMax-Remover\par\vspace{2pt}\includegraphics[width=\linewidth,keepaspectratio]{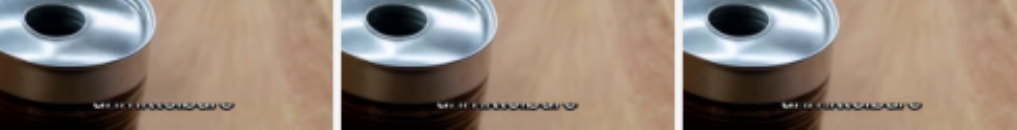}\end{minipage} \\
\begin{minipage}[t]{\linewidth}\strut OmnimatteZero\par\vspace{2pt}\includegraphics[width=\linewidth,keepaspectratio]{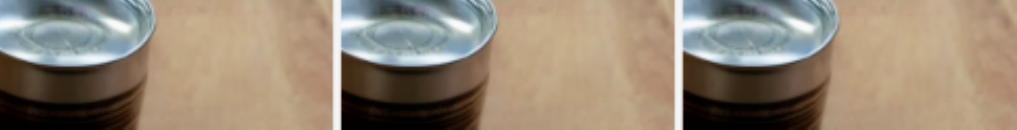}\end{minipage} & \begin{minipage}[t]{\linewidth}\strut EffectErase\par\vspace{2pt}\includegraphics[width=\linewidth,keepaspectratio]{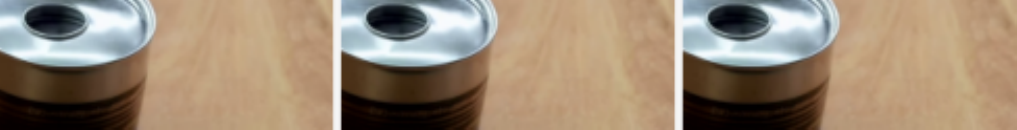}\end{minipage} \\
\bottomrule
\end{tabularx}
\endgroup

\smallskip
\noindent Subtitle remnants persist across the strips for CLEAR in T1 and
ProPainter in T3. Such errors can look stable, yet differ from the clean
target. Temporal fidelity must account for target changes as well as output
stability, helping explain the TC--TWE-Excess disagreement in
Appendix~\ref{app:finding1-details}.

\clearpage
\subsection*{Temporal Fidelity under Subtitle Motion}
T4 changes subtitle position without character animation; T5 combines
motion with bilingual shadowed text.
\par\smallskip
\begingroup\footnotesize\setlength{\tabcolsep}{5pt}\setlength{\extrarowheight}{1pt}
\renewcommand{\arraystretch}{1.0}
\begin{tabularx}{\textwidth}{XX}
\toprule
\rowcolor{AppHeader}
\textbf{T4 / Moving static text} & \hfill Frames 0, 1, 2 $\longrightarrow$ \\
\midrule
\rowcolor{AppBand}\begin{minipage}[t]{\linewidth}\strut Input\par\vspace{2pt}\includegraphics[width=\linewidth,keepaspectratio]{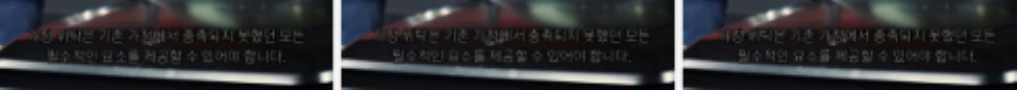}\end{minipage} & \begin{minipage}[t]{\linewidth}\strut Clean target\par\vspace{2pt}\includegraphics[width=\linewidth,keepaspectratio]{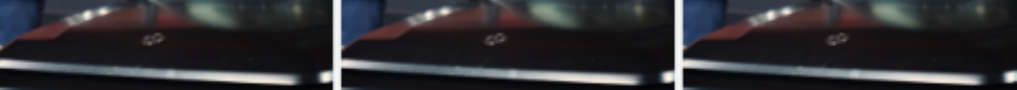}\end{minipage} \\
\begin{minipage}[t]{\linewidth}\strut CLEAR (blind)\par\vspace{2pt}\includegraphics[width=\linewidth,keepaspectratio]{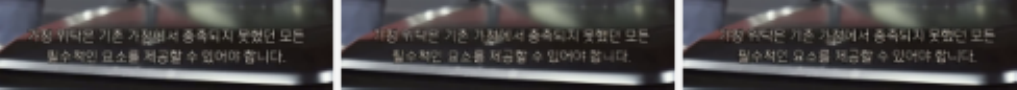}\end{minipage} & \begin{minipage}[t]{\linewidth}\strut ProPainter\par\vspace{2pt}\includegraphics[width=\linewidth,keepaspectratio]{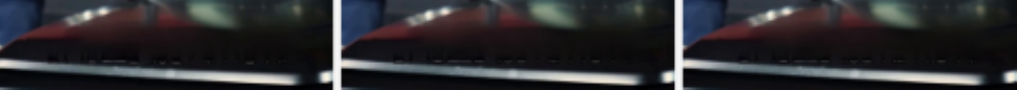}\end{minipage} \\
\rowcolor{AppBand}\begin{minipage}[t]{\linewidth}\strut DiffuEraser\par\vspace{2pt}\includegraphics[width=\linewidth,keepaspectratio]{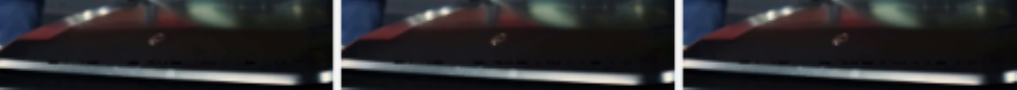}\end{minipage} & \begin{minipage}[t]{\linewidth}\strut MiniMax-Remover\par\vspace{2pt}\includegraphics[width=\linewidth,keepaspectratio]{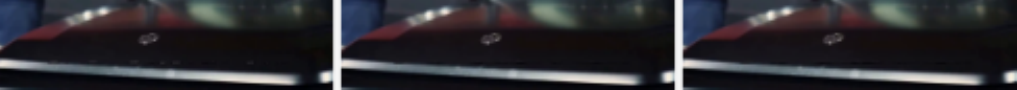}\end{minipage} \\
\begin{minipage}[t]{\linewidth}\strut OmnimatteZero\par\vspace{2pt}\includegraphics[width=\linewidth,keepaspectratio]{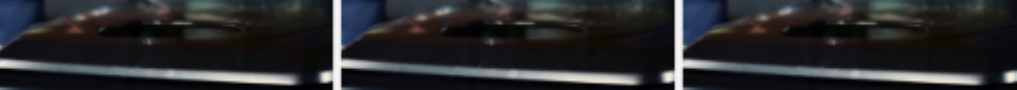}\end{minipage} & \begin{minipage}[t]{\linewidth}\strut EffectErase\par\vspace{2pt}\includegraphics[width=\linewidth,keepaspectratio]{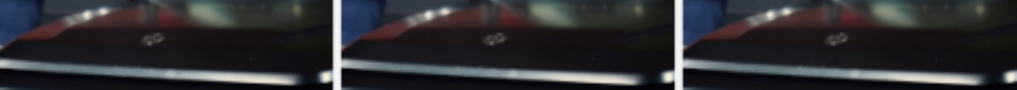}\end{minipage} \\
\bottomrule
\end{tabularx}
\endgroup
\par\smallskip
\begingroup\footnotesize\setlength{\tabcolsep}{5pt}\setlength{\extrarowheight}{1pt}
\renewcommand{\arraystretch}{1.0}
\begin{tabularx}{\textwidth}{XX}
\toprule
\rowcolor{AppHeader}
\textbf{T5 / Moving bilingual} & \hfill Frames 0, 1, 2 $\longrightarrow$ \\
\midrule
\rowcolor{AppBand}\begin{minipage}[t]{\linewidth}\strut Input\par\vspace{2pt}\includegraphics[width=\linewidth,keepaspectratio]{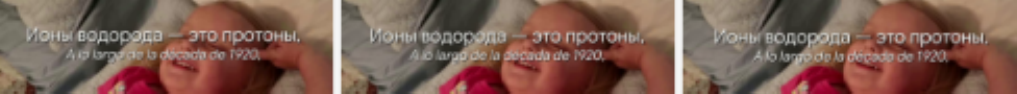}\end{minipage} & \begin{minipage}[t]{\linewidth}\strut Clean target\par\vspace{2pt}\includegraphics[width=\linewidth,keepaspectratio]{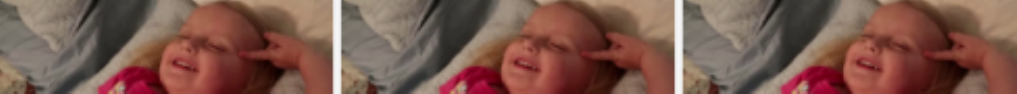}\end{minipage} \\
\begin{minipage}[t]{\linewidth}\strut CLEAR (blind)\par\vspace{2pt}\includegraphics[width=\linewidth,keepaspectratio]{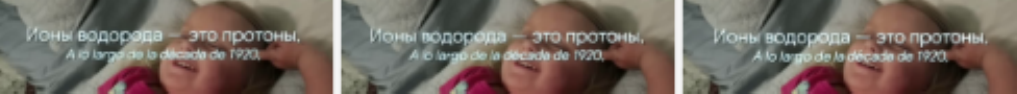}\end{minipage} & \begin{minipage}[t]{\linewidth}\strut ProPainter\par\vspace{2pt}\includegraphics[width=\linewidth,keepaspectratio]{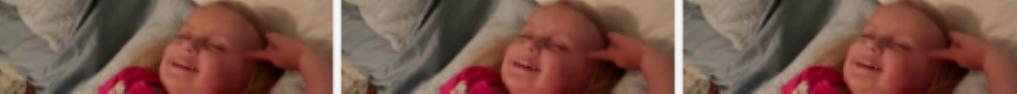}\end{minipage} \\
\rowcolor{AppBand}\begin{minipage}[t]{\linewidth}\strut DiffuEraser\par\vspace{2pt}\includegraphics[width=\linewidth,keepaspectratio]{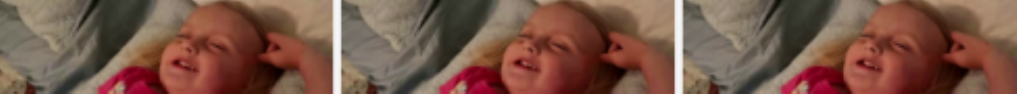}\end{minipage} & \begin{minipage}[t]{\linewidth}\strut MiniMax-Remover\par\vspace{2pt}\includegraphics[width=\linewidth,keepaspectratio]{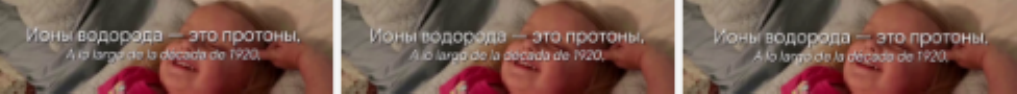}\end{minipage} \\
\begin{minipage}[t]{\linewidth}\strut OmnimatteZero\par\vspace{2pt}\includegraphics[width=\linewidth,keepaspectratio]{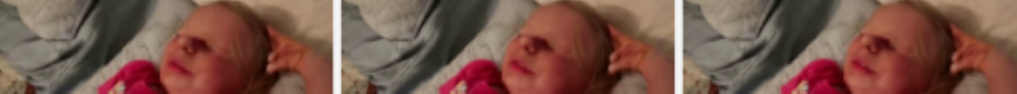}\end{minipage} & \begin{minipage}[t]{\linewidth}\strut EffectErase\par\vspace{2pt}\includegraphics[width=\linewidth,keepaspectratio]{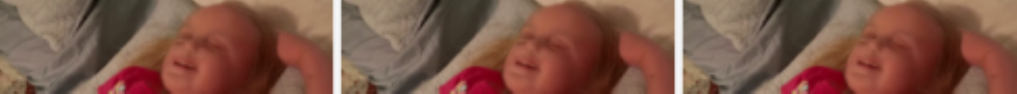}\end{minipage} \\
\bottomrule
\end{tabularx}
\endgroup

\medskip
\noindent Spatial errors and temporal changes can occur together, making
an overall visual preference difficult to interpret. The separate
Restoration and Temporal tasks help distinguish these judgments, though the
high undecided rate in the held-out temporal study limits conclusions about
metric alignment (Appendix~\ref{app:finding5-human-qualitative}).

Three-frame strips locate transitions for inspection; they cannot establish
perceptible flicker over a full video. That assessment requires playback on
shared support, as provided by the original evaluation interface in
Figure~\ref{fig:human-evaluation-interface}.

\end{document}